\pdfoutput=1
\documentclass[12pt]{article}
\input{preamble/preamble}

\newcommand{\appropto}{\mathrel{\vcenter{
  \offinterlineskip\halign{\hfil$##$\cr
    \propto\cr\noalign{\kern2pt}\sim\cr\noalign{\kern-2pt}}}}}

\crefname{lemma}{lemma}{lemmas}
\Crefname{lemma}{Lemma}{Lemmas}
\crefname{thm}{theorem}{theorems}
\Crefname{thm}{Theorem}{Theorems}
\crefname{prop}{proposition}{propositions}
\Crefname{prop}{Proposition}{Propositions}
\crefname{defn}{definition}{definitions}
\Crefname{defn}{Definition}{Definitions}
\crefname{problem}{problem}{problems}
\Crefname{problem}{Problem}{Problems}

\newtheorem{lemma}{Lemma}
\newtheorem{fact}{Fact}
\newtheorem{theorem}{Theorem}

\newtheorem{proposition}{Proposition}

\DeclareMathOperator*{\argmax}{argmax}

\usepackage{multirow}
\usepackage{caption}
\usepackage{subcaption}

\newcommand{\giv}{\,|\,}
\DeclareRobustCommand{\parhead}[1]{\textbf{#1}~}

\DeclareMathOperator*{\argmin}{arg\,min}

\newcommand{\mbx}{\mathbf{x}}

\newcommand\dif{\mathop{}\!\mathrm{d}}

\usepackage{booktabs,arydshln}
\makeatletter
\def\adl@drawiv#1#2#3{%
        \hskip.5\tabcolsep
        \xleaders#3{#2.5\@tempdimb #1{1}#2.5\@tempdimb}%
                #2\z@ plus1fil minus1fil\relax
        \hskip.5\tabcolsep}
\newcommand{\cdashlinelr}[1]{%
  \noalign{\vskip\aboverulesep
           \global\let\@dashdrawstore\adl@draw
           \global\let\adl@draw\adl@drawiv}
  \cdashline{#1}
  \noalign{\global\let\adl@draw\@dashdrawstore
           \vskip\belowrulesep}}
\makeatother

\newcommand{\rd}{\mathrm{d}}

\def\vx{{\bf x}}

\def\rd{\mathrm{d}}
\def\rT{{\top}}

\def\mI{\mathrm{I}}

\def\mI{\mathrm{I}}

\newcommand*\R[0]{\mathbb{R}}

\newcommand*\tr[0]{\mathrm{tr}}
\newcommand*\diag[0]{\mathrm{diag}}

\newcommand{\KL}[2]{\mathrm{KL}\left(#1~\Vert~ #2\right)}

\newcommand*\E[1]{\mathbb{E}\left[#1\right]}
\newcommand*\Ep[2]{\mathbb{E}_{#1}\left[#2\right]}

\newcommand*\lrb[1]{\left[#1\right]}
\newcommand*\lrbb[1]{\left\{#1\right\}}
\newcommand*\lrp[1]{\left(#1\right)}
\newcommand*\lrn[1]{\left\|#1\right\|}
\newcommand*\lrw[1]{\left\langle#1\right\rangle}

\renewcommand*{\P}{\mathbf{P}}
\def\mI{\mathrm{I}}

\def\vx{{\bf x}}

\def\rd{\mathrm{d}}

\def\mI{\mathrm{I}}

\newcommand{\1}{\ensuremath{{\sf (i)}}}
\newcommand{\2}{\ensuremath{{\sf (ii)}}}

\newcommand{\real}{\ensuremath{\mathbb{R}}}

\newcommand{\N}{\mathcal N}

\newcounter{assumption}%
\renewcommand{\theassumption}{\arabic{assumption}}

\newcommand{\Gt}{{G_k^{(t)}}}
\newcommand{\Zt}{{Z^{(t)}}}
\newcommand{\Xt}{{U_k^{(t)}}}
\newcommand{\Xtp}{{U_k^{(t+1)}}}
\newcommand{\matrixS}{{\Omega^{(\infty)}}}
\newcommand{\tP}{{\widetilde{P}}}

\newcommand{\lsi}{\frac{1}{\alpha} \log\frac{L}{\alpha}}
\newcommand{\map}{T}

\newacronym{POMDP}{pomdp}{partially observable {M}arkov decision process}
\newacronym{MDP}{mdp}{{M}arkov decision process}

\newacronym{PNS}{pns}{probability of necessity and sufficiency}
\newacronym{PS}{ps}{probability of sufficiency}
\newacronym{PN}{pn}{probability of necessity}
\newacronym{POC}{poc}{probabilities of causation}
\newacronym{PPCA}{ppca}{probabilistic principal component analysis}

\newacronym{GMM}{gmm}{Gaussian mixture model}

\newacronym{VAE}{vae}{variational autoencoder}
\newacronym{RDR}{rdr}{related disequilibrium regression}

\newacronym{SCM}{scm}{structural causal model}

\newacronym{Causal-Rep}{causal-rep}{Causal Representation}

\newacronym{ELBO}{elbo}{evidence lower bound}
\newacronym{OOD}{ood}{out-of-distribution}

\newacronym{IOSS}{ioss}{independence-of-support score}
\newacronym[
  longplural={factors of variation}
]{FOV}{fov}{factor of variation}

\SetKwInput{KwInput}{Input}             
\SetKwInput{KwReturn}{Return}

\usepackage{tikz}
\usetikzlibrary{bayesnet}
\usetikzlibrary{automata,arrows}
\usetikzlibrary{calc,arrows.meta,positioning}
\usepackage{pgfplots}
\pgfplotsset{compat=newest}
\pgfplotsset{plot coordinates/math parser=false}
\usepgfplotslibrary{statistics}

\pgfdeclarelayer{edgelayer}
\pgfdeclarelayer{nodelayer}
\pgfsetlayers{edgelayer,nodelayer,main}

\definecolor{hexcolor0xbfbfbf}{rgb}{0.749,0.749,0.749}

\tikzset{>=latex}
\tikzstyle{none}   = [inner sep=0pt]
\tikzstyle{line}   = [ thick, -, shorten <=1pt, shorten >=1pt ]
\tikzstyle{arrow}  = [ thick,  ->, shorten <=1pt, shorten >=1pt ]
\tikzstyle{ardash} = [ thick dotted, ->, shorten <=1pt, shorten >=1pt ]

\tikzstyle{empty}=[circle,opacity=0.0,text opacity=1.0,minimum width=4pt,minimum height=4pt]
\tikzstyle{box}=[rectangle,fill=White,draw=Black]
\tikzstyle{filled}=[circle,fill=hexcolor0xbfbfbf,draw=Black]
\tikzstyle{hollow}=[circle,fill=White,draw=Black]
\tikzstyle{param}=[rectangle,fill=Black,draw=Black,inner sep=0pt,minimum width=4pt,minimum height=4pt]
\tikzstyle{paramhollow}=[rectangle,fill=White,draw=Black,inner sep=0pt,minimum
width=4pt,minimum height=4pt]

\usepackage{environ}
\makeatletter
\newsavebox{\measure@tikzpicture}
\NewEnviron{scaletikzpicturetowidth}[1]{%
  \def\tikz@width{#1}%
  \begin{lrbox}{\measure@tikzpicture}%
  \BODY
  \end{lrbox}%
  \pgfmathparse{#1/\wd\measure@tikzpicture}%
  \BODY
}
\makeatother

\usepackage{adjustbox}

\newcommand*\patchAmsMathEnvironmentForLineno[1]{%
  \expandafter\let\csname old#1\expandafter\endcsname\csname
    #1\endcsname
  \expandafter\let\csname oldend#1\expandafter\endcsname\csname
    end#1\endcsname
  \renewenvironment{#1}%
  {\linenomath\csname old#1\endcsname}%
  {\csname oldend#1\endcsname\endlinenomath}%
}
\newcommand*\patchBothAmsMathEnvironmentsForLineno[1]{%
  \patchAmsMathEnvironmentForLineno{#1}%
  \patchAmsMathEnvironmentForLineno{#1*}%
}
\patchBothAmsMathEnvironmentsForLineno{equation}
\patchBothAmsMathEnvironmentsForLineno{align}
\patchBothAmsMathEnvironmentsForLineno{flalign}
\patchBothAmsMathEnvironmentsForLineno{alignat}
\patchBothAmsMathEnvironmentsForLineno{gather}
\patchBothAmsMathEnvironmentsForLineno{multline}

\usepackage[defaultlines=3,all]{nowidow}

\title{\Large{Statistical and Computational Trade-offs in Variational
Inference: \\ A Case Study in Inferential Model Selection}}

\author{
   \normalsize{Kush Bhatia\footnote{Authors are listed in alphabetical order.}}\\
   \normalsize{University of California, Berkeley}\\
  \normalsize{\textit{kushbhatia@berkeley.edu}}
   \and
   \normalsize{Nikki Lijing Kuang$^*$}\\
   \normalsize{University of California, San Diego}\\
   \normalsize{\textit{l1kuang@ucsd.edu}}
   \and
   \makebox[1cm][r]{\normalsize{Yi-An Ma$^*$}}\\
   \makebox[5cm][r]{\normalsize{University of California, San Diego}}\\
   \makebox[2.2cm][r]{\normalsize{\textit{yianma@ucsd.edu}}}
   \and
   \makebox[2.7cm][r]{\normalsize{Yixin Wang$^*$}}\\
   \makebox[4.5cm][r]{\normalsize{University of Michigan}}\\
   \makebox[4cm][r]{\normalsize{\textit{yixinw@umich.edu}}}
  }

\definecolor{salmon}{RGB}{234,153,153}
\definecolor{cornflowerblue}{RGB}{100,149,237}
\hypersetup{
    colorlinks,
    linkcolor={black},
    citecolor={cornflowerblue},
    urlcolor={salmon}
}

\date{}
\begin{document}
\maketitle
  


\begin{abstract} 

Variational inference has recently emerged as a popular alternative to
the classical Markov chain Monte Carlo (MCMC) in large-scale Bayesian
inference. The core idea is to trade
statistical accuracy for computational efficiency. 
In this work, we
study these statistical and computational trade-offs in variational
inference via a case study in inferential model selection. Focusing on
Gaussian inferential models (or variational approximating
families) with diagonal plus low-rank precision matrices, we initiate
a theoretical study of the trade-offs in two aspects, Bayesian
posterior inference error and frequentist uncertainty quantification
error. From the Bayesian posterior inference perspective, we
characterize the error of the variational posterior relative to the
exact posterior. We prove that, given a fixed computation budget, a
lower-rank inferential model produces variational posteriors with a
higher statistical approximation error, but a lower computational
error; it reduces variance in stochastic optimization and, in turn,
accelerates convergence. From the frequentist uncertainty
quantification perspective, we consider the precision matrix of the variational posterior as an uncertainty estimate, which involves an additional statistical error originating from the sampling uncertainty of the data.
As a consequence, for small datasets, the inferential model need not
be full-rank to achieve optimal estimation error (even with unlimited
computation budget). 

\end{abstract}


\clearpage




\section{Introduction}

Modern Bayesian inference relies on scalable algorithms that can
perform posterior inference on large datasets. One such algorithm is
variational inference, which has recently emerged as a popular
alternative to the classical Markov chain Monte Carlo (MCMC)
algorithms~\citep{Saul_VI,blei2017variational}. Unlike MCMC that relies on
sampling, variational inference infers the posterior by solving a
constrained optimization problem, and scales to large datasets by
leveraging modern advances in stochastic optimization~\citep{Hoffman_SVI,Huggins2022}.

The key idea of variational inference is to trade statistical accuracy
for computational efficiency. It aims to \textit{approximate} the
posterior, reducing the computation costs but also potentially compromising its
statistical accuracy. To
perform posterior approximation in variational inference, we solve an
optimization problem. We first choose an inferential model---also
known as a variational approximating family---and then find the member
within this family that is closest to the exact posterior in KL
divergence. Herein, the choice of the inferential model plays a key role
in trading off statistical accuracy and computational efficiency. A
less flexible inferential model incurs a higher statistical
approximation error. Yet, it can make the computation more efficient.

\begin{figure}[t]
    \centering
    \begin{subfigure}[t]{1\linewidth}
        \centering
        \includegraphics[width = 0.7
        \textwidth]{{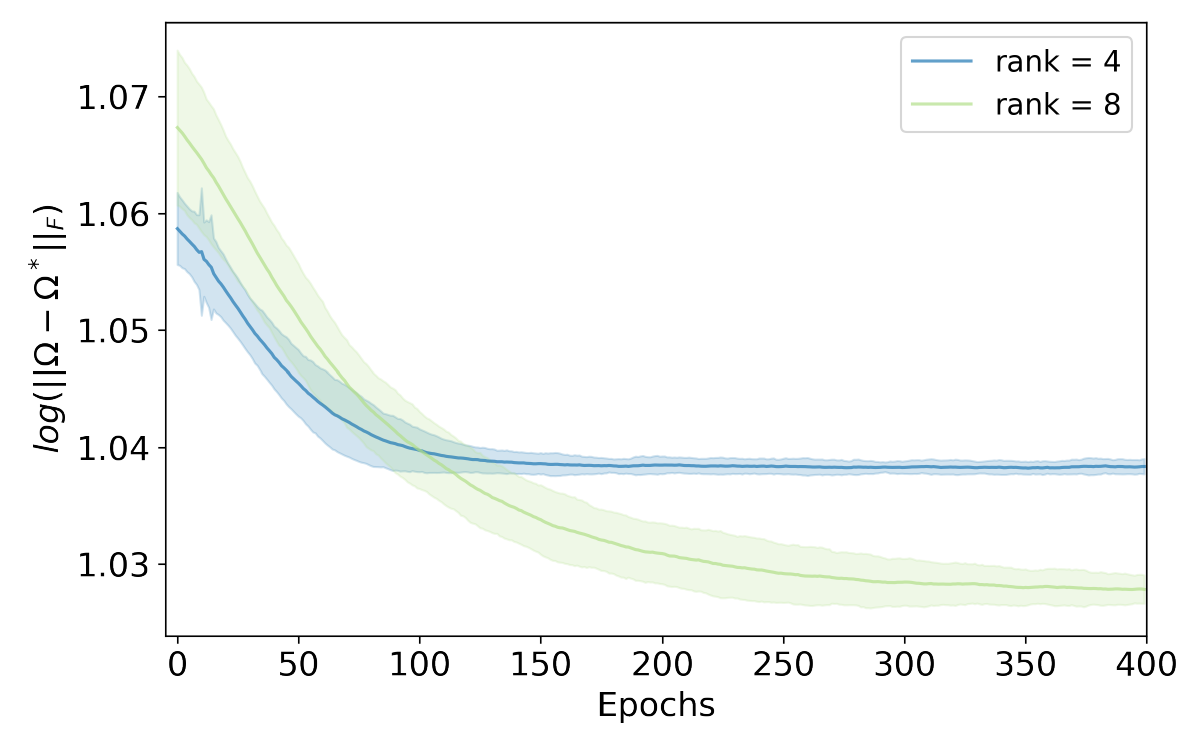}}
    \end{subfigure}
    \caption{The choice of the optimal inferential model varies with the computational budget in a Bayesian logistic regression on the cardiac arrhythmia dataset.
    When the computational budget can only permit training 10 epochs, the less flexible rank-4 Gaussian inferential model (\Cref{eq:VI_family}) achieves a lower posterior approximation error, measured by the Frobenius distance between the precision matrices of the approximate and the exact posterior. With a higher computational budget (e.g. training 100 epochs), the more flexible rank-8 inferential model achieves a lower error. \label{fig:cardiacNormDifference}
    }
    \vspace{-10pt}
\end{figure}

This trade-off between statistical accuracy and computational efficiency bears important practical implications when the computational budget is limited, a setting prevalent in large-scale Bayesian inference. With a limited computational budget, choosing a more flexible inferential model may not lead to a better posterior approximation. While it shall theoretically return a closer approximation to the exact posterior in KL divergence, we may not reach this close approximation in practice due to optimization complications; the flexible inferential model may make the optimization problem so hard that the optimization algorithm can not converge within the limited computational budget, leading to suboptimal solutions to the optimization and hence a poor posterior approximation; see \Cref{fig:cardiacNormDifference} for an example in Bayesian logistic regression.

The statistical and computational trade-off in variational inference suggests that choosing the most flexible inferential models can be suboptimal under a limited computational budget. Then how does statistical accuracy trade-off with computational efficiency? How can we choose the inferential model to achieve the optimal trade-off? We study these questions in this paper.

\subsection{Main ideas}

We describe the setup and the main results of the paper. Consider a
standard statistical inference task on a dataset $\mathbf{x} =
\{x_i\}_{i=1}^n$. We posit a probability model
$p(\mathbf{x}\giv\theta)$ with a $d$-dimensional latent
variable $\theta$ and a prior $p(\theta)$ on
the latent. The goal is to estimate the latent variable
$\theta$ by inferring its posterior
$p(\theta \giv \mathbf{x}) = p(\theta)
\cdot p(\mathbf{x} \giv \theta) / p(\mathbf{x})$. The
posterior is hard to compute because its denominator is the marginal
likelihood of $\mathbf{x}$, which is an integral $p(\mathbf{x})=\int
p(\theta) p(\mathbf{x}\giv \theta) \dif
\theta$ and is often computationally intractable.

Variational inference seeks to \emph{approximate} the posterior
$p(\theta \giv \mathbf{x})$. We first posit an
inferential model $\mathcal{Q}$, also known as a variational
approximating family. A classical choice of $\mathcal{Q}$ is the
mean-field family, which contains all distributions with factorizable
densities~\citep{wainwright2008graphical}. After positing
$\mathcal{Q}$, we then find the member within $\mathcal{Q}$ that is
closest in KL divergence to the exact posterior,
\begin{align}
\label{eq:vb-def}
    q_{\phi^*}(\theta\giv \mathbf{x}) = \argmin_{q_\phi\in \mathcal{Q}}
    \mathrm{KL}\left(q_\phi(\theta \giv \mathbf{x}) \| p(\theta \giv
    \mathbf{x})\right).
\end{align}
The closest member $q_{\phi^*}(\theta\giv \mathbf{x})$,
sometimes called the \emph{variational posterior}, is used in
downstream analysis in the place of the exact posterior.

Variational inference circumvents the computation of the intractable
integral. The reason is that \Cref{eq:vb-def} is equivalent to
maximizing an objective that does not involve the hard-to-compute
integral $p(\mathbf{x})=\int p(\theta) p(\mathbf{x}\giv \theta) \dif
\theta$, that is,
\begin{align}
\label{eq:elbo}
    q_{\phi^*}(\theta\giv \mathbf{x}) = \argmax_{q_\phi\in \mathcal{Q}} \mathbb{E}_{\theta \sim q_\phi(\theta\giv \mathbf{x})}\left[\log p(\theta, \mathbf{x}) - \log q_\phi(\theta\giv \mathbf{x})\right].
\end{align}
The objective in \Cref{eq:elbo} is known as the \emph{evidence lower
bound} (ELBO), whose expectation can be calculated via Monte Carlo
samples of $q_\phi(\theta\giv \mathbf{x})$.

The inferential model $\mathcal{Q}$ drives the statistical accuracy
and computational efficiency of variational inference. The variational
posterior is a biased estimate of the exact posterior when the
inferential model $\mathcal{Q}$ is too restrictive to contain the
exact posterior. Yet, such a variational posterior can be easier to
compute, because searching over a more restricted inferential model is
more computationally efficient. Thus, variational inference trades
statistical accuracy for computational efficiency by restricting the
capacity of its inferential model. Though this trade-off in
variational inference is observed in empirical studies (e.g.
\Cref{fig:cardiacNormDifference}), few theoretical results exist
around the characterization of this trade-off.

Here we theoretically characterize this statistical and computational
trade-off by varying the choice of inferential models. We consider
\emph{low-rank Gaussian inferential models}: a rank-$p$ Gaussian
inferential model is a Gaussian family for variational approximation
whose precision matrix has a structure of a diagonal matrix plus a rank-$p$ matrix,
\begin{align}
    \mathcal{Q}_p = \{\mathcal{N}(\hat{\theta}, \Omega^{-1}):
    \hat{\theta}\in \mathbb{R}^d, \Omega = D + \sum_{i=1}^p \lambda_i
    u_i u_i^\top, \text{$D$ is a $d\times d$ diagonal matrix}, p\leq
    d\},
    \label{eq:VI_family}
\end{align} 
where we denote the parameters as $\phi = (\hat{\theta}, \Omega).$

These low-rank inferential models extend the classical Gaussian
mean-field family: the Gaussian mean-field family is a special case of
$\mathcal{Q}_p$ with $p=0$. Unlike the mean-field family, however,
this low-rank family can capture the dependence structure among latent
variables. As we increase the rank $p$ to be $p=d$, $\mathcal{Q}_p$
becomes the Gaussian full-rank family. The rank of an inferential
model indicates its capacity, so we will consider the low-rank
inferential models with different $p$'s to study the statistical and
computational trade-offs in variational inference.

Using this Gaussian low-rank family of inferential models, we study
the statistical and computational trade-offs of variational inference
in two aspects: Bayesian posterior inference error and frequentist
uncertainty quantification error. Bayesian posterior inference error
characterizes the error of the approximate posterior relative to the
exact posterior in KL divergence.

\parhead{Informal version of \Cref{thm:KL_convergence}.} After running
stochastic variational inference (specifically, the preconditioned SGD
algorithm detailed in \Cref{sec:SGD_VI}) with at most $\Pi$ gradient
evaluations, the resulting variational posterior $q_{\phi^{(T)}}(
\theta\giv
\mathbf{x} ) $ satisfies
\begin{align}
\KL{ q_{\phi^{(T)}}( \theta\giv \mathbf{x} )
}{p(\theta\giv \mathbf{x})} &\leq \underbrace{c_1\cdot \sum_{k=p+1}^d
\lrp{\lambda^*_k}^2}_{\text{Variational Approximation Error}} + \underbrace{c_2 \cdot
\lrp{\frac{pd}{\Pi}}^{1/3} \sum_{k=1}^p \lambda^*_k}_{\text{Optimization Error}} ,
\end{align}
with high probability, where $\lambda^*_k$ is the $k$th largest
eigenvalues of the true posterior precision; $c_1, c_2$ are two
constants; $T$ is the number of iterations corresponding to $\Pi$
gradient evaluations.

Theorem 1 shows that the total error of the variational posterior
(i.e. the KL divergence between the variational and exact posterior)
is composed of two terms, an irreducible statistical approximation
error and a reducible numerical error. Lower-rank approximation of the
posterior suffers from a larger irreducible statistical approximation
error. Yet, its numerical error can be reduced by more computation,
which scales linearly with the rank. Accordingly, a higher-rank
approximation achieves lower statistical error but requires more
computation to reduce the numerical error.

Theorem 1 also suggests an optimal choice of inferential model that can  significantly improve the required computation budget. In particular, given a fixed error tolerance, the optimal inferential model requires a total computational budget that only scales linearly with the dimension $d$ in \Cref{eq:optimal_compute}, as opposed to the quadratic scaling if we use the naive full-rank parametrization~\citep{hoffman2019langevin}. Moreover, the faster the eigenvalue decay of $\lambda_k^*$ is, the less computation the optimal inferential model requires to achieve the error tolerance.  Finally, the computation budget scales with the condition number of the true posterior precision.


We next turn to evaluate variational inference in its resulting
frequentist uncertainty estimate, considering the
covariance matrix of the variational posterior as an uncertainty
estimate.

\parhead{Informal version of \Cref{prop:stat}.} 
After stochastic variational inference with at most $\Pi$ number of
gradient evaluations, the precision matrix $\Omega^{(T)}$ of the
resulting variational posterior satisfies
\begin{align}
    \lrn{ \frac{1}{n} \Omega^{(T)}-\Omega^* }_F^2 \leq  \underbrace{c_3\sum_{k =
    p+1}^d(\lambda_k^*)^2}_{\text{Variational Approximation Error}} +
    \underbrace{c_4 \cdot \lrp{\frac{pd}{\Pi}}^{1/3} \sum_{k=1}^p \lambda^*_k}_{\text{Optimization Error}}
    + \underbrace{\frac{c_5}{n}}_{\text{Statistical Error}},
\end{align}
with high probability, where $\Omega^*$ is the true (rescaled)
posterior precision matrix with infinite i.i.d. data, and $c_3, c_4,
c_5$ are some constants.

Theorem 2 implies that, relative to the true asymptotic covariance matrix, variational inference with lower-rank inferential models
bears lower computational costs (the second term) but suffers from
higher approximation error (the first term). Further, when the sample
size is small, the third term may dominate; thus the optimal
inferential model need not be full-rank to achieve the optimal order of estimation error (even with unlimited computational budget). In particular, under unlimited computational budget, any inferential model is optimal as long as its approximation error is of (at most) the same order as the statistical error; further decreasing the approximation error with more flexible inferential models cannot decrease the order of the total error.

The key observation behind these results is the connection between the
natural gradient descent algorithm commonly used in variational
inference and the stochastic power method, whose convergence rate
depends on the noise of the gradient~\citep{hardt2014}. Specifically,
this connection implies that the convergence rate of variational
inference also depends on the noise of its gradients, which increases
as we increase the flexibility of the inferential model $\mathcal{Q}$.
In more detail, the gradient of the ELBO decomposes into two terms
\begin{multline*}
\nabla_{\phi}\mathbb{E}_{\theta \sim q_\phi(\theta\giv \mathbf{x})}\left[\log p(\theta, \mathbf{x}) - \log q_\phi(\theta\giv \mathbf{x})\right]\\
= -\Ep{\theta \sim q_\phi(\theta\giv
\mathbf{x})}{ \nabla_{\phi}  \log
q_{\phi}(\theta\giv\mathbf{x})\cdot \log{
q_{\phi}(\theta\giv\mathbf{x})}  }
+ \Ep{\theta \sim q_\phi(\theta\giv \mathbf{x})}{ \nabla_{\phi} \log  q_{\phi}(\theta\giv \mathbf{x}) \cdot \log p(\theta, \mathbf{x})  }.
\end{multline*}
The first term can be computed deterministically. The second term,
however, can only be estimated via drawing samples from
$\theta \sim q_\phi(\theta\giv \mathbf{x})$,
and induce noise in gradients; this noise increases as the inferential
model becomes more flexible, i.e. $\mathcal{Q}$ has a higher-rank
precision matrix.


Beyond theoretical understanding, Theorems 1 and 2 can also inform
optimal inferential model selection in practice, which we demonstrate
in \Cref{sec:trade-off-normal}. We also corroborate the theoretical
findings with empirical studies. Across synthetic experiments, we
find that empirical observations confirm the theoretical results:
higher-rank approximation exhibits better statistical accuracy but
takes more stochastic gradient steps to converge.

\parhead{Related work.} This work draws on several threads of previous
research in theoretical characterizations of variational inference and
statistical--computational trade-offs in statistical methods.

Theoretical results around variational inference have mostly centered
around its statistical properties, including asymptotic
properties~\citep{wang2018frequentist,wang2019,cherief2020convergence,chen2020consistency,knoblauch2019frequentist,bhattacharya2021statistical,bhattacharya2020variational,hajargasht2019approximation,han2019statistical,guha2020statistical,banerjee2021pac,cherief2019consistency,jaiswal2020asymptotica,cherief2018consistency,womack2013ldainconsistency,hall2011theory,hall2011asymptotic,celisse2012consistency,bickel2013asymptotic,westling2015establishing,you2014variational,wang2006convergence,zhang2017convergence,wang2004convergence,pati2017statistical,yang2017alpha,alquier2017concentration,alquier2016properties,campbell2019},
finite sample approximation
error~\citep{huggins2018practical,huggins2020validated,giordano2017covariances,sheth2017excess,chen2017use},
robustness to model
misspecification~\citep{medina2021robustness,cherief2018consistency,wang2019,alquier2017concentration},
and properties in high-dimensional
settings~\citep{ray2021variational,yang2020variational,ray2020spike,mukherjee2021variational}.
Beyond statistical properties, a few works have explored the role of
optimization algorithms in variational
inference~\citep{mukherjee2018mean,sarkar2021random,plummer2020dynamics,zhang2017theoretical,ghorbani2018instability,pmlr-v119-hoffman20a,xu2021computational}. Along this line, closest to our work is \citet{xu2021computational}, which investigates the asymptotic properties of the nonconvex optimization problem in Gaussian variational inference. Our work differs from these works: we focus on explicitly characterizing the statistical and computational trade-offs in variational inference, enabling us to identify the optimal choice of inferential model that strikes the balance.

Despite being rarely explored in variational inference, statistical
and computational trade-offs have been formalized in other statistical
problems, including principal component
analysis~\citep{wang2016statistical,sriperumbudur2017approximate},
clustering~\citep{calandriello2018statistical,balakrishnan2011statistical},
matrix completion~\citep{dasarathy2017computational}, denoising
problems~\citep{chandrasekaran2013computational}, high-dimensional
problems~\citep{berthet2014statistical}, single-index
models~\citep{wang2019statistical}, two-sample
testing~\citep{ramdas2015adaptivity}, weakly supervised
learning~\citep{yi2019more}, combinatorial
problems~\citep{lu2018edge,khetan2016computational,jin2020computational,khetan2018generalized},
and other estimators like stochastic composite
likelihood~\citep{dillon2009statistical} and Dantzig-type
estimators~\citep{listatistical}. These analyses focus on studying
point estimators for the parameters of interest. In contrast, we
consider distributional estimators in this work. We characterize the
statistical and computational trade-offs for variational posterior distributions.

Finally, this work relates to the body of literature on Bayesian model
selection; see~\citet{o2009review} for a review. Closely related works
include~\citet{chandrasekaran2010latent} which focuses on
identifiability and tractability of latent variable model
selection, \citet{yang2016computational} which analyzes an MCMC
approach to Bayesian variable selection based on designing priors, and
\citet{song2020extended} which proposes an extended stochastic
gradient MCMC for Bayesian variable selection. These works analyze
different aspects of Bayesian model selection algorithms than we
do; they focus on identifiability, asymptotic convergence, or
computational complexity, while we analyze non-asymptotic convergence
and statistical guarantees.


\parhead{This paper.} The rest of the paper is organized as follows.
\Cref{sec:SGD_VI} introduces the optimization algorithms we analyze
for stochastic variational inference.
In \Cref{subsec:sgd-analysis}, we analyze the
convergence of stochastic variational inference for Gaussian
posteriors, via transforming it into an equivalent algorithm that
performs the stochastic power method. In \Cref{sec:trade-off-normal},
we study the statistical and computational trade-offs in stochastic
variational inference. We first focus on the Bayesian posterior
inference error, measured by KL divergence between the variational
approximation and the true posterior. We then examine the frequentist
uncertainty quantification error, where the precision matrix of the
inferential model is used to quantify the sampling uncertainty. In
both cases, we demonstrate how the theoretical results inform optimal
inferential model selection: they identify the optimal rank $p$ of the
low-rank inferential model so that they achieve a given error
threshold with minimum total computation. In
\Cref{sec:general_posterior}, we extend the convergence analysis of
\Cref{subsec:sgd-analysis} beyond Gaussian posteriors and analyze the
inference error due to the low-rank inferential model. We conclude with empirical studies in
\Cref{sec:empirical}.

\section{Stochastic Optimization for Variational Inference}
\label{sec:SGD_VI}


To study the statistical and computational trade-off in variational inference, we first study the stochastic optimization algorithms, focusing on their convergence behaviors in variational inference with low-rank Gaussian inferential models. In the next section, we will discuss the implications of these convergence behaviors on the statistical and computational trade-off in variational inference.

\subsection{Stochastic variational inference (SVI) algorithm}

Recall that our goal is to find a variational approximation $q(\theta|\mathbf{x})$ to the posterior within an inferential model, a.k.a. a variational approximating family. We focus on the family of rank-$p$ Gaussian inferential model $\mathcal{Q}_p$; all members of $\mathcal{Q}_p$ are multivariate Gaussian $\mathcal{N}(\hat{\theta}, \Omega^{-1})$ with a diagonal plus rank-$p$ precision matrix:
$\Omega = D + \sum_{i=1}^p \lambda_i u_i u_i^\rT = D + U \Lambda U^\rT$,
where $U\in\R^{d\times p}$ is a semi-orthonormal matrix  and its associated eigenvalues are $\Lambda\in\R^{p\times p}$.
For simplicity of exposition, we assume that this Gaussian inferential model is centered with $\hat{\theta}=0$ and the diagonal structure of its precision matrix is contributed by an isotropic normal prior, i.e. $D=\alpha \mI_d$. The discussion easily extends to non-centered $\hat{\theta}$ and anisotropic $D$.



To find the closest variational approximation in $\mathcal{Q}_p$, we solve an optimization problem that minimize $\KL{q(\theta\,|\,\mathbf{x})}{p(\theta\,|\,\mathbf{x})}$---often abbreviated to $\KL{q}{p}$---subject to the constraint that $q(\theta\,|\,\mathbf{x}) \in \mathcal{Q}_p.$ The most common approach to this optimization is to perform gradient descent over the KL divergence to find the optimal parameters for the variational approximation. Specifically, we write $q_{(U,\Lambda)}$ as the variational approximation with parameters $U$ and $\Lambda$, and perform  gradient descent over $\KL{q}{p}$ with respect to the parameters $\phi\in\{u_i,\lambda_i\}_{i=1}^p$,
\begin{align}
\nabla_{\phi} \KL{q}{p}
&= \Ep{\theta\sim q}{ \nabla_{\phi} \log q_{(U,\Lambda)}(\theta|\mathbf{x}) \log\frac{q_{(U,\Lambda)}(\theta|\mathbf{x})}{p(\theta|\mathbf{x})}  }.
\end{align}


The gradient $\nabla_{\phi} \KL{q}{p}$ can be further decomposed into two terms,
\begin{align}
\nabla_{\phi} \KL{q}{p}
= \Ep{\theta\sim q}{ \nabla_{\phi}  \log q_{(U,\Lambda)}(\theta|\mathbf{x}) \cdot \log{q_{(U,\Lambda)}(\theta|\mathbf{x})}  }
+ \Ep{\theta\sim q}{ \nabla_{\phi} \log q_{(U,\Lambda)}(\theta|\mathbf{x}) \cdot \psi(\theta)  },\label{eq:T1T2_decomp}
\end{align}
where we denote $\psi(\theta)$ as the unnormalized component of the true posterior, $p(\theta|\mathbf{x}) \propto\exp\lrp{-\psi(\theta)}$ with $\psi(\theta)$ being a positive definite function. (\Cref{eq:T1T2_decomp} is due to integration by parts, which implies that $\Ep{\theta\sim q}{ \nabla_{\phi} \log q_{(U,\Lambda)}(\theta|\mathbf{x}) \cdot C } = 0$ for any constant $C$.) The decomposition in \Cref{eq:T1T2_decomp} implies that its first term has a closed form and can be explicitly computed for $q\in \mathcal{Q}_p$ from the low-rank Gaussian inferential model. The second term, in contrast, does not always have a closed form. As it is an expectation over $q_{(U,\Lambda)}(\ \cdot\ |\mathbf{x})$, we often adopt a Monte Carlo estimate 
\[
\Ep{\theta\sim q}{ \nabla_{\phi} \log q_{(U,\Lambda)}(\theta|\mathbf{x}) \cdot \psi(\theta)  }\approx \frac{1}{N} \sum_{j=1}^N  \nabla_{\phi} \log q_{(U,\Lambda)}(\theta_j|\mathbf{x}) \cdot \psi(\theta_j),
\]
where $\theta_j$ are i.i.d. samples from $q_{(U,\Lambda)}(\ \cdot\ |\mathbf{x})$; the variance of this Monte Carlo estimate decreases with larger $N$.


\sloppy
In practice, we often perform gradient descent in two stages. We first
learn the semi-orthonormal matrix $U=[u_1, \ldots, u_p]$ and then
proceed to learn the corresponding eigenvalues
$\Lambda=\mathrm{diag}(\lambda_1, \ldots, \lambda_p)$. Below we detail the stochastic optimization
algorithm for $U$ and $\Lambda$ respectively.




\parhead{Optimizing for the eigenvectors $U$.} In the first stage of the algorithm, we focus on optimizing the semi-orthonormal matrix $U$, fixing the other parameters $\Lambda$ at their initial values, e.g.~$\Lambda^{(0)}~=~\mI_p$.

To optimize $U$, we perform preconditioned gradient descent of KL divergence over $U$ \citep{Preconditioning}. We employ the preconditioner $\mI\otimes\Omega$ such that the preconditioned gradient of the KL divergence has a simple form,
\begin{align}
\Omega \nabla_{U} \KL{ q}{p} 
&= U\Lambda + \Ep{\theta\sim q}{ \lrp{- \Omega \theta \theta^\rT U \Lambda + U\Lambda} \psi(\theta)  } \label{eq:precond-step-1}\\
&= U\Lambda - \Ep{\theta\sim q}{ \nabla\psi(\theta) \theta^\rT } U \Lambda,
\label{eq:precond_gradient}
\end{align}
where \Cref{eq:precond-step-1} is obtained by calculating $\nabla_{U} \log q_{(U,\Lambda)}(\theta|\mathbf{x})
= - \theta \theta^\rT U \Lambda + \Omega^{-1} U \Lambda$ and plugging it into \Cref{eq:T1T2_decomp}. Thus, given the eigenvalues $\Lambda$, the preconditioned stochastic gradient descent (SGD) over $U$ operates as follows. At the $t$th iteration,
\begin{align}
& \text{Sample i.i.d.} \; \theta_j^{(t)} \sim q_{\lrp{U^{(t)},\Lambda}}, j\in\{1,\dots,N\} \label{eq:precond_GD_1},\\
& \tilde{U}^{(t+1)} = U^{(t)} - h_t U^{(t)} \Lambda + \frac{h_t}{N} \sum_{j=1}^N \lrp{ \nabla\psi\lrp{\theta_j^{(t)}} \theta_j^\rT } U^{(t)} \Lambda, \label{eq:precond_GD_2}\\
& U^{(t+1)} = \mathrm{QR}\lrp{\tilde{U}^{(t+1)}}, \label{eq:precond_GD_3}
\end{align}
where $\Omega^{(t)} = D + U^{(t)} \Lambda \lrp{U^{(t)}}^\rT$. \Cref{eq:precond_GD_3} employs 
a QR decomposition to ensure that the matrix $U^{(t)}$ is a semi-orthonormal matrix at all iterations.
We choose a simple
initialization: $U^{(0)} = [e_1,\dots,e_p]$, where $e_i \in \R^d$ is a column vector with $1$ in the $i$-th entry and $0$'s elsewhere.
\begin{algorithm}[t]
\SetAlgoLined
\setstretch{0.5}
\KwInput{$D = \alpha\mI$, $U^{(0)} = \left[e_1,\dots,e_p\right]$, $\Lambda^{(0)}=\mI_p$}
\For{$t = 0, \ldots,T-1$}{
\begin{flalign}
& \text{Sample i.i.d.} \; \theta_j^{(t)} \sim q_{\lrp{U^{(t)},\Lambda^{(t)}}}, \forall j\in\{1,\dots,N\}, &&\nonumber\\
& \tilde{U}^{(t+1)} = U^{(t)} - h_t U^{(t)} \Lambda^{(t)} + \frac{h_t}{N} \sum_{j=1}^N \lrp{ \nabla\psi\lrp{\theta_j^{(t)}} \theta_j^\rT } U^{(t)} \Lambda^{(t)},&&\nonumber\\
& U^{(t+1)} = \mathrm{QR}\lrp{\tilde{U}^{(t+1)}}. &&\nonumber
\end{flalign}
Sample i.i.d. $\theta_j^{(t)} \sim q_{\lrp{U^{(t)},\Lambda^{(t)}}}, \forall j\in\{1,\dots,M\}$.\\
\For{$k=1,\dots,p$}{
\begin{flalign}
& \lambda^{(t)}_k = \frac{1}{M} \sum_{j=1}^M \lrp{ u_k^{(t)} }^\rT \lrp{ \frac{\nabla \psi(\theta_j+\Delta \cdot u_k^{(t)}) - \nabla \psi(\theta_j-\Delta \cdot u_k^{(t)})}{2\Delta} } - D_{k,k} .&& \label{eq:lambda_step}
\end{flalign}
}
}
\KwReturn{$U^{(T)}, \Lambda^{(T)} = \mathrm{diag}(\lambda^{(T)}_1, \ldots, \lambda^{(T)}_p)$}
\caption{Stochastic variational inference (SVI) with low-rank Gaussian inferential models}
\label{alg:General}
\end{algorithm}

\parhead{Optimizing for the eigenvalues $\Lambda$.} In the second stage of the algorithm, we learn the diagonal eigenvalue matrix $\Lambda$ given the estimated semi-orthonormal matrix $U$. We find the optimal $\Lambda$ by finding the values of $\lambda_i$ achieving a zero derivative of the log-likelihood over $\lambda_i$. In the following lemma, we obtain  an explicit expression for solving $\lambda_i$, given vector $u_i$ in the semi-orthonormal matrix $U$.
\begin{lemma}
\label{fact:lambda}
The solution to $\nabla_{\lambda_i} \KL{q_{(U,\Lambda)}(\theta|\mathbf{x})}{p(\theta|\mathbf{x})} = 0$ is $\lambda_i = \Ep{\theta\sim q}{ u_i^\rT \nabla_{\theta}^2 \psi(\theta) u_i } - D_{i,i}$.
\end{lemma}
This fact implies that solving for optimal $\lambda_i$ is equivalent to evaluating $\Ep{\theta\sim q}{ u_i^\rT \nabla_{\theta}^2 \psi(\theta) u_i }$, which we use $M$ samples of $\theta\sim q$ to estimate. We then note that the Hessian vector product $\nabla_{\theta}^2 \psi(\theta) u_i$ can be accurately estimated by the gradient difference: $\lrp{\nabla \psi(\theta+\Delta \cdot u_i) - \nabla \psi(\theta-\Delta \cdot u_i)}/\lrp{2\Delta}$.
Combining these two facts yields the update rule of \Cref{eq:lambda_step} in \Cref{alg:General}.



Taken together the two stages of optimization, \Cref{alg:General} summarizes the full stochastic optimization algorithm for variational inference with low-rank Gaussian inferential models.

\section{Convergence analysis of SVI for Gaussian posteriors}

\label{subsec:sgd-analysis}

We next analyze the non-asymptotic convergence properties of
the SVI algorithm in \Cref{alg:General}. 
This analysis will characterize the
computational properties of variational inference and facilitate the
characterization of statistical and computational trade-offs in
\Cref{sec:trade-off-normal}. As in \Cref{sec:SGD_VI}, we focus on the
Gaussian posterior cases in \Cref{subsec:sgd-analysis} and
\Cref{sec:trade-off-normal}, where we can express the posterior as
$p(\theta|\vx) \propto \exp\lrp{-\psi(\theta)}$, where $\psi(\theta) =
\frac{1}{2} \theta^\rT \Omega^{(\infty)} \theta$ for a positive
definite matrix $\Omega^{(\infty)} \in \R^{d \times d}$. (We extend to
non-Gaussian cases in \Cref{sec:general_posterior}.)

To analyze \Cref{alg:General} for the Gaussian posterior
(corresponding to a quadratic function $\psi$), we rely on a key
observation that the learning of $U$ and $\Lambda$ decouples: the
optimization algorithm does not require we alternate between
optimizing $U$ and $\Lambda$; rather, it can be separated into two
separate stages because the optimization of $U$ does not depend on the
value of $\Lambda$ in variational inference with low-rank Gaussian
inferential models. In particular, the update in
\Cref{eq:precond_GD_2}, together with the QR decomposition in
\Cref{eq:precond_GD_3}, implies that the stationary solution to the
semi-orthonormal $U$ is always composed of the first $p$ eigenvectors
of $\Omega^{(\infty)}$, regardless of the choice of $\Lambda$ in the
update. From \Cref{fact:lambda}, we also know that, if the eigenspace
$U$ is learned, then the solution $\Lambda$ simply reads out the
eigenvalues of $\Ep{\theta\sim q}{\nabla^2 \psi(\theta)}$, which is
$\Omega^{(\infty)}$ in the Gaussian posterior case. Therefore, the
optimization of $U$ and $\Lambda$ can be performed in two stages by
first fixing $\Lambda=\mI_p$ and letting $U$ converge first and then
computing $\Lambda$ conditioning on $U$.

These observations enables us to consider \Cref{alg:Gaussian}
(SVI\_Gauss)---a simplified version of \Cref{alg:General} tailored to
Gaussian posteriors---to facilitate the analysis for Gaussian
posteriors in \Cref{sec:trade-off-normal}. (The details of \Cref{alg:Gaussian} is in \Cref{sec:alg_details}.) Moreover, as we will see,
the update of $U$ in \Cref{alg:Gaussian} (SVI\_Gauss) closely connects
to the stochastic power method \citep{hardt2014}; this connection will
allow us to borrow analysis tools of the stochastic power method for
analyzing stochastic optimization for variational inference. Below we
discuss these observations and connections in detail.

\parhead{SVI and the stochastic power method.} To establish the connection between the stochastic power method and the 
SVI algorithm, 
we prove that the first stage of \Cref{alg:General}---which performs preconditioned SGD to optimize for the eigenvectors $U$---is equivalent to the stochastic power method for principal component analysis (PCA).


\begin{lemma}[SVI $\Leftrightarrow$ Stochastic power method]
\label{fact:precond_equiv}
Suppose the diagonal matrix $D=\alpha\mI_d$ and fix $\Lambda=\mI_p$ in the update of the matrix $U^{(t)}$ in \Cref{alg:General} (\Cref{eq:precond_GD_1}--\Cref{eq:precond_GD_3}).
The preconditioned SGD in \Cref{eq:precond_GD_1,eq:precond_GD_2,eq:precond_GD_3} with step size $h_t = 1$ is equivalent to the stochastic power method:
\begin{align}
& \text{Sample i.i.d.} \; \theta_j^{(t)} \sim \mathcal{N} \lrp{ 0, (\Omega^{(t)})^{-1} }, \forall j\in\{1,\dots,N\}, \label{eq:sample_Gauss} \\
& \tilde{U}^{(t+1)} = \sum_{j=1}^N \lrp{ \nabla\psi\lrp{\theta_j^{(t)}} \lrp{\theta_j^{(t)}}^\rT \Omega^{(t)} } U^{(t)},  \label{eq:mult_Gauss} \\
& U^{(t+1)} = \mathrm{QR}\lrp{\tilde{U}^{(t+1)}},  \label{eq:QR_Gauss}
\end{align}
where $\Omega^{(t)} = D + U^{(t)} \Lambda \lrp{U^{(t)}}^{\rT} = \alpha\mI_d + U^{(t)} \lrp{U^{(t)}}^\rT$.

In addition, for the Gaussian posterior, \Cref{alg:Gaussian} (which substitutes $\Omega^{(t)}$ in \Cref{eq:sample_Gauss} and \Cref{eq:mult_Gauss} with a fixed input matrix $\Omega$) yields the same stationary solution $U$ as that of the above update, for any positive definite input $\Omega$.
\end{lemma}

\Cref{fact:precond_equiv} is an immediate consequence of equating the updates (\Cref{eq:precond_GD_1,eq:precond_GD_2,eq:precond_GD_3} and \Cref{eq:sample_Gauss,eq:mult_Gauss,eq:QR_Gauss}) in both algorithms. 
%
%
Hence for Gaussian posteriors, we can directly analyze \Cref{alg:Gaussian} for the convergence of \Cref{alg:General}.

\parhead{Convergence analysis of SVI.} \Cref{fact:precond_equiv} established the equivalence between the first stage of \Cref{alg:General} and the stochastic power method. 
Existing works on the stochastic power method discover that, even though the objective is nonconvex, with small enough noise during each iteration, the algorithm will converge and approximately recover the top $p$ eigenspace of $\Omega^{(\infty)}$. Leveraging this connection, below we study the convergence properties of \Cref{alg:General} for each of its two stages and establish per-iteration noise bounds to determine number of samples to generate in the sampling step (\Cref{eq:sample_step_Gauss}) of \Cref{alg:Gaussian}. 

We first establish the convergence of the first stage of \Cref{alg:General} that optimizes over the $d\times p$ semi-orthonormal matrix $U$.
For the Gaussian posterior $p(\theta|\vx) \propto \exp\lrp{-\psi(\theta)}$, if function $\psi=\frac{1}{2}\theta^\rT \Omega^{(\infty)} \theta$ is $\alpha$-strongly convex, then we can express $\Omega^{(\infty)} = U^{(\infty)} \Lambda^{(\infty)} \lrp{U^{(\infty)}}^\rT + \alpha\mI$, for unitary matrix $U^{(\infty)} \in \R^{d\times d}$ and positive semi-definite diagonal matrix $\Lambda^{(\infty)} = \mathrm{diag}\lrp{\lambda^{(\infty)}_1,\dots,\lambda^{(\infty)}_d} \in \R^{d \times d}$; $\alpha$ is the minimum eigenvalue of $\Omega^{(\infty)}$.
For the general non-Gaussian posteriors, we will let $\alpha = \inf_{\theta\in\R^d} \min_{i=1,\dots,d} \sigma_i(\nabla^2 \psi(\theta))$ to be the minimum eigenvalue of the Hessian of the negative log posterior $\nabla^2 \psi(\theta)$ across the space $\R^d$ (See \Cref{sec:general_posterior}).

In what follows, we will measure the accuracy of the semi-orthonormal matrix $U=[u_1,\dots,u_p]$ via the Rayleigh quotient: $u_k^\rT \Omega^{(\infty)} u_k$ for all $k\in\lrbb{1,\dots,p}$.
A uniformly large (and close to $\lambda_k^{(\infty)} + \alpha$) Rayleigh quotient means that matrix $U$ is close to covering the top $p$ eigenspace of $\Omega^{(\infty)}$.
Denoting $U^{(\infty)}_{p} = \lrb{u^{(\infty)}_1,\dots,u^{(\infty)}_p}$ which contains the top $p$ eigenvectors of $U^{(\infty)}$, we have the following result.
\begin{proposition}[Convergence of SVI over $U$]
\label{lem:Oja_convergence}
Assume that $\psi(\theta) = \frac{1}{2} \theta^\rT \Omega^{(\infty)} \theta$ (so that the posterior $p(\theta|\vx)$ is Gaussian) and is $\alpha$-strongly convex and $L$-Lipschitz smooth; the matrix $\Omega^{(\infty)}$ has condition number $\kappa(\Omega^{(\infty)}) = L/\alpha.$
We run the SVI algorithm (described in in equations~\eqref{eq:sample_Gauss} to~\eqref{eq:QR_Gauss} and in \Cref{alg:Gaussian}) for the Gaussian posterior with input matrix $\Omega$.
Let $m_0 = \sigma_{\min}\lrp{ (U^{(\infty)}_p)^\rT U^{(0)} }$ denote the initial condition.
For any $\epsilon, \delta>0$, with 
number of stochastic gradient samples per iteration $N = \widetilde{\boldsymbol{\Omega}}\lrp{ \max\lrbb{ \frac{pd}{ m_0^2 } , \frac{p^2}{ m_0^4 } } \cdot \frac{L^2}{\alpha^2} \cdot \frac{L^2}{\epsilon^2} \cdot \frac{\kappa(\Omega)}{\sqrt{\delta} } }$, 
and number of iterations $T = \boldsymbol{\Theta}\lrp{ \frac{L}{\epsilon} \log\frac{1}{m_0}}$,
we obtain that 
\begin{align}
\label{eq:conv-of-u}
\lrp{u^{(T)}_k}^\rT \Omega^{(\infty)} u^{(T)}_k \geq \lambda^{(\infty)}_k + \alpha - 2\epsilon,
\qquad\forall k\in\{1,\dots,p\}
\end{align}
with probability $1-\delta$, where $\Omega^{(\infty)} = U^{(\infty)} \Lambda^{(\infty)} \lrp{U^{(\infty)}}^\rT + \alpha\mI$ with minimum eigenvalue $\alpha$.
\end{proposition}

The proof of \Cref{lem:Oja_convergence} is in \Cref{sec:pf_U}.
The key proof idea is that we first obtain high probability bound of the distance between the gradient estimate $\frac{1}{N} \sum_{j=1}^N \lrp{ \nabla\psi\lrp{\theta_j^{(t)}} \theta_j^\rT } U^{(t)}$ and the true gradient $\Ep{\theta\sim q}{ \nabla\psi(\theta) \theta^\rT } U$.
This helps us decide the number of samples $N$ to generate in the first step of the algorithm. 
We then recursively bound how much does the top $k$ eigenspace of the $t$-th iterate: $\lrb{ u_1^{(t)},\dots,u_k^{(t)} }$ overlap with the bottom eigenspace of $\Omega^{(\infty)}$ in \Cref{proposition:computation_main}, adopting similar techniques as in the analysis of stochastic power method~\citep{hardt2014, balcan2016improved}.
Summing up the inner products between $u_k^{(T)}$ and the eigenvectors of $\Omega^{(\infty)}$ yields the result. (The additive term of $\alpha$ on the right side of \Cref{eq:conv-of-u} is due to $\alpha$ being the minimum eigenvalue of $\Omega^{(\infty)}$.

\Cref{lem:Oja_convergence} provides the scaling of the total computation complexity of the preconditioned SGD over the semi-orthonormal matrix $U$.
Let $\bar{U}^{(\infty)}_{p\times p} \in \R^{p\times p}$ denote the $p\times p$ leading principal submatrix of $U^{(\infty)}$.
Then we have $m_0 = \sigma_{\min}\lrp{\bar{U}^{(\infty)}_{p\times p}}$ according to our choice of initialization.
Assuming that $\sigma_{\min}\lrp{\bar{U}^{(\infty)}_{p\times p}} > 0$, we have $N=\widetilde{\boldsymbol{\Omega}}\lrp{ p d \cdot \frac{L^2}{\alpha^2} \cdot \frac{L^2}{\epsilon^2} \cdot \frac{\kappa(\Omega)}{\sqrt{\delta} } }$ 
and $T = \widetilde{\boldsymbol{\Theta}}\lrp{ \frac{L}{\epsilon} }$.
Taking the input matrix $\Omega=\mI$, the total computation complexity of the optimization step over $U$ is $N \cdot T = \widetilde{\boldsymbol{\Omega}}\lrp{pd\cdot \frac{L^2}{\alpha^2}\cdot\frac{L^3}{\epsilon^3\sqrt{\delta}}}$.



We next analyze the second stage of the SVI algorithm which solves for $\Lambda$. We quantify the error of the resulting $\Lambda$, assuming that \Cref{lem:Oja_convergence} stands and that the first stage recovers a good representation for the top $p$ eigenspace in $U$.
\begin{proposition}[Convergence of SVI over $\Lambda$]
\label{lem:lambda}
Assume the conditions for \Cref{lem:Oja_convergence} are satisfied such that \Cref{eq:conv-of-u} holds.
Then taking the number of samples for eigenvalue computation $M \geq 2 p \log(4p/\delta) \cdot \frac{L^2}{\epsilon^2}$ and $\varDelta \leq \epsilon/\lrp{ \sqrt{p} \cdot L_{\rm{Hess}} }$ in \Cref{alg:General} \Cref{eq:lambda_step} (or \Cref{alg:Gaussian} \Cref{eq:lambda_step_Gauss}), we obtain that for Gaussian posterior,
\[
\sum_{k=1}^p \lrp{\lambda^{(\infty)}_k - \lambda^{(T)}_k}^2 
\leq 2 \sum_{k=p+1}^d \lrp{\lambda^{(\infty)}_k}^2 + 8 \epsilon \sum_{k=1}^p \lambda^{(\infty)}_k + 8 \epsilon^2,
\]
with probability $1-\delta$.
\end{proposition}

The proof of \Cref{lem:lambda} is in \Cref{sec:pf_Lambda}. Similar to the proof of \Cref{lem:Oja_convergence}, we first bound the stochastic gradient noise in estimating $\Ep{\theta\sim q}{ u_i^\rT \nabla_{\theta}^2 \psi(\theta) u_i }$, which according to \Cref{lem:lambda} achieves the critical point of $\KL{q_{(U,\Lambda)}(\theta|\mathbf{x})}{p(\theta|\mathbf{x})}$ if $u_i$ is the $i$-th eigenvector of $\Omega^{(\infty)}$. We then note that the gradient difference in \Cref{alg:Gaussian} gives an accurate estimate of the Hessian vector product $\nabla_{\theta}^2 \psi(\theta) u_i$. Combining with \Cref{lem:Oja_convergence} that matrix $U$ approximately recovers the top $p$ eigenspace of $\Omega^{(\infty)}$, we obtain the result.

\Cref{lem:Oja_convergence,lem:lambda} imply that the total computation complexity for the SVI algorithm to achieve $\boldsymbol{\mathcal{O}}(\epsilon)$ convergence (plus bias) is
$ N \cdot T + 2 M = \widetilde{\boldsymbol{\Omega}}\lrp{pd\cdot \frac{L^2}{\alpha^2}\cdot\frac{L^3}{\epsilon^3\sqrt{\delta}}}$. 
Moreover, they imply that the bound on $M$ is less than that on $N$ and that the overall complexity is dominated by the computation complexity of the preconditioned SGD over $U$.
The two propositions will help us characterize how far the variational approximation is from the true posterior given fixed computational cost.

\section{Statistical and Computational Trade-offs in Variational Inference}

\label{sec:trade-off-normal}

Leveraging the analysis of the SVI algorithm in \Cref{subsec:sgd-analysis}, we characterize the non-asymptotic error of the variational posterior $q_{\lrp{U^{(T)},\Lambda^{(T)}}}(\theta \giv \mathbf{x})$ after $T$ steps of gradient descent. This non-asymptotic error will come from two sources: the variational approximation error due to the low-rank inferential model, and the computational error due to the stochastic optimization algorithm. How the two sources contribute to the total error of $q_{\lrp{U^{(T)},\Lambda^{(T)}}}(\theta \giv \mathbf{x})$ thus depicts the statistical and computational trade-offs in variational inference.

Specifically, we evaluate the variational posterior at step $T$, $q_{\lrp{U^{(T)},\Lambda^{(T)}}}(\theta \giv \mathbf{x})$, from two aspects: Bayesian posterior inference and frequentist uncertainty quantification. For simplicity of exposition, we adopt the setting where the exact posterior is a  centered Gaussian, i.e.
$p(\theta \giv \mathbf{x}) = \mathcal{N}( 0, (\Omega^{(\infty)} )^{-1} )$ for some $\Omega^{(\infty)}$. Locating the mean parameter requires much less computation; hence this simplication does not alter the analysis.
The Gaussianity of the exact posterior simplifies the SVI algorithm to \Cref{alg:Gaussian}. (We extend to general non-Gaussian posteriors in \Cref{sec:general_posterior}.)



\subsection{Bayesian posterior inference error}

\label{subsec:bayes}

We first study the Bayesian posterior inference error of variational posteriors, where we evaluate its KL divergence to the true Bayesian posterior, $\KL{q_{\lrp{U^{(T)},\Lambda^{(T)}}}(\theta \giv \mathbf{x})}{p(\theta \giv \mathbf{x}))}$.


Combining the convergence analysis of both stages of \Cref{alg:General} in \Cref{subsec:sgd-analysis}, the following theorem establishes the computation-approximation trade-off. To focus on how the error depends on the rank $p$ of the inferential model, below we adopt the initialization of $U^{(0)} = \left[e_1,\dots,e_p\right]$ and that $m_0 = \sigma_{\min}\lrp{\bar{U}^{(\infty)}_{p\times p}} > 0$.

\begin{theorem}
\label{thm:KL_convergence}
Suppose that we are given a computation budget that allows for $\Pi=N\cdot T + 2M$ gradient evaluations in \Cref{alg:General} with input matrix $\Omega=\mI$.
Then under the allocation rule of $N = \boldsymbol{\Theta}\lrp{ \Pi^{2/3} \cdot \frac{(pd)^{1/3}}{\delta^{1/6} } \cdot \lrp{\frac{L}{\alpha}}^{2/3} }$, $T = \boldsymbol{\Theta}\lrp{ \Pi^{1/3} \cdot \frac{\delta^{1/6}}{(pd)^{1/3}} \cdot \lrp{\frac{\alpha}{L}}^{2/3} }$, and $M = \widetilde{\boldsymbol{\Theta}} \lrp{ \Pi^{2/3} \cdot \frac{p^{1/3}}{d^{2/3}} \cdot \lrp{\frac{\alpha}{L}}^{4/3} \delta^{1/3} }$ (also laid out in \Cref{lem:Oja_convergence} and \Cref{lem:lambda}), the variational posterior $q_{\lrp{U^{(T)},\Lambda^{(T)}}}(\theta \giv \mathbf{x})$ satisfies
\begin{align}
\KL{q_{\lrp{U^{(T)},\Lambda^{(T)}}}(\theta \giv \mathbf{x})}{p(\theta \giv \vx)}
&\lesssim \underbrace{ \frac{1}{\alpha^2} \sum_{k=p+1}^d \lrp{\lambda^{(\infty)}_k}^2}_{E_1: \text{ Approximation}} 
+ \underbrace{ \lrp{\frac{pd}{\Pi}}^{1/3} \cdot \lrp{\frac{L}{\alpha}}^{5/3} \cdot \frac{1}{\delta^{1/6}} \cdot \frac{1}{\alpha} \sum_{k=1}^p \lambda^{(\infty)}_k }_{E_2: \text{ Optimization}},
\end{align}
with $(1-\delta)$ probability, where we have the exact posterior satisfying $p(\theta\giv\mathbf{x})\propto\exp\lrp{-\theta^\rT\Omega^{(\infty)}\theta}$ with $\Omega^{(\infty)} = \alpha\mI_d + U^{(\infty)} \Lambda^{(\infty)} \lrp{U^{(\infty)}}^\rT$ and $\alpha$ being the minimum eigenvalue of $\Omega^{(\infty)}$.
\end{theorem} 
Alternatively, we can express the result that with the number of stochastic gradient samples per iteration $N = \widetilde{\boldsymbol{\Omega}}\lrp{ \max\lrbb{ \frac{pd}{ m_0^2 } , \frac{p^2}{ m_0^4 } } \cdot \frac{L^2}{\alpha^2} \cdot \frac{L^2}{\epsilon^2 \sqrt{\delta} } }$, the number of iterations $T = \boldsymbol{\Theta}\lrp{ \frac{L}{\epsilon} \log\frac{1}{m_0}}$, as well as the number of samples for eigenvalue computation $M = {\boldsymbol{\Omega}}\lrp{ p \log(p/\delta) \cdot \frac{L^2}{\epsilon^2} }$,
\begin{align}
\KL{q_{\lrp{U^{(T)},\Lambda^{(T)}}}(\theta \giv \mathbf{x})}{p(\theta \giv \vx)} 
&\leq \frac{1}{2\alpha^2} \lrn{\Omega^{(\infty)}-\Omega^{(T)}}_F^2 \nonumber\\
&\leq \underbrace{ \frac{3}{2\alpha^2} \sum_{k=p+1}^d \lrp{\lambda^{(\infty)}_k}^2 }_{E_1:\text{ Approximation}} 
+ \underbrace{ \frac{6}{\alpha^2} \epsilon \sum_{k=1}^p \lambda^{(\infty)}_k + \frac{4\epsilon^2}{\alpha^2} }_{E_2:\text{ Optimization}}.
\label{eq:thm_1_alt}
\end{align}

The proof of \Cref{thm:KL_convergence} is in \Cref{appnd:A}. The key idea of the proof is to combine the convergence analyses of \Cref{lem:Oja_convergence,lem:lambda} and obtain an upper bound on the Frobenius norm difference $\lrn{ \Omega^{(\infty)} - \Omega^{(T)} }_F$ via the difference in the eigenvalues of $\Omega^{(\infty)}$ and $\Omega^{(T)}$.
We then upper bound the KL divergence with $\lrn{ \Omega^{(\infty)} - \Omega^{(T)} }_F$.

\Cref{thm:KL_convergence} implies that the posterior estimation error decomposes into two terms, $E_1$ and $E_2$.
The term $E_1$ corresponds to an irremovable bias due to the variational approximation of \Cref{eq:VI_family}, namely we use a $p$-dimensional subspace to approximate the $d$-dimensional covariance structure. If $\lambda_k^*=0$ for all $k>p^*$, then it is apparent that we should choose $p\leq p^*$. 

The second term $E_2$ is numerical error that scales as $\boldsymbol{\mathcal{O}}\lrp{ \lrp{{p \cdot d}/{\Pi}}^{1/3} }$. It decays with the total computational budget $\Pi$ and increases with the rank $p$. 
It is worth noting that, for a fixed accuracy requirement, the computation resource scales linearly with $p\cdot d$.
The scaling of $(1/\Pi)^{1/3}$ in $E_2$ stems from \Cref{proposition:computation_main}, which requires $1/\epsilon$ number of iterations and $1/\epsilon^2$ samples per iteration for the SVI algorithm to converge to $\epsilon$ accuracy in recovering the leading Rayleigh quotients.

Finally, we remark that both error terms $E_1$ and $E_2$ in \Cref{thm:KL_convergence} are invariant to the scale of the precision matrix; they scale with $\lambda_k^{(\infty)}/\alpha$,  where the eigenvalues $\lambda_k^{(\infty)}$ are normalized by the smallest eigenvalue of the true posterior covariance $\alpha$.

\parhead{Practical implications of \Cref{thm:KL_convergence} on
inferential model selection.} In what follows, we demonstrate how
\Cref{thm:KL_convergence} informs optimal inferential model selection.
Given the tolerance level of the posterior inference error and the
computational budget, we find the optimal rank $p$ of the inferential
model: it shall minimize the bound in \Cref{thm:KL_convergence}, which
implies an optimal trade-off between the approximation error and the
optimization error. 

We focus on the setting where the covariance matrix of the true
posterior has eigenvalues  $\lambda_k^{(\infty)}$ with a power law
decay, i.e. $\lambda_k^{(\infty)} = c \cdot (1/k)^\beta$, $\beta>1/2$,
for some constant $c$.
We first determine the optimal rank $p^*$ given a fixed computational
budget $\Pi$. We then compute, under this optimal choice of the rank,
the minimal computational cost $\Pi$ to guarantee an overall accuracy
requirement: $\KL{q_{\lrp{U^{(T)},\Lambda^{(T)}}}(\theta \giv
\mathbf{x})}{p(\theta \giv \vx)} = E_1 + E_2 \leq \nu_{\mathrm{KL}}$.

In more detail, we first simplify the bounds on $E_1$ and $E_2$ in \Cref{thm:KL_convergence} as follows,
$$E_1 \lesssim \frac{c^2}{\alpha^2} \max\lrbb{\frac{d^{1-2\beta}}{1-2\beta}, \frac{p^{1-2\beta}}{2\beta-1}};$$
$$E_2 \lesssim \frac{c}{\alpha} \cdot \lrp{\frac{pd}{\Pi}}^{1/3} \cdot \lrp{\frac{L}{\alpha}}^{5/3} \cdot \frac{1}{\delta^{1/6}} \cdot \max\lrbb{\frac{p^{1-\beta}}{1-\beta}, \frac{1-p^{1-\beta}}{\beta-1} + 1}.$$
Under the assumption that $\beta>1/2$, we obtain that the optimal rank $p^*$ given the total computational budget $\Pi$ is
\begin{align}
p^* \asymp
\left\{
\begin{array}{l}
     \lrp{\frac{\Pi}{d}}^{\frac{1}{6\beta - 2}} \cdot \lrp{ \frac{c}{\alpha} }^{\frac{1}{2\beta - 2/3}} \cdot \lrp{\frac{\alpha}{L}}^{\frac{5}{6 \beta - 2}} \cdot \delta^{\frac{1}{12\beta - 4}} , \qquad \beta > 1 \\
     \lrp{\frac{\Pi}{d}}^{\frac{1}{3\beta +1}} \cdot \lrp{ \frac{c}{\alpha} }^{\frac{1}{\beta + 1/3}} \cdot \lrp{\frac{\alpha}{L}}^{\frac{5}{3\beta +1}} \cdot \delta^{\frac{1}{6\beta + 2}} , \qquad 1/2<\beta<1 \\
     \min\lrbb{\lrp{\frac{\Pi}{d}}^{1/4},d} , \qquad \qquad \qquad \qquad \beta=1
\end{array}
\right..
\label{eq:optimal_rank}
\end{align}
In particular, when $\beta=1$, the optimal choice of the inferential model is smaller than the latent variable dimension $d$ when the computation budget $\Pi$ is limited.

Under the optimal choice of $p^*$ laid out in \Cref{eq:optimal_rank}, the budget $\Pi$ to achieve an overall error of $E_1+E_2\leq \nu_{\mathrm{KL}}$ is
\begin{align}
\Pi \asymp
\left\{
\begin{array}{l}
     d \cdot \nu_{\mathrm{KL}}^{-3-\frac{1}{2\beta-1}} \cdot \lrp{\frac{L}{\alpha}}^5 \lrp{\frac{c}{\alpha}}^{3+\frac{2}{2\beta-1}} \lrp{\frac{1}{\delta}}^{1/2}, \qquad \beta>1 \\
     d \cdot \nu_{\mathrm{KL}}^{-\frac{3\beta+1}{2\beta-1}} \cdot \lrp{\frac{L}{\alpha}}^5 \lrp{\frac{c}{\alpha}}^{\frac{5}{2\beta-1}} \lrp{\frac{1}{\delta}}^{1/2}, \qquad 1/2<\beta<1  \\
     d \cdot \nu_{\mathrm{KL}}^{-4} \cdot \lrp{\frac{L}{\alpha}}^5 \lrp{\frac{c}{\alpha}}^{5} \lrp{\frac{1}{\delta}}^{1/2}, \qquad \qquad \qquad \beta=1
\end{array}
\right..
\label{eq:optimal_compute}
\end{align}

From \Cref{eq:optimal_rank}, we observe that the optimal rank $p^*$ scales with the computational budget $\Pi$ but does not increase when the dimension $d$ alone increases.
This $p^*$ results in a total computational budget that only scales linearly with the dimension $d$ in \Cref{eq:optimal_compute}, as opposed to the quadratic scaling if we use the naive full-rank parametrization~\citep{hoffman2019langevin}.
The main difference among different exponents $\beta$ is in the scaling with the error tolerance $\nu_{\mathrm{KL}}$:
the larger $\beta$ is, the more appealing $\Pi$ is with respect to the error tolerance $\nu_{\mathrm{KL}}$. 
For $\beta \gg 1$, $\Pi = \boldsymbol{\mathcal{O}}\lrp{ d \cdot \nu_{\mathrm{KL}}^{-3} }$;
for $\beta=1$, $\Pi = \boldsymbol{\mathcal{O}}\lrp{ d \cdot \nu_{\mathrm{KL}}^{-4} }$; when $\beta<1$, $\Pi$ is of even higher order in $\nu_{\mathrm{KL}}^{-1}$. Finally, the total computation budget also scales with $(L/\alpha)^5$, where $L/\alpha$ is the condition number of the true posterior precision.

\subsection{Frequentist uncertainty quantification error}


We next evaluate how well the variational posterior provides frequentist uncertainty quantification, i.e. how well the variational posterior provides an estimate of the precision matrix for the latent. (The posterior mean is much easier to approximate).

We consider a dataset $\mbx=\{ x_i, y_i \}_{i=1}^n$ drawn i.i.d. from a statistical model wherein each datapoint $x_i \stackrel{iid}{\sim} \mathbb{P}$ where the distribution $\mathbb{P}$ has mean $0$, covariance $\mathbb{E}[x_i x_i^\top] = \Omega^{*}$, and $\|x_i\|_2^2 \leq R$ almost surely. We study a linear regression model:
$p(y_i|x_i, \theta) \propto \prod_{i=1}^n\exp\lrp{-\frac{1}{2}(\theta^\top x_i - y_i)^2}$.
With a uniform prior distribution, the posterior precision of $\theta$ is $\Omega^{(\infty)} = \sum_{i=1}^n x_i x_i^\top$.
The scaled posterior precision $\frac{1}{n} \Omega^{(\infty)}$ provides an estimate of the data covariance $\Omega^* = \Ep{}{x_ix_i^\top}$. We compute the variational posterior of $\theta$, $q_{\Omega,\mu}(\theta|\mathbf{x}) \propto \exp\left( -(\theta-\mu)^\top \Omega (\theta-\mu)\right)$. After $T$ steps, the variational parameter $\Omega^{(T)}$ we obtain provides an estimate of $\Omega^*$, whose quality is measured by its distance to the data covariance: $\lrn{ \Omega^* - \frac{1}{n}\Omega^{(T)} }_F$.
This evaluation takes a frequentist perspective, where we consider the precision matrix of the variational posterior as a quantification of the sampling uncertainty of the data.
It is in contrast to the Bayesian perspective where we consider the variational posterior as an approximation of the exact posterior. 


Below we quantify the error of $\Omega$ in Frobenius norm, $\lrn{ \Omega^* - \frac{1}{n}\Omega^{(T)} }_F^2$, associated with the precision matrix $\Omega^{T}$ returned by $T$ steps of the SVI algorithm.

\begin{theorem}\label{prop:stat}
Assume we have $n$ i.i.d. data samples $\{ x_i \}_{i=1}^n$ drawn from some zero-mean distribution $\mathbb{P}$ with bounded support ($\lrn{x_i}^2 \leq R$) and covariance matrix $\Omega^*=\E{x_i x_i^\top}$ satisfying $\alpha^* \mI \preceq \Omega^* \preceq L^* \mI$. 
We run the SVI algorithm (\Cref{alg:General}) with input matrix $\Omega=\mI$.
Suppose that we are given a total computation budget that allows for $\Pi=N\cdot T + 2M$ gradient evaluations in \Cref{alg:General} and that we use the same allocation rule for $N$, $T$, and $M$ as in \Cref{thm:KL_convergence}.
If  $n \geq 8 \frac{R^2}{(\alpha^*)^2} \log\frac{2}{\delta}$, then the precision matrix of the resulting variational posterior, i.e. $\Omega^{(T)} = \alpha\mI_d + U^{(T)} \Lambda^{(T)} \lrp{U^{(T)}}^\top$, satisfies
\begin{align*}
\lrn{\Omega^{*} - \frac{1}{n} \Omega^{(T)}}_F^2 \lesssim
\underbrace{ \sum_{k=p+1}^d \lrp{\lambda^{*}_k}^2 }_{E_1:\text{ Approximation}} 
+ \underbrace{ \lrp{\frac{pd}{\Pi}}^{1/3} \cdot \lrp{\frac{L^*}{\alpha^*}}^{2/3} \cdot \frac{L^*}{\delta^{1/6}} \cdot \sum_{k=1}^p \lambda^{*}_k }_{E_2:\text{ Optimization}}
+ \underbrace{ \frac{R^2}{n}\log\lrp{\frac{1}{\delta}} }_{E_3:\text{ Statistical}},
\end{align*}
with probability at least $1-\delta$.
\end{theorem}
The proof of \Cref{prop:stat} is in \Cref{app:stat}. 
Additional to those errors in \Cref{thm:KL_convergence}, we also need to take into account in this theorem the statistical error that is intrinsic to the data samples.
The key idea is to
adapt the Hoeffding's inequality to the sum of rank $1$ matrices and measure the distance in terms of the Frobenius norm.



\Cref{prop:stat} implies that the frequentist uncertainty quantification error consists of three terms, the variational approximation error $E_1$, the optimization error $E_2$, and the statistical estimation error $E_3$. 
The first two terms correspond exactly to those in \Cref{thm:KL_convergence}:
the $E_1$ term is an approximation error that decreases with increasing values of $p$;
the $E_2$ term  is an optimization error that can be decreased by increasing the number of gradient evaluations in the algorithm.



The third term $E_3$ is a statistical error term that decays with the number of samples $n$. This term only depends on $n$; it does not depend on the inferential model rank $p$ or the computational parameters $T, N$, and $M$. Therefore, this term can be a bottleneck in minimizing the total frequentist uncertainty quantification error in practice. 
In other words, finding the optimal inferential model for frequentist uncertainty quantification often requires we choose the computational parameters $T, N, M$ in a way that the first two error terms $E_1, E_2$ are of the same order as the third term $E_3$. Larger $T, N, M$ may further decrease $E_1, E_2$ but cannot decrease the order of the total.

\parhead{Practical implications of \Cref{prop:stat} on inferential model selection.} Similar to \Cref{thm:KL_convergence}, \Cref{prop:stat} can inform the optimal rank $p$ of the inferential model. Given a tolerance level of the frequentist uncertainty quantification error $\nu_{\mathrm{UQ}}$, we can find the optimal $p,T,N$ that achieves the desired $\nu_{\mathrm{UQ}}$ similar to the calculations in \Cref{subsec:bayes}.

One implication of these calculations is that choosing the most flexible inferential model may not be optimal even given infinite computational budget. For example, suppose  the eigenvalues of the true precision matrix $\Omega^* = \mI + U\Lambda U^\top$ follows a power-law decay, i.e. $\lambda_j = c^* j^{-\beta}$ for some parameter $\beta > 1/2$. 
We also know that, for sub-Gaussian random vectors $x$, its norm typically scale as: $\lrn{x}^2 \leq R = c_x \cdot d$.
Then in the limit of $T \to \infty$, the frequentist uncertainty quantification error satisfies
\begin{align}
\label{eq:run-to-conv}
    \lrn{ \Omega^* - \frac{1}{n} \Omega^{(\infty)} }_F^2 
    \lesssim \frac{(c^*)^2}{2\beta-1}\left( p^{-2\beta+1} - d^{-2\beta+1}\right)+  c_x^2 \frac{d^2}{n} \log\left(\frac{1}{\delta}\right)\;.
\end{align}
The minimal choice of the rank $p^*$ that maintains the optimal error of order $\frac{d^2}{n}\log(1/\delta)$ is
\begin{equation}
p^* \asymp \left( \frac{n}{d^2} \cdot \frac{\lrp{c^*/c_x}^2}{2\beta-1} \right)^{\frac{1}{2\beta-1}} .
\label{eq:optimal_rank_stats}
\end{equation}

When $\beta = 1$, the optimal choice of the inferential model has $p^* \asymp \min \lrbb{ \frac{n}{d^2}, d }$, which can be smaller than $d$ when the number of data points $n$ is small. 
In this case, the total computational budget required to achieve this optimal error can be computed as
\[
\Pi \asymp \frac{n^4}{d^7} \cdot \lrp{ \frac{L^*}{\alpha^*} }^2 \lrp{ \frac{1}{\delta} }^{1/2} \frac{L^3}{c^* c_x^2} .
\]
It can be observed that when $n<d^2$, the computational budget to achieve minimal error scales sub-linearly with the latent variable dimension $d$.

This calculation implies that, even when we run the optimization algorithm to convergence, it may not be optimal to choose the most flexible inferential model; choosing one with rank larger than $p^*$ cannot further decrease the frequentist uncertainty quantification error. Further, if we instead run the algorithm with a finite computational budget $\Pi$, the optimal choice of the rank is the minimum of that in \Cref{eq:optimal_rank} and \Cref{eq:optimal_rank_stats}. 
When $\beta = 1$, $p^* \asymp \min\lrbb{ \lrp{\frac{\Pi}{d}}^{1/4} , \frac{n}{d^2} , d } $.
As long as $\Pi \gtrsim \frac{n^4}{d^7}$, the same optimal error of order $\boldsymbol{\mathcal{O}}\lrp{\frac{d^2}{n}}$ is achieved.





\section{Extensions to General Non-Gaussian Posteriors}

\label{sec:general_posterior}

We next extend the inferential error analysis in \Cref{thm:KL_convergence} to general non-Gaussian posteriors.
In this case, due to the use of the Gaussian inferential model, there always exists an asymptotic bias in the variational posterior, even with an infinite computational budget.
We hence compare the numerical solution $(U^{(T)},\Lambda^{(T)})$ (obtained from $T$ steps of stochastic optimization algorithms)
with the optimal Gaussian approximation of the posterior $q_{\infty}\triangleq \mathcal{N} ( 0, ( \Omega^{(\infty)}_\infty )^{-1} )$ with $\Omega=\argmin_{\Omega}\KL{\mathcal{N}\lrp{0,\Omega^{-1}}}{p(\theta\giv\vx)}$. The theoretical optimum $\Omega^{(\infty)}_\infty$ can also be viewed the asymptotic solution from the SVI algorithm with full-rank Gaussian variational family and infinite computational budget, hence the notation.
Below we compare the variational posterior $q_{(U^{(T)},\Lambda^{(T)})}(\theta\giv \vx)$ against the optimal Gaussian approximation to the posterior $q_{\infty}$. 

To analyze the convergence properties of \Cref{alg:General} for general non-Gaussian posteriors, we extend the results in \Cref{subsec:sgd-analysis,sec:trade-off-normal}.
Specifically, instead of directly studying \Cref{alg:General} itself, we separate it into an inner loop (\Cref{alg:Gaussian}), which fixes $\Omega$ when generating i.i.d. samples $\lrbb{\theta_1,\dots,\theta_N}$ from the variational distribution $q_{\Omega} \sim \mathcal{N}\lrp{ 0, \Omega^{-1} }$; and an outer loop, which updates $\Omega$ (\Cref{alg:outer_loop}).

To facilitate the convergence analyses, we make three assumptions about the regularity and convexity of the function $\psi$.
\begin{itemize}
    \item We first assume that function $\psi$ is Lipschitz smooth: $\exists$ positive semi-definite matrix $\Psi$ such that $\nabla^2 \psi(\theta) \preceq \alpha\mI + \Psi, \forall \theta\in\real^d$.
    \item We then assume that the expectation of $\psi(\theta)$ is strongly convex: $\forall$ positive definite $\Omega$ satisfying $\alpha \mI\preceq\Omega\preceq L \mI$, we have $\Ep{\theta\sim \mathcal{N}\lrp{0,\Omega^{-1}}}{ \nabla^2 \psi(\theta) } \succeq \alpha \mI$, where $L = \alpha + \sigma_{\max}(\Psi)$ denotes the Lipschitz smoothness of $\psi$.
    \item At last, we assume that the expectation of $\psi(\theta)$ is Hessian Lipschitz: $\exists 0<\rho<1$ such that for any $\alpha \mI \preceq \Omega_1, \Omega_2 \preceq L \mI$, we have $d_W\lrp{ \Ep{ \theta\sim \mathcal{N}\lrp{0,\ \Omega_1^{-1}} }{ \nabla^2 \psi(\theta) } , \Ep{ \theta'\sim \mathcal{N}\lrp{0,\ \Omega_2^{-1}} }{ \nabla^2 \psi(\theta') } } \leq \rho \cdot d_W\lrp{\Omega_1,\Omega_2}$, where $d^2_W(\cdot, \cdot)$ is the Bures-Wasserstein distance on the space of positive definite matrices
$d^2_W\lrp{\Omega_1,\Omega_2}
= \tr\lrp{ \Omega_1 + \Omega_2 - 2 \lrp{\Omega_1^{1/2}\Omega_2\Omega_1^{1/2}}^{1/2} }$.
\end{itemize}

Under the above assumptions, we obtain the following convergence result for the general posterior.
\begin{theorem}
\label{thm:non-normal}
Assume that the above assumptions about the regularity and convexity (in expectation) of function $\psi$ hold.
Further assume, in the inner loop of \Cref{alg:outer_loop} (i.e. \Cref{eq:global_iter} that invokes \Cref{alg:Gaussian}), the number of stochastic gradient samples per iteration is $N = \widetilde{\boldsymbol{\Omega}}\lrp{ \max\lrbb{ \frac{pd}{ m_0^2 } , \frac{p^2}{ m_0^4 } } \cdot \frac{L^3}{\alpha^3} \cdot \frac{L^2}{\epsilon^2 \sqrt{\delta} } }$, the number of iterations is $T = \widetilde{\boldsymbol{\Theta}}\lrp{ \frac{L}{\epsilon}}$, and the samples for eigenvalue computation is $M = \widetilde{{\boldsymbol{\Omega}}}\lrp{ p \cdot \frac{L^2}{\epsilon^2} }$.
Then, after $K = \widetilde{\boldsymbol{\Omega}}(1)$ global iterations, we obtain $\lrp{U^{(T)}_{K},\Lambda^{(T)}_{K}}$ such that, for $\epsilon \leq \frac{1}{2} \sum_{i=1}^p \Lambda^{(\infty)}_{k+1}(i)$, 
\begin{align*}
\KL{q_{\lrp{U^{(T)}_{K},\Lambda^{(T)}_{K}}}(\theta \giv \mathbf{x})}{q_{\Omega^{(\infty)}_\infty}(\theta \giv \mathbf{x})}  
\lesssim \lrp{ \frac{1}{(1-\rho)^2} \frac{L}{\alpha} \log\frac{L}{\alpha} \cdot \frac{1}{\alpha^2} } \cdot \Bigg( \underbrace{ \sum_{i=p+1}^d \sigma_i(\Psi)^2 }_{\text{Approximation}} + \underbrace{ \epsilon \sum_{i=1}^p \sigma_i(\Psi) }_{\text{Optimization}} \Bigg),
\end{align*}
for $\epsilon\leq\alpha\sqrt{p}$, with $1-\delta$ probability.
\end{theorem}



The proof and a more formal statement of \Cref{thm:non-normal} and its assumptions is in \Cref{sec:non-normal_proof}. The main idea of the proof is that we separate the convergence of \Cref{alg:outer_loop} into two steps: its inner loop and its outer loop. 
We apply \Cref{thm:KL_convergence} to bound the error incurred in the inner loop of the algorithm that invokes \Cref{alg:Gaussian}. 
We then prove a contraction result in the global iteration step that maps different inputs closer to each other in the output space.

\Cref{thm:non-normal} generalizes \Cref{thm:KL_convergence} in characterizing the statistical and computational trade-off in stochastic variational inference for general posterior. 
Comparing with \Cref{eq:thm_1_alt} from \Cref{thm:KL_convergence}, \Cref{thm:non-normal} differs in a factor of $\frac{1}{(1-\rho)^2} \frac{L}{\alpha} \log\frac{L}{\alpha}$.
It implies that, under low-rank Gaussian inferential models, the variational approximation of non-Gaussian posteriors has the same approximation-optimization trade-off as that of Gaussian posteriors.
That said, we note that the additional asymptotic bias due to the use of Gaussian inferential models is not captured in \Cref{thm:non-normal}; this bias can easily dominate both the approximation and the optimization errors.

We conclude this section with a discussion of the assumptions of \Cref{thm:non-normal}. The three assumptions stated above are implied by a simpler but stronger point-wise conditions, i.e. the function $\psi$ is strongly convex and Lipschitz smooth:
$\alpha\mI \preceq \nabla^2 \psi(\theta) \preceq \alpha\mI + \Psi, \forall \theta\in\real^d$, 
and that function $\psi$ is Hessian Lipschitz (assuming constant $\rho < 1$):
\[ \lrn{ \lrp{\nabla^2 \psi(\theta)}^{1/2} - \lrp{\nabla^2 \psi(\theta')}^{1/2}}_F \leq  \rho \sqrt{\frac{\alpha}{L} \log\frac{\alpha}{L}} \cdot \lrn{\theta-\theta'}, \forall \theta, \theta' \in \real^d .\]

We finally compare the assumptions of \Cref{thm:non-normal} with those used for establishing fast convergence in MCMC methods.
The first assumption of \Cref{thm:non-normal} is a Lipschitz smooth condition that is frequently used in the analyses of gradient-based MCMC or optimization methods~\citep{Dalalyan_JRSSB,Dimension_free_MCMC,Nesterov_intro}.
The second assumption is an expected strong convexity condition, which relaxes the usual strong convexity condition being used in the optimization literature.
In the literature of MCMC methods for Bayesian inference, there are other conditions that relax the strong convexity condition and achieve fast convergence.
Such conditions are often instantiations of the log-Sobolev condition; they take the form of perturbations to a strongly convex function in a bounded region~\citep{MCMC_nonconvex,Xiang_Nonconvex,Yian_underdamped}. 
In comparison, the second assumption of \Cref{thm:non-normal} is a much weaker condition to obtain similar convergence rates.
The third assumption of expected Hessian Lipschitzness is often not required in gradient-based MCMC methods. 
It is required for technical reasons, due to the precision matrix parametrization as well as the proof technique we use.

\section{Empirical Studies}

\label{sec:empirical}

To illustrate the statistical and computational trade-offs in variational inference, we study the low-rank Gaussian inferential model in two empirical studies: one is a mean estimation problem on synthetic multivariate Gaussian data; the other is a Bayesian logistic regression problem on a real cardiac arrhythmia dataset.
In both studies, we evaluate the variational posterior obtained by both the mean-field inferential model and the low-rank Gaussian inferential model~(\Cref{eq:VI_family}). Across both studies, we find that Gaussian inferential models with a higher rank suffers from lower approximation error but incurs much higher computational costs. In contrast, the low-rank Gaussian inferential model can often yield significant computational benefits while keeping variational approximation at similar quality.


\subsection{Low-rank variational approximation of Gaussian posteriors}
\label{subsec:empirical-gaussian}

\begin{figure}
    \centering
    \begin{subfigure}[t]{0.49\linewidth}
        \centering
        \includegraphics[width=\linewidth]{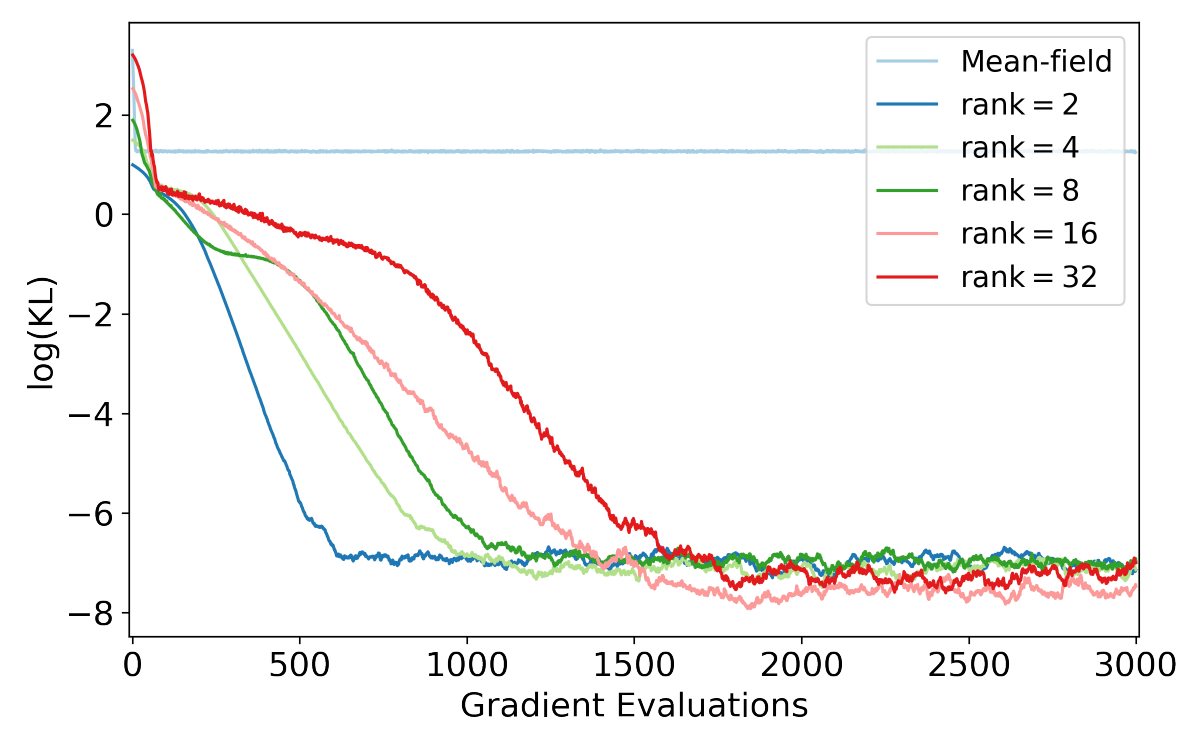}
    \end{subfigure}
    \begin{subfigure}[t]{0.49\linewidth}
        \centering
        \includegraphics[width=\linewidth]{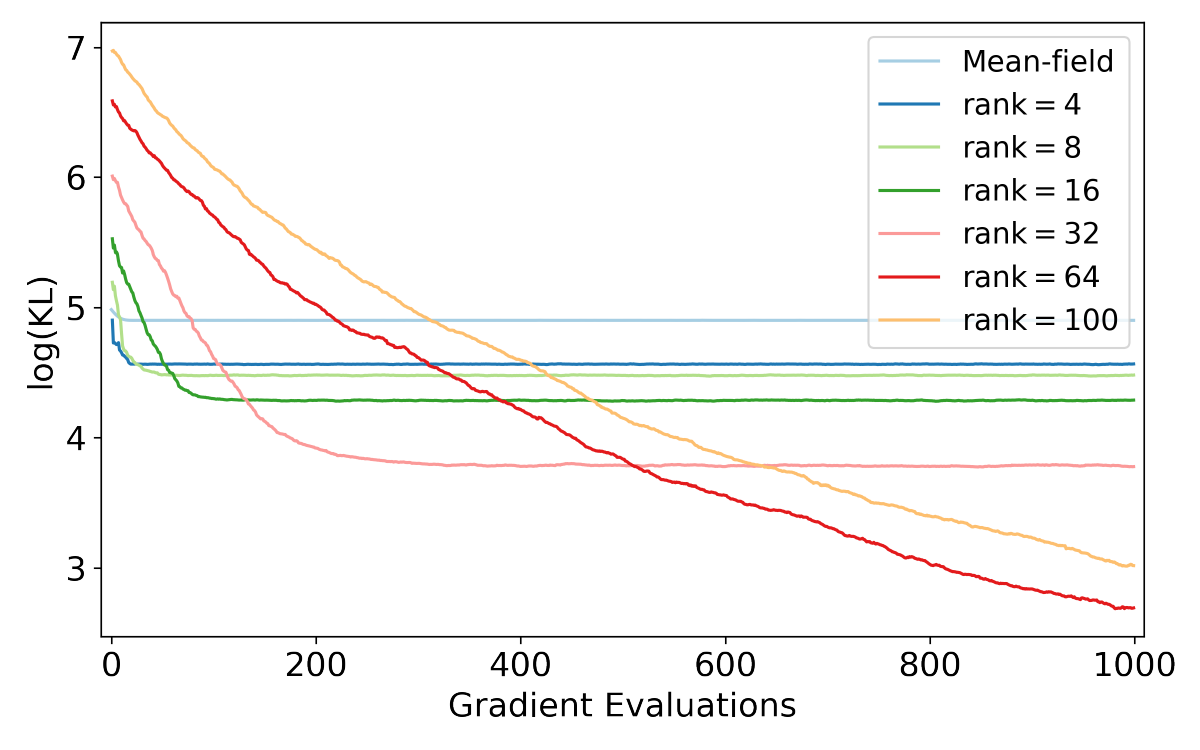}
    \end{subfigure}
    \caption{Significant computational
benefits can be achieved with employing the low-rank Gaussian inferential model, while high ranks trade computational efficiency for statistical accuracy. Evolution of KL is compared across Gaussian inferential models of different ranks as well as the mean-field inferential model. Left: dimension $d = 100$, true rank $p^*=2$. Right: dimension $d = 100$, true rank $p^*=64$.}
    \label{fig:synthetic_KL}
\end{figure}

We first study the impact of low-rank Gaussian variational approximation on Gaussian posteriors. This task is akin to the setup in \Cref{thm:KL_convergence,prop:stat} that performs variational inference on Gaussian posteriors; it evaluates the quality of low-rank Gaussian approximation of posteriors.

\parhead{Experimental setup.} We consider a setting where the true posterior 
is assumed to be centered multivariate Gaussian: $p(\theta \giv \mathbf{x}) \sim \mathcal{N}(0, \lrp{ \Omega^*}^{-1})$, with dimension $d = 100$. We consider $\Omega^* = D + M^*$, where $D = \alpha^* \mI_d$ and $M^*$ is a random symmetric positive semi-definite matrix. The goal is to find a variational approximation $q(\theta \giv \mathbf{x}) \in \mathcal{Q}_p$ to $p(\theta \giv \mathbf{x})$. 
Specifically, two configurations are examined, one constrains the random matrix $M^*$ to be of rank $p^*=2$ and the other of rank $p^*=64$.
In both cases, nonzero eigenvalues of $M^*$ are all bounded away from~0.



To illustrate the trade-offs in variational inference as in \Cref{thm:KL_convergence,prop:stat}, we parameterize the precision matrix as $\Omega = \alpha \mI_d + U \Lambda U^\rT$ and restrict our variational family accordingly: $q = q_{(U,\Lambda)}$. We adopt our stochastic variational inference algorithm (\Cref{alg:General}) and update the parameters $U, \Lambda$ in a sequential way. 
We first learn $U$ by setting $\Lambda$ to be $\mI_p$, and use QR decomposition to ensure that $U$ is semi-orthonormal.
In this experiment, we fix the number of stochastic gradient samples $N = 1$ in \Cref{alg:General}.
We then determine $\Lambda$ with the optimized $U$. The training process continues until convergence of $\KL{q}{p}$ is achieved. 


\parhead{Results.} To empirically evaluate the statistical and computational trade-offs, we study the posterior inference error of variational inference under different inferential models, namely the KL divergence between the exact posterior and its variational approximation as in \Cref{thm:KL_convergence}, including both the mean-field inferential model and the low-rank Gaussian one. 


\Cref{fig:synthetic_KL} illustrates how the posterior inference error decreases over training epochs given Gaussian inferential models of different ranks. It shows that significant computational benefits can be achieved with low-rank approximation of the posteriors. By decreasing the rank of the inferential model, we achieve fast convergence within a small number of epochs (equivalently, a small number of gradient evaluations). However, the variational approximation accuracy degrades; the correlation structure of the posterior cannot be captured in fine granularity. 

\Cref{fig:synthetic_KL} also shows that, when higher variational approximation accuracy is desired, we can increase the rank of the inferential model up to the true rank of $M^*$. When the rank of the inferential model is greater than or equal to the rank of the true model $M^*$, increasing rank would no longer improve accuracy but would notably increase the computational time. 

\begin{figure}
    \centering
    \includegraphics[height=0.8\textheight, width=0.8\textwidth, keepaspectratio]{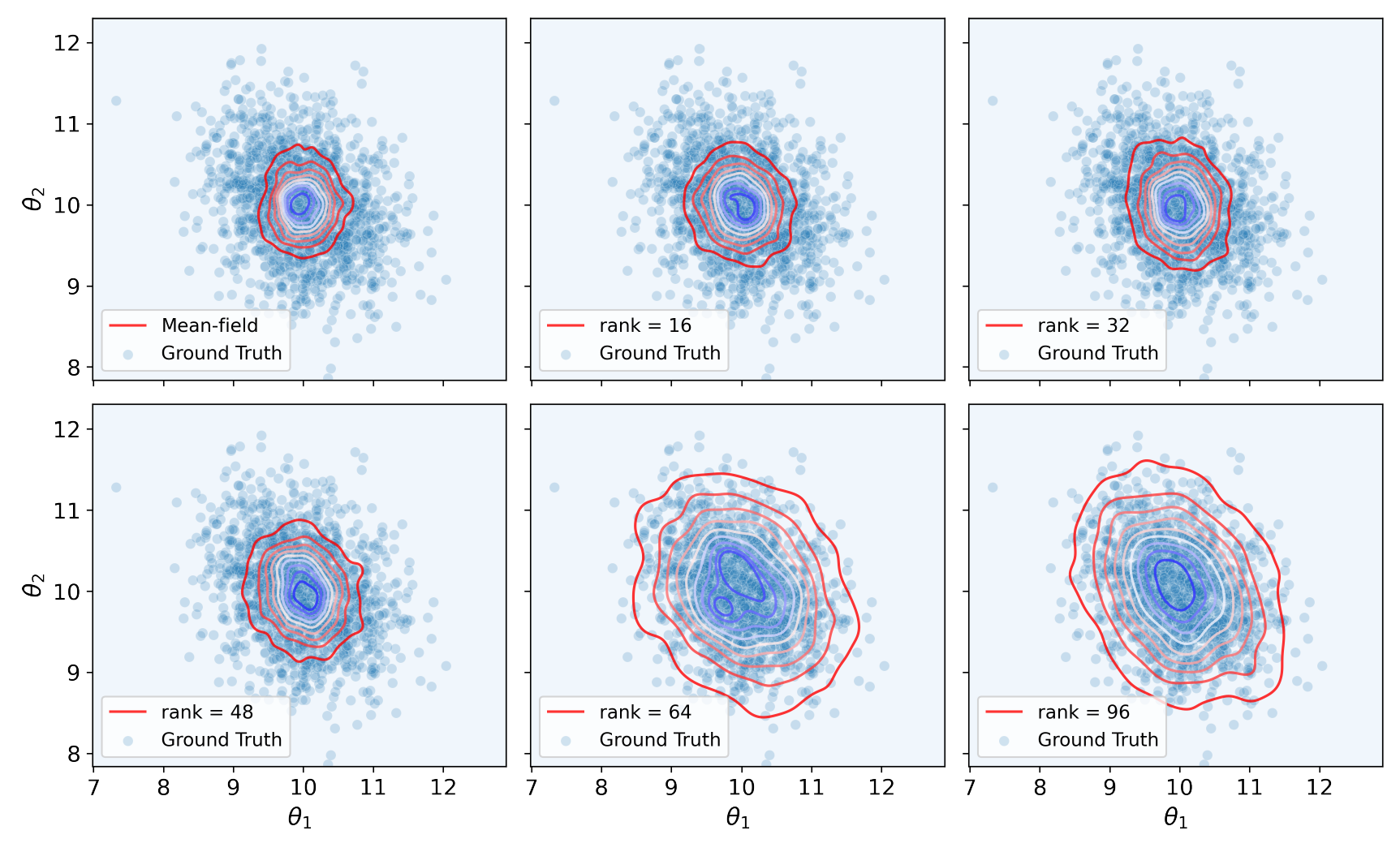}
    \caption{Gaussian inferential models with higher ranks lead to better approximation to the Gaussian posterior. The figure compares the bivariate marginals of the variational posterior with different ranks. True rank of the exact posterior is configured to be $p^* = 64$ with dimension of data $d=100$. Each distribution contour graph results from $1500$ samples from the exact posterior and its variational approximation.} 
    \label{fig:synthetic_contour}
\end{figure}

\Cref{fig:synthetic_contour} demonstrates how variational approximation quality varies under different inferential models. By comparing the contours of the  selected bivariate marginals of both the exact and approximate posteriors, we observe that variational approximation quality can be improved only up to when the true rank is reached. It also shows that the mean-field inferential model, though computationally efficient, tends to under-estimate the variance and correlation among different variables. Increasing the rank of the inferential model helps capture the posterior covariance.

\subsection{Bayesian Logistic Regression}
\label{subsec:empirical-logistic-sec}

We next study variational inference in a Bayesian logistic regression task on a real dataset of cardiac arrhythmia in the patients.\footnote{See more details of the experiment and the dataset in \Cref{sec:empirical-logistic}.} 
The goal of the logistic regression is to infer the probability of presence of cardiac arrhythmia in patients.

\begin{figure}[t]
    \centering
    \begin{subfigure}{0.49\linewidth}
        \centering
        \includegraphics[width=\linewidth]{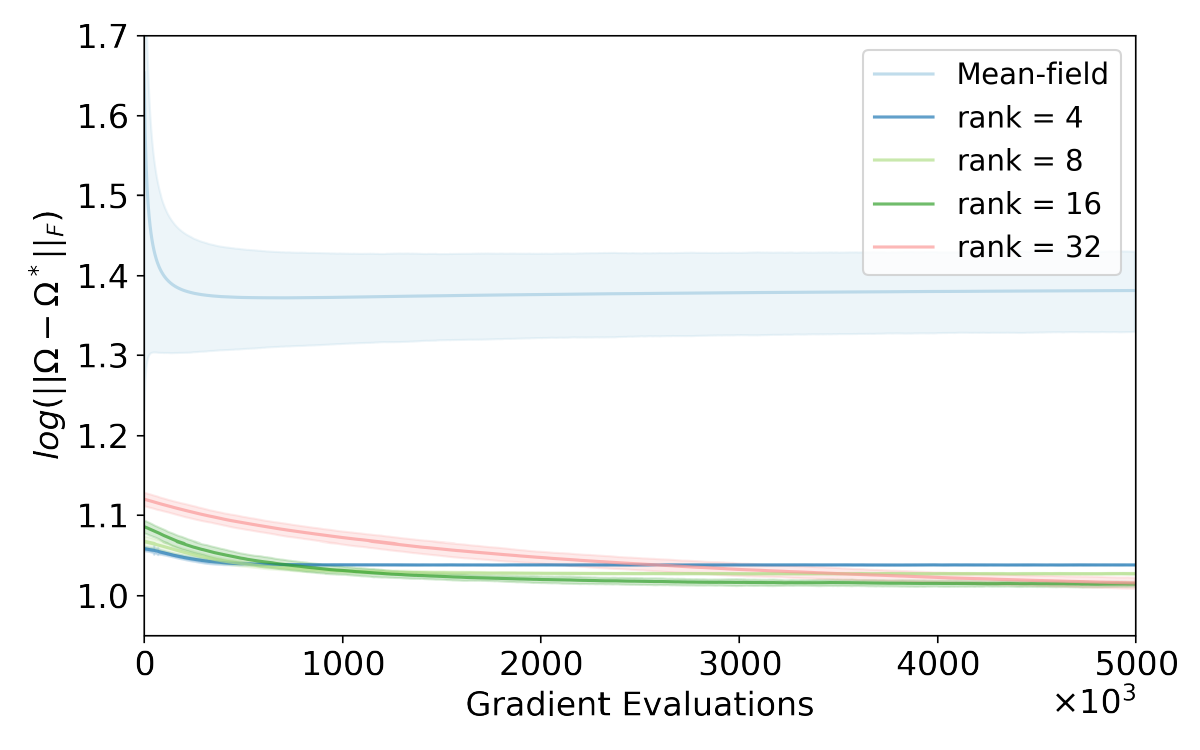}
    \end{subfigure}
    \begin{subfigure}{0.49\linewidth}
        \centering
        \includegraphics[width=\linewidth]{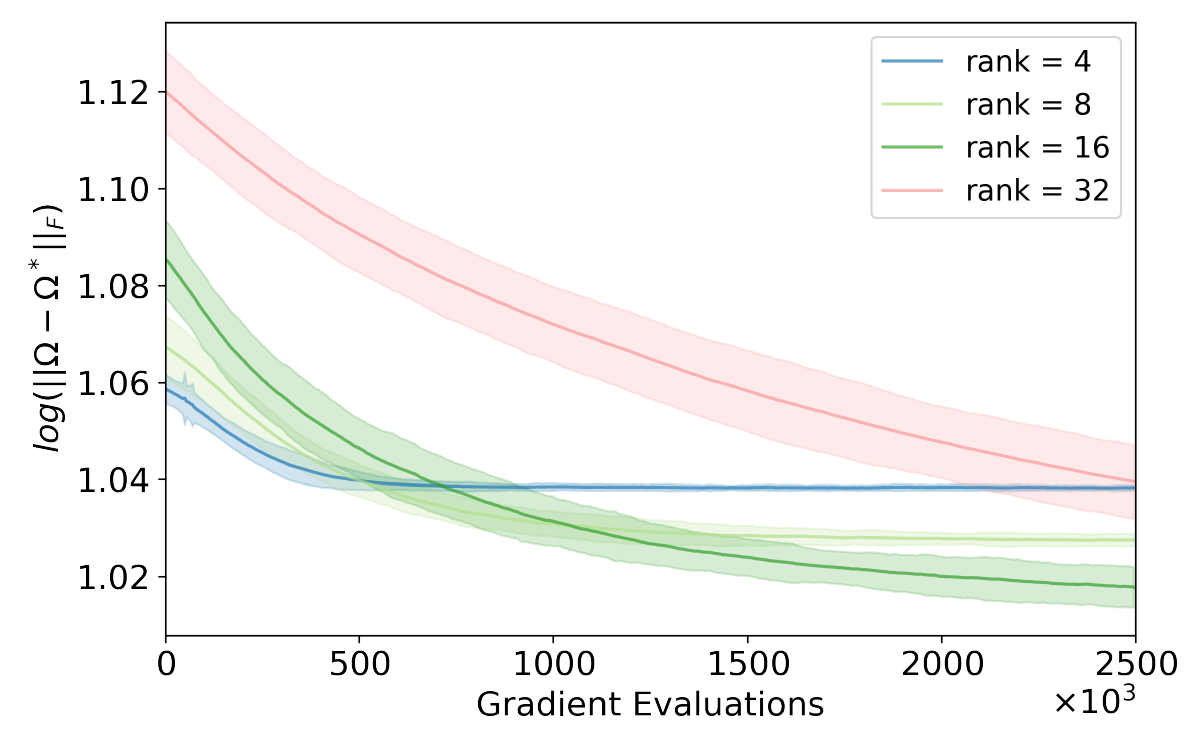}
    \end{subfigure}
    \caption{The optimal inferential model varies under different computational budget (e.g. the number of epochs) on the cardiac arrhythmia dataset (dimension d = 110). Left: Compared with low-rank inferential model, the mean-field inferential model converges to a higher posterior inference error; the convergence also exhibits higher variance. Right: Convergence of inferential models with rank $p=4, 8, 16, 32$. Inferential models with lower ranks trade statistical accuracy for computational efficiency. }
    \label{fig:Logistic_Regression}
\end{figure}

\parhead{Experimental setup.} 
We first learn the mean parameter $\mu$ and then fix $\mu$ and learn $\mathcal{N}(\mu, \Omega^{-1})$ in the Gaussian inferential model $\mathcal{Q}_p$, which is the crux of the estimation problem and our focus.
We also adopt a prior distribution of $\N(\mu, \beta \mI_d)$, with some positive constant $\beta$, over the logistic regression parameter $\theta$. 
Applying the low-rank variational inferencial model with the same parameterization as $\Omega^{(t)} = \alpha \mI_d + U^{(t)} \Lambda \lrp{U^{(t)}}^\rT$, we again run \Cref{alg:General} to sequentially optimize the learnable parameters: 
unitary matrix $U$ and diagonal matrix $\Lambda$. 
In this experiment, we fix the number of stochastic gradient samples $N = 5000$ in \Cref{alg:General} due to the higher variance in samples of the gradient log posterior as compared to the Gaussian case.

To evaluate the effectiveness of the low-rank Gaussian inferential model in variational inference, we use a long run of the No-U-Turn Sampler (NUTS) \citep{hoffman2014nuts} with multiple Markov chains to obtain a baseline estimate $\Omega^*$ of the posterior precision given the dataset.
We evaluate the distance between the baseline $\Omega^*$ and the precision matrix $\Omega$ returned by variational approximation, via measuring the Frobenius norm difference $\lrn{\Omega-\Omega^*}_F$, for both the mean-field inferential model and the low-rank Gaussian model with rank $p=4, 8, 16, 32$.


\parhead{Results.}  \Cref{fig:Logistic_Regression} compares the convergence behaviors of variational approximation under inferential models of different ranks. 
Similar to the earlier synthetic multivariate Gaussian experiment, we observe the statistical and computational trade-offs under different choices of inferential models. Specifically, the optimal inferential model varies under different computational budget (e.g. the number of epochs). 
Higher variance is also detected with increasing ranks of the inferential model. 
Moreover, low-rank Gaussian inferential models outperforms the mean-field inferential model in terms of posterior inference accuracy.
The observed discrepancy in statistical accuracy is further demonstrated in \Cref{fig:Logistic_Regression_hist,fig:datacontour} in \Cref{appnd:logistic_results}. It compares the marginal distributions obtained by both variational families against the NUTS baseline, where low-rank variational model achieves significantly better approximation with higher accuracy.


\vspace{-12pt}
\section{Discussion}
This paper initiates a theoretical study of the statistical and computational trade-offs arising in variational inference. Through a case study in inferential model selection, we establish this trade-off under the low-rank Gaussian inferential model. We prove that, as the rank of the inferential model increases, it can capture a larger subspace of the posterior uncertainty. Yet, this gain comes at the cost of slower convergence of the optimization algorithm. These results shed light on the practical success of low-rank approximations in Bayesian inference \citep{miller2017variational}. 


\section*{Acknowledgement}
We thank Sinho Chewi for supplying the proof of \Cref{lem:W_2_bound}.
This work is supported in part by the National Science Foundation Grants NSF-SCALE MoDL(2134209) and NSF-CCF-2112665 (TILOS), the U.S. Department of Energy Office of Science, and the Facebook Research Award.

\clearpage
\bibliographystyle{apalike}
\bibliography{vb_tradeoff}

\appendix
\onecolumn
\textbf{\Large{Supplementary Materials}}




\section{Details of \Cref{alg:Gaussian,alg:outer_loop}}
\label{sec:alg_details}

\begin{algorithm}[H]
\SetAlgoLined
\setstretch{0.5}
\KwInput{$\Omega$, $p$, $N$, $M$, $T$}
Initialize $U^{(0)} = \left[e_1,\dots,e_p\right]$, $\Lambda^{(0)}=\mI_p$\\
\For{$t = 0, \ldots,T-1$}{
\begin{align}
    & \text{Sample i.i.d.} \; \theta_j^{(t)} \sim \mathcal{N} \lrp{ 0, \Omega^{-1} }, \forall j\in\{1,\dots,N\}, \label{eq:sample_step_Gauss} \\
    & \tilde{U}^{(t+1)} = \sum_{j=1}^N \lrp{ \nabla\psi\lrp{\theta_j^{(t)}} \lrp{\theta_j^{(t)}}^\rT \Omega } U^{(t)},          \label{eq:mult_step_Gauss} \\
    &U^{(t+1)} = \mathrm{QR} \lrp{\widetilde{U}^{(t+1)}}. \label{eq:QR_step_Gauss}
\end{align}
 }
Draw $\lrbb{\theta_1,\dots,\theta_M}$ i.i.d. from $\mathcal{N}\lrp{ 0, \Omega^{-1} }$ \\
\For{$i=1,\dots,p$}{
\begin{align}
\Lambda_{i,i}^{(T)} = \frac{1}{M} \sum_{j=1}^M \lrp{ u_i^{(T)} }^\rT \lrp{ \frac{\nabla \psi(\theta_j+\Delta \cdot u_i^{(T)}) - \nabla \psi(\theta_j-\Delta \cdot u_i^{(T)})}{2\Delta} } - D_{i,i} . \label{eq:lambda_step_Gauss}
\end{align}
}
\KwReturn{$U^{(T)}, \Lambda^{(T)}$}
\caption{SVI algorithm for Gaussian posterior (SVI\_Gauss)}
\label{alg:Gaussian}
\end{algorithm}


\begin{algorithm}[H]
\SetAlgoLined
\setstretch{0.5}
\KwInput{$U_0 = \left[e_1,\dots,e_p\right]$, $\Lambda_0=\mI_p$, $p$, $N$, $M$, $T$, $K$}
\For{$k = 0, \ldots,K-1$}{
\begin{align}
\lefteqn{\Omega_k = D + U_k \Lambda_k \lrp{U_k}^\rT} \nonumber\\
&\lrp{ U_{k+1}, \Lambda_{k+1} } = \textrm{SVI\_Gauss}\lrp{\Omega_{k}, p, N, M, T}. \label{eq:global_iter}
\end{align}
}
\KwReturn{$ U_K, \Lambda_K $}
\caption{SVI algorithm for general posterior (SVI\_General)}
\label{alg:outer_loop}
\end{algorithm}

\section{Proofs for the Convergence of Stochastic Optimization over the KL Divergence}
\label{appnd:A}
\begin{proof}[Proof of \Cref{thm:KL_convergence}]
\begin{align*}
\KL{q_{\lrp{U^{(T)},\Lambda^{(T)}}}(\theta \giv \mathbf{x})}{p(\theta \giv \vx)} 
&= \frac{1}{2} \lrp{ \log\frac{|\Omega^{(T)}|}{|\Omega^{(\infty)}|} - d + \tr\lrp{ \lrp{ \Omega^{(T)} }^{-1} \Omega^{(\infty)} } } \\
&= \frac{1}{2} \lrp{ - \log\big| \lrp{ \Omega^{(T)} }^{-1} \Omega^{(\infty)} \big| + \tr\lrp{  \lrp{ \Omega^{(T)} }^{-1} \Omega^{(\infty)} - \mI } }.
\end{align*}
Since 
\[
\log\Big| \lrp{ \Omega^{(T)} }^{-1} \Omega^{(\infty)} \Big| \geq \tr\lrp{ \mI -  \lrp{\Omega^{(\infty)}}^{-1} \Omega^{(T)}  },
\]
\begin{align*}
\KL{q_{\lrp{U^{(T)},\Lambda^{(T)}}}(\theta \giv \mathbf{x})}{p(\theta \giv \vx)} 
&\leq \frac{1}{2} \tr\lrp{ \lrp{ \Omega^{(T)} }^{-1}\Omega^{(\infty)} + \lrp{\Omega^{(\infty)}}^{-1} \Omega^{(T)} - 2\mI } \\
&= \frac{1}{2} \tr\lrp{ \lrp{\lrp{ \Omega^{(T)} }^{-1}-\lrp{\Omega^{(\infty)}}^{-1}} \lrp{\Omega^{(\infty)}-\Omega^{(T)}} } \\
&= \frac{1}{2} \tr\lrp{ \lrp{ \Omega^{(T)} }^{-1} \lrp{\Omega^{(\infty)}-\Omega^{(T)}} \lrp{\Omega^{(\infty)}}^{-1} \lrp{\Omega^{(\infty)}-\Omega^{(T)}}  } \\
&\leq \frac{1}{2} \sigma_{\max}\lrp{\lrp{\Omega^{(\infty)}}^{-1}} \sigma_{\max}\lrp{\lrp{ \Omega^{(T)} }^{-1}} \cdot \lrn{\Omega^{(\infty)}-\Omega^{(T)}}_F^2 \\
&\leq \frac{1}{2\alpha^2} \lrn{\Omega^{(\infty)}-\Omega^{(T)}}_F^2.
\end{align*}

We then bound $\lrn{\Omega^{(\infty)}-\Omega^{(T)}}_F^2$ in what follows.
Since $\Omega^{(T)}$ and $\Omega^{(\infty)}$ share the same $D=\alpha\mI_d$ matrix,
we can define $M^{(T)} = \Omega^{(T)} - \alpha\mI_d = U^{(T)} \Lambda^{(T)} \lrp{U^{(T)}}^\rT$ and $M^{(\infty)} = \Omega^{(\infty)} - \alpha\mI_d = U^{(\infty)} \Lambda^{(\infty)} \lrp{U^{(\infty)}}^\rT$ and obtain:
\begin{align}
\tr\lrp{ \lrp{\Omega^{(\infty)}-\Omega^{(T)}}^2 }
&= \tr\lrp{\lrp{\Lambda^{(T)}}^2} + \tr\lrp{\lrp{\Lambda^{(\infty)}}^2} - 2 \tr\lrp{ \Lambda^{(T)} \lrp{U^{(T)}}^\rT M^{(\infty)} U^{(T)} }.
\label{eq:trace_eq}
\end{align}
Since $\lrp{ \lrp{U^{(T)}}^\rT M^{(\infty)} U^{(T)} }_{i,j} = \lrp{u_i^{(T)}}^\rT M^{(\infty)} u_j^{(T)}$, we have
$\lrp{ \Lambda^{(T)} \lrp{U^{(T)}}^\rT M^{(\infty)} U^{(T)} }_{i,j} = \lambda_i^{(T)} \lrp{u_i^{(T)}}^\rT M^{(\infty)} u_j^{(T)}$.
Therefore, 
\[
\tr\lrp{\Lambda^{(T)} \lrp{U^{(T)}}^\rT M^{(\infty)} U^{(T)} } = \sum_{k=1}^p \lambda_k^{(T)} \lrp{u_k^{(T)}}^\rT M^{(\infty)} u_k^{(T)}.
\]

From \Cref{lem:Oja_convergence}, we know that taking number of stochastic gradient samples per iteration $N = \widetilde{\boldsymbol{\Omega}}\lrp{ \max\lrbb{ \frac{pd}{ m_0^2 } , \frac{p^2}{ m_0^4 } } \cdot \frac{L^4}{\alpha^2} \cdot \frac{\kappa(\Omega)}{\epsilon^2 \sqrt{\delta} } }$, 
and number of iterations $T = \boldsymbol{\Theta}\lrp{ \frac{L}{\epsilon} \log\frac{1}{m_0}}$,
we have with probability $1-\delta/2$:
\[
\lrp{u^{(T)}_k}^\rT M^{(\infty)} u^{(T)}_k \geq \lambda^{(\infty)}_k - 2\epsilon,
\qquad\forall k\in\{1,\dots,p\}.
\]
Hence $\tr\lrp{\Lambda^{(T)} \lrp{U^{(T)}}^\rT M^{(\infty)} U^{(T)} } \geq \sum_{k=1}^p \lrp{ \lambda_k^{(T)} \lambda_k^{(\infty)} - 2\epsilon \lambda^{(\infty)}_k }$.
Therefore, 
\begin{align}
&\tr\lrp{ \lrp{\Omega^{(\infty)}-\Omega^{(T)}}^2 } \\
&= \sum_{k=1}^p \lrp{ \lambda_k^{(T)} }^2 + \sum_{k=1}^d \lrp{\lambda^{(\infty)}_k}^2 - 2 \sum_{k=1}^p \lambda_k^{(T)} \lrp{u_k^{(T)}}^\rT M^{(\infty)} u_k^{(T)} \nonumber\\
&= \sum_{k=p+1}^d \lrp{\lambda^{(\infty)}_k}^2 + \sum_{k=1}^p \lrp{ \lambda_k^{(T)} }^2 + \sum_{k=1}^p \lrp{\lambda^{(\infty)}_k}^2 - 2 \sum_{k=1}^p \lambda_k^{(T)} \lrp{ u_k^{(T)}} ^\rT M^{(\infty)} u_k^{(T)} \nonumber\\
&\leq \sum_{k=p+1}^d \lrp{\lambda^{(\infty)}_k}^2 + \sum_{k=1}^p \lrp{\lambda^{(\infty)}_k - \lambda_k^{(T)}}^2 + 4 \epsilon \sum_{k=1}^p \lambda^{(\infty)}_k.
\label{eq:appnd_trace_bound}
\end{align}
From \Cref{lem:lambda}, we know that when we take $n \geq 2 p \log(8p/\delta) \cdot \frac{L^2}{\epsilon^2}$ and $\varDelta \leq \epsilon/\lrp{ \sqrt{p} \cdot L_{\rm{Hess}} }$ in \Cref{alg:Gaussian} (\Cref{eq:lambda_step_Gauss}), we can achieve with $1-\delta/2$ probability that:
\begin{align}
\sum_{k=1}^p \lrp{\lambda^{(\infty)}_k - \lambda^{(T)}_k}^2 
\leq 2 \sum_{k=p+1}^d \lrp{\lambda^{(\infty)}_k}^2 + 8 \epsilon \sum_{k=1}^p \lambda^{(\infty)}_k + 8 \epsilon^2.
\label{eq:appnd_lambda_bound}
\end{align}
Plugging equation~\eqref{eq:appnd_lambda_bound} into \eqref{eq:appnd_trace_bound}, we arrive at the result that with $1-\delta$ probability,
\begin{align*}
\lrn{ \Omega^{(\infty)} - \Omega^{(T)} }_F^2
\leq 3 \sum_{k=p+1}^d (\lambda^{(\infty)}_k)^2 + 12\epsilon \sum_{k=1}^p \lambda_k^{(\infty)} + 8 \epsilon^2.
\end{align*}
Therefore with $1-\delta$ probability,
\begin{align*}
\KL{q_{\lrp{U^{(T)},\Lambda^{(T)}}}(\theta \giv \mathbf{x})}{p(\theta \giv \vx)} 
&\leq \frac{1}{2\alpha^2} \lrn{\Omega^{(\infty)}-\Omega^{(T)}}_F^2 \\
&\leq \frac{3}{2\alpha^2} \sum_{k=p+1}^d \lrp{\lambda^{(\infty)}_k}^2 + \frac{6}{\alpha^2} \epsilon \sum_{k=1}^p \lambda^{(\infty)}_k + \frac{4\epsilon^2}{\alpha^2}.
\end{align*}

Plugging in the computation budget of $\Pi=N\cdot T + 2M = \widetilde{\boldsymbol{\Omega}}\lrp{pd\cdot \frac{L^5}{\alpha^2}\cdot\frac{1}{\epsilon^3\sqrt{\delta}}}$, we obtain that 
\begin{align*}
\KL{q_{\lrp{U^{(T)},\Lambda^{(T)}}}(\theta \giv \mathbf{x})}{p(\theta \giv \vx)} 
\lesssim \frac{1}{\alpha^2} \sum_{k=p+1}^d \lrp{\lambda^{(\infty)}_k}^2 + \lrp{\frac{pd}{\Pi}}^{1/3} \cdot \lrp{\frac{L}{\alpha}}^{5/3} \cdot \frac{1}{\delta^{1/6}} \cdot \frac{1}{\alpha} \sum_{k=1}^p \lambda^{(\infty)}_k,
\end{align*}
with $1-\delta$ probability.
\end{proof}
\subsection{Proofs for solving for $U$}
\label{sec:pf_U}

\begin{proof}[Proof of \Cref{fact:precond_equiv}]
Taking $h_t=1$ and fixing $\Lambda = \mI_p$ in the updates of equations~\eqref{eq:precond_GD_1}--\eqref{eq:precond_GD_3}, we have
\begin{align}
& \text{Sample i.i.d.} \; \theta_j^{(t)} \sim q_{ \lrp{U^{(t)},\Lambda} }, \forall j=\{1,\dots,N\} \\
& \tilde{U}^{(t+1)} = \frac{1}{N} \sum_{j=1}^N \lrp{ \nabla\psi\lrp{\theta_j^{(t)}} \theta_j^\rT } U^{(t)} \label{eq:precond_GD_2_appnd} \\
& U^{(t+1)} = \mathrm{QR}\lrp{\tilde{U}^{(t+1)}},
\end{align}
where $q_{ \lrp{U^{(t)},\Lambda} } = \mathcal{N}\lrp{0, \lrp{\Omega^{(t)}}^{-1} }$, since $\Omega^{(t)} = D + U^{(t)} \Lambda \lrp{U^{(t)}}^\rT$.

For $D=\alpha\mI_{d}$, and for $\Lambda = \mI_p$,
\begin{align*}
\Omega^{(t)} U^{(t)} 
= \lrp{ D + U^{(t)} \Lambda \lrp{U^{(t)}}^\rT } U^{(t)} 
= \lrp{ \alpha\mI_d + U^{(t)} \mI_p \lrp{U^{(t)}}^\rT } U^{(t)}
= \lrp{\alpha+1} U^{(t)}.
\end{align*}
Equation~\eqref{eq:precond_GD_2_appnd} becomes 
\[
\tilde{U}^{(t+1)} = \frac{1}{(\alpha+1)N} \sum_{j=1}^N \lrp{ \nabla\psi\lrp{\theta_j^{(t)}} \theta_j^\rT } \Omega^{(t)} U^{(t)}.
\]
Combining with the fact that the QR decomposition will remove any scaling factor, we obtain the final result that the updates of equations~\eqref{eq:precond_GD_1}--\eqref{eq:precond_GD_3} is equivalent to:
\begin{align}
& \text{Sample i.i.d.} \; \theta_j^{(t)} \sim \mathcal{N}\lrp{0, \lrp{\Omega^{(t)}}^{-1} }, \forall j=\{1,\dots,N\} \\
& \tilde{U}^{(t+1)} = \sum_{j=1}^N \lrp{ \nabla\psi\lrp{\theta_j^{(t)}} \theta_j^\rT } \Omega^{(t)} U^{(t)} \\
& U^{(t+1)} = \mathrm{QR}\lrp{\tilde{U}^{(t+1)}},
\end{align}
where $\Omega^{(t)} = D + U^{(t)} \Lambda \lrp{U^{(t)}}^{\rT} = \alpha\mI_d + U^{(t)} \lrp{U^{(t)}}^\rT$.

In addition, if the posterior is a normal distribution, then $\psi$ is a quadratic function and $\psi(\theta) = \frac{1}{2} \theta^\rT \Omega^{(\infty)} \theta$.
In that case, define stochastic update matrix $\widetilde{F}_\psi = \frac{1}{N} \sum_{i=1}^N \nabla \psi(\theta_i) \theta_i^\rT \Omega^{(t-1)}$. Then
$$F_\psi = \mathbb{E}[\widetilde{F}_\psi] = \Ep{ \theta\sim\mathcal{N}\lrp{ 0, \lrp{\Omega^{(t-1)}}^{-1} } }{ \Omega^{(\infty)} \theta \theta^\rT \Omega^{(t-1)} } = \Omega^{(\infty)}.$$
Substituting $\lrp{\Omega^{(t-1)}}$ with an arbitrary $\Omega$ yields the same result that
$$\Ep{ \theta\sim\mathcal{N}\lrp{ 0, \lrp{\Omega}^{-1} } }{ \Omega^{(\infty)} \theta \theta^\rT \Omega } = \Omega^{(\infty)}
= F_\psi.$$
We can therefore apply the following update rule in the Gaussian posterior case:
\begin{align}
& \text{Sample i.i.d.} \; \theta_j^{(t)} \sim \mathcal{N}\lrp{0, \lrp{\Omega}^{-1} }, \forall j=\{1,\dots,N\} \\
& \tilde{U}^{(t+1)} = \sum_{j=1}^N \lrp{ \nabla\psi\lrp{\theta_j^{(t)}} \theta_j^\rT } \Omega \cdot U^{(t)} \\
& U^{(t+1)} = \mathrm{QR}\lrp{\tilde{U}^{(t+1)}},
\end{align}
for any positive definite $\Omega$.
\end{proof}

\begin{proof}[Proof of \Cref{lem:Oja_convergence}]
When $p(\theta|\vx)$ is a normal distribution, $\nabla_{\theta}^2 \psi(\theta) = \Omega^{(\infty)}$, and that $\nabla_{\theta} \psi(\theta) = \Omega^{(\infty)} \theta$, $\forall \theta\in\mathbb{R}^d$.
%
Since $\psi$ is $\alpha$-strongly convex, we consider the following form: 
$$ \Omega^{(\infty)} = U^{(\infty)} \Lambda^{(\infty)} \lrp{U^{(\infty)}}^\rT + \alpha \mI_{d\times d} = U^{(\infty)} \lrp{ \Lambda^{(\infty)} + \alpha \mI_{d\times d} } \lrp{ U^{(\infty)} }^\rT, $$ 
where $U^{(\infty)}, \Lambda^{(\infty)}\in\R^{d\times d}$, and the eigenvalues $\lambda^{(\infty)}_k = \Lambda^{(\infty)}_{k,k}$ are non-negative and are ranked in a descending order.

We denote ${U}^{(\infty)}_{k}$ as the matrix consisting of the top $k$ eigenvectors of $\Omega^{(\infty)}$.
Conversely, denote $U^{(\infty)}_{-k}$ as the matrix consisting of the bottom $(d-k)$ eigenvectors of $\Omega^{(\infty)}$ and similarly ${U}^{(\infty)}_{-k-\iota(\epsilon)}$ as the matrix consisting of all eigenvectors of $\Omega^{(\infty)}$ with eigenvalues less than or equal to $\lambda^{(\infty)}_k + \alpha - \epsilon$ (assuming there are $(d-k-\iota(\epsilon))$ of them).
%
For $U^{(t)} \in\R^{p\times p}$,
we similarly denote for $k\leq p$: $U^{(t)}_k = \left[u^{(t)}_1,\dots,u^{(t)}_k\right]$ as the matrix consisting of the first $k$-columns of $U^{(t)}$.


We now leverage the following \Cref{proposition:computation_main} to bound the Rayleigh quotient $\lrp{u^{(T)}_k}^\rT \Omega^{(\infty)} u^{(T)}_k$.

We apply \Cref{proposition:computation_main} and similarly let $m_0 = \sigma_{\min}\lrp{ (U^{(\infty)}_p)^\rT U^{(0)} } \geq \max\lrp{0, 1- \lrn{U^{(\infty)}_p-U^{(0)}}}$ to prove the lemma.
We run the stochastic variational inference algorithm for Gaussian posterior described in equations~\eqref{eq:sample_Gauss} to~\eqref{eq:QR_Gauss} with input matrix $\Omega$,
number of stochastic gradient samples per iteration $N = \widetilde{\boldsymbol{\Omega}}\lrp{ \max\lrbb{ \frac{pd}{ m_0^2 } , \frac{p^2}{ m_0^4 } } \cdot \frac{L^4}{\alpha^2} \cdot \frac{\kappa(\Omega)}{\epsilon^2 \sqrt{\delta} } }$, 
and number of iterations $T = \Theta\lrp{ \frac{L}{\epsilon} \log\frac{1}{m_0}}$.
We obtain that 
with probability $1 - \delta$, for all $m\geq k$, for any $k\leq p$,
\[
\lrn{ \lrp{{U}^{(\infty)}_{-m-1}}^\rT U_k^{(T)} } \leq \frac{\epsilon}{\lambda_k^{(\infty)} - \lambda_{m+1}^{(\infty)}},
\]
which implies that with probability $1 - \delta$,
\begin{align}
\forall k\in\{1,\dots,p\}, \ \forall \gamma_{k,i} = \frac{\lambda_k^{(\infty)} - \lambda_i^{(\infty)}}{\epsilon} \geq 1, 
\qquad \lrn{ \lrp{{U}^{(\infty)}_{-k-\iota(\gamma\cdot\epsilon)}}^\rT u^{(t)}_k }_2^2 \leq 1/\gamma_{k,i}^2.
\label{eq:appnd_eig_union}
\end{align}

We use the above result to bound the Rayleigh quotient for $\Omega^{(\infty)} 
= U^{(\infty)} \lrp{ {\Lambda}^{(\infty)} + \alpha \mI_{d\times d} } \lrp{{U}^{(\infty)}}^\rT$.
For each $k\in[p]$, denote $k_0 = k+\iota(\epsilon)$. 
Then $(k_0+1)$ indexes the first eigenvector with eigenvalue less than or equal to $\lambda^{(\infty)}_k + \alpha - \epsilon$.
Define $b_{k,j} = \sum_{s=j}^d \lrw{u^{(t)}_k,u^{(\infty)}_s}^2$ where $u^{(\infty)}_s$ is the $s$-th eigenvector of $\Omega^{(\infty)}$.
It satisfies $b_{k,1} = \lrp{u^{(t)}_k}^\rT \lrp{\sum_{s=1}^d u^{(\infty)}_s \lrp{u^{(\infty)}_s}^\rT} u^{(t)}_k = 1$. By Abel's transformation,
\begin{align}
\lrp{u^{(t)}_k}^\rT \Omega^{(\infty)} u^{(t)}_k 
&= \sum_{j=1}^d \lrp{\lambda^{(\infty)}_j+\alpha} \lrw{u^{(t)}_k, u^{(\infty)}_j}^2 \nonumber\\
&= \sum_{j=1}^{k_0-1} \lrp{\lambda^{(\infty)}_j+\alpha} \lrw{u^{(t)}_k, u^{(\infty)}_j}^2 + \lrp{\lambda^{(\infty)}_{k_0}+\alpha} b_{k,k_0} - \sum_{j=k_0+1}^d b_{k,j} \lrp{\lambda^{(\infty)}_{j-1}-\lambda^{(\infty)}_j} \nonumber\\
&\geq (\lambda^{(\infty)}_k + \alpha - \epsilon) b_{k,1} - \sum_{j=k_0+1}^d b_{k,j} \lrp{\lambda^{(\infty)}_{j-1}-\lambda^{(\infty)}_j} \nonumber\\
&= (\lambda^{(\infty)}_k + \alpha - \epsilon) - \sum_{j=k_0+1}^d b_{k,j} \lrp{\lambda^{(\infty)}_{j-1}-\lambda^{(\infty)}_j}.
\label{eq:appnd_eig_sum}
\end{align}
According to equation~\eqref{eq:appnd_eig_union}, for every $i\geq k_0+1$, $b_{k,j} = \lrn{ {U}^{(\infty)}_{-k-\iota(\gamma_{k,j}\cdot\epsilon)} u^{(t)}_k }_2^2 \leq \frac{1}{\gamma_{k,j}^2}$.
Plugging this result into equation~\eqref{eq:appnd_eig_sum}, we obtain
\[
\sum_{j=k_0+1}^d b_{k,j} \lrp{\lambda^{(\infty)}_{j-1}-\lambda^{(\infty)}_j}
\leq \sum_{j=k_0+1}^d \frac{1}{\gamma_{k,j}^2} \epsilon\lrp{\gamma_{k,j}-\gamma_{k,j-1}}
\leq \epsilon \int_{1}^{1/\epsilon} \frac{1}{\gamma^2} d\gamma
\leq \epsilon.
\]
Therefore, with probability $1-\delta$,
\[
\lrp{u^{(t)}_k}^\rT \Omega^{(\infty)} u^{(t)}_k \geq \lambda^{(\infty)}_k + \alpha - 2\epsilon.
\]
\end{proof}

\begin{proposition}
\label{proposition:computation_main}
Assume that function $\psi(\theta) = \frac{1}{2} \theta^\rT \Omega^{(\infty)} \theta$ (so that the posterior $p(\theta|\vx)$ is Gaussian) and is $\alpha$-strongly convex and $L$-Lipschitz smooth.
We denote the initial condition: $m_0 = \sigma_{\min}\lrp{ (U^{(\infty)}_p)^\rT U^{(0)} } \geq \max\lrp{0, 1- \lrn{U^{(\infty)}_p - U^{(0)}}}$.
We run the stochastic variational inference algorithm for Gaussian posterior described in equations~\eqref{eq:sample_Gauss} to~\eqref{eq:QR_Gauss} with input matrix $\Omega$, number of stochastic gradient samples per iteration, and number of iterations:
\begin{align*}
N &= \widetilde{\Omega}\lrp{ \max\lrbb{ \frac{pd}{ m_0^2 } , \frac{p^2}{ m_0^4 } } \cdot \frac{L^4}{\alpha^2} \cdot \frac{\kappa(\Omega)}{\epsilon^2 \sqrt{\delta} } }, \\
T &= \Theta\lrp{ \frac{L}{\epsilon} \log\frac{1}{m_0}},
\end{align*}
where $\epsilon\leq\lambda_k$.
Then with probability $1-\delta$,
\[
\lrn{ \lrp{U^{(\infty)}_{-m-1}}^\rT U_k^{(T)} } \leq \frac{\epsilon}{\lambda_k^{(\infty)} - \lambda_{m+1}^{(\infty)}},
\]
for all $m\geq k$, for any $k\leq p$.
\end{proposition}

\begin{proof}[Proof of \Cref{proposition:computation_main}]
We now provide the proof of \Cref{proposition:computation_main}.
For succinctness, we denote in this section $V_k={U}^{(\infty)}_{k}$;
$W_m={U}^{(\infty)}_{-m-1}$;
and the stochastic gradient noise restricted to the first $k$ columns: \[
\Gt = \frac{1}{N} \sum_{j=1}^N \nabla \psi(\theta_j) \theta_j^\rT \Omega \Xt - \Ep{\theta\sim\mathcal{N}\lrp{0,\Omega^{-1}}}{\nabla \psi(\theta) \theta^\rT \Omega \Xt}.
\]
We first state the following main lemma that will help us bound our objective via contraction.
\begin{lemma}
\label{lemma:cos_iter}
Assume that $\lrn{\Gt} \leq \Delta_1 \cdot \lambda_k^{(\infty)} \cdot \sigma_{\min}\lrp{V_k^\rT \Xt}$ and $\lrn{V_k^\rT \Gt} \leq \Delta_2 \cdot \lambda_k^{(\infty)} \cdot \sigma^2_{\min}\lrp{V_k^\rT \Xt}$, where $\Delta_2\leq1/3$.
Then for
$h_{t} = \lrn{W_m^\rT \Xt \lrp{V_k^\rT \Xt}^+}$,
\[
h_{t+1} - \frac{ 6 \lambda_k^{(\infty)} \Delta_1 + 15 \lambda_{m+1}^{(\infty)} \Delta_2 }{ \lambda_k^{(\infty)}-\lambda_{m+1}^{(\infty)} } \leq \frac{\lambda_{m+1}^{(\infty)}}{\lambda_k^{(\infty)}} \lrp{ h_t - \frac{ 6 \lambda_k^{(\infty)} \Delta_1 + 15 \lambda_{m+1}^{(\infty)} \Delta_2 }{ \lambda_k^{(\infty)}-\lambda_{m+1}^{(\infty)}} }.
\]
\end{lemma}
Expanding the recursion, we obtain that 
\[
h_t \leq \lrp{\frac{\lambda_{m+1}^{(\infty)}}{\lambda_k^{(\infty)}}}^{t} h_0 + \frac{ 6 \lambda_k^{(\infty)} \Delta_1 + 15 \lambda_{m+1}^{(\infty)} \Delta_2 }{ \lambda_k^{(\infty)}-\lambda_{m+1}^{(\infty)} }.
\]

We now develop the conditions that will satisfy the premise of \Cref{lemma:cos_iter} with a uniform bound on $\sigma_{\min}\lrp{V_k^\rT \Xt}$ by the initial condition.
\begin{lemma}
\label{lem:init}
For $\lrn{\Gt} \leq \eta \lambda_k^{(\infty)}$ and for $\lrn{V_k^\rT \Gt} \leq \frac{1}{10} \eta \lambda_k^{(\infty)} \cdot \sigma_{\min}\lrp{V_k^\rT \Xt}$,
$$\sigma_{\min}\lrp{V_k^\rT \Xt} \geq \frac{1}{6} \sigma_{\min}\lrp{V_k^\rT U_k^{(0)}},$$ 
for all $t\leq 1/\eta$.
\end{lemma}
Applying this to \Cref{lemma:cos_iter}, we achieve the result that when $\Delta_1 = \mathcal{O}\lrp{ \frac{\epsilon}{L} }$ and $\Delta_2 = \mathcal{O}\lrp{ \frac{\epsilon}{L} }$,
and when $T = \Theta\lrp{ \frac{L}{\epsilon} \log\frac{1}{\sigma_{\min}\lrp{V_k^\rT U_k^{(0)}}}}$,
\[
\lrn{W_m^\rT U_k^{(T)}} \leq \frac{h_T}{\sigma_{\min}\lrp{V_k^\rT U_k^{(T)}}} \leq \frac{\epsilon}{\lambda_k^{(\infty)} - \lambda_{m+1}^{(\infty)}},
\]
with $W_m={U}^{(\infty)}_{-m-1}$, for any $m\geq k$.
This leads to the requirement that for $\epsilon\leq\lambda_k^{(\infty)}$,
\begin{align}
\lrn{\Gt} = \mathcal{O} \lrp{\epsilon \cdot \frac{\alpha}{L} \cdot \sigma_{\min}\lrp{V_k^\rT U_k^{(0)}}},
\qquad
\lrn{V_k^\rT \Gt} 
= \mathcal{O} \lrp{\epsilon \cdot \frac{\alpha}{L} \cdot \sigma_{\min}^2\lrp{V_k^\rT U_k^{(0)}}},
\label{eq:requirement_SG_bound}
\end{align}
since $\lambda_k^{(\infty)}\geq\alpha$, $\forall k$.

We then bound the stochastic gradient noise $\Gt$ to fulfill this requirement.
\begin{lemma}
\label{lem:sg_bound}
For $\theta_j^{(t)} \sim \mathcal{N}\big( 0,\Omega^{-1} \big)$, and for orthogonal matrix $V_k\in\real^{d\times k}$, the following bounds hold for $L$-Lipschitz smooth function $\psi$ with probability $1-\mathcal{O}\lrp{1/N^2}$, for any $k\leq p$:
\[
\lrn{\Gt} \lesssim L \sqrt{\kappa(\Omega)} \cdot \sqrt{\frac{pd \log^4(N) \log(d)}{N}} + L \sqrt{\kappa(\Omega)} \cdot \frac{\sqrt{pd}}{N^2},
\]
and
\[
\lrn{V_k^\rT \Gt}  \lesssim L \sqrt{\kappa(\Omega)} \cdot \sqrt{\frac{p^2 \log^4(N) \log(d)}{N}} + L \sqrt{\kappa(\Omega)} \cdot \frac{p}{N^2},
\]
for any $k\leq p$.
Here $\kappa(\Omega) = \sigma_{\max}(\Omega) / \sigma_{\min}(\Omega)$ is the condition number of the positive definite input matrix $\Omega$.
\end{lemma}

Applying \Cref{lem:sg_bound} to the requirements in~\eqref{eq:requirement_SG_bound}, we establish that when 
$$N=\widetilde{\Omega}\lrp{ \max\lrbb{ \frac{pd}{ \sigma_{\min}^2\lrp{V_k^\rT U_k^{(0)}} } , \frac{p^2}{ \sigma_{\min}^4\lrp{V_k^\rT U_k^{(0)}} } } \cdot \frac{L^4}{\alpha^2} \cdot \frac{\kappa(\Omega)}{\epsilon^2 \sqrt{\delta} } },$$ 
and when 
$T = \Theta\lrp{ \frac{L}{\epsilon} \log\frac{1}{\sigma_{\min}\lrp{V_k^\rT U_k^{(0)}}}}$,
the following holds with probability $1-\delta$ for any $k\leq p$:
\[
\lrn{ \lrp{{U}^{(\infty)}_{-m-1}}^\rT U_k^{(T)} } = \lrn{W_m^\rT U_k^{(T)}} \leq \frac{\epsilon}{\lambda_k^{(\infty)}-\lambda_{m+1}^{(\infty)}}.
\]
Plugging in the following Weyl's inequality for the initial condition yields the final result
\[
1 - \sigma_{\min}\lrp{V_k^\rT U_k^{(0)}} 
\leq 1 - \sigma_{\min}\lrp{ (U^{(\infty)}_p)^\rT U^{(0)} }
\leq \lrn{\mI-\lrp{U^{(\infty)}_p}^\rT U^{(0)}} 
\leq \lrn{U^{(\infty)}_p - U^{(0)}}.
\]
\end{proof}

\subsubsection{Supporting Proofs}

\begin{proof}[Proof of \Cref{lemma:cos_iter}]
We first note that the update of the lower ranked $\Xt$ is the same as the update of $U^{(t)}$ in equations~\eqref{eq:mult_step_Gauss} and~\eqref{eq:QR_step_Gauss} because of the properties of matrix multiplication and QR factorization.
Writing $\matrixS = \Ep{\theta\sim \mathcal{N} \lrp{ 0, \Omega^{-1} }}{\nabla \psi(\theta) \theta^\rT \Omega}$, the update rule for $\Xt$ can be expressed as the mean update plus stochastic noise: $\widetilde{U}_k^{(t+1)} = \matrixS \cdot \Xt + \Gt$,
$\Xtp = \mathrm{QR} \lrp{\widetilde{U}_k^{(t+1)}}$.

We then use this update rule to develop our object $W_m^\rT \Xtp \lrp{V_k^\rT \Xtp}^+$, where we define the pseudo-inverse: $Z^+=Z^\rT \lrp{Z Z^\rT}^{-1}$.
Under this definition, 
\[V_k^\rT \lrp{\matrixS \cdot \Xt + \Gt} \lrp{R^{(t+1)}}^{-1} R^{(t+1)} \lrp{V_k^\rT \lrp{\matrixS \cdot \Xt + \Gt}}^+ = \mI.\]
Hence $$\lrp{ V_k^\rT \Xtp }^+ = \lrp{ V_k^\rT \lrp{\matrixS \cdot \Xt + \Gt} \lrp{R^{(t+1)}}^{-1} }^+ = R^{(t+1)} \lrp{V_k^\rT \lrp{\matrixS \cdot \Xt + \Gt}}^+.$$
Expanding $W_m^\rT \Xtp \lrp{V_k^\rT \Xtp}^+$ using this result, we obtain that
\[
W_m^\rT \Xtp \lrp{V_k^\rT \Xtp}^+ = \lrp{ \Lambda_{-m}^{(\infty)} W_m^\rT \Xt + W_m^\rT \Gt } \lrp{ V_k^\rT \Xt + (\Lambda_k^{(\infty)})^{-1} V_k^\rT \Gt }^+ (\Lambda_k^{(\infty)})^{-1}.
\]
Denote $\Zt = V_k^\rT \Xt + (\Lambda_k^{(\infty)})^{-1} V_k^\rT \Gt$.
Then we can upper bound $h_{t+1}$ as follows.
\begin{align*}
h_{t+1} &= \lrn{ \lrp{ \Lambda_{-m}^{(\infty)} W_m^\rT \Xt + W_m^\rT \Gt } \lrp{ \Zt }^+ (\Lambda_k^{(\infty)})^{-1} } \\
&\leq \underbrace{\lrn{ \Lambda_{-m}^{(\infty)} W_m^\rT \Xt \lrp{ \Zt }^+ (\Lambda_k^{(\infty)})^{-1} }}_{T_1}
+ \underbrace{\lrn{ W_m^\rT \Gt \lrp{ \Zt }^+ (\Lambda_k^{(\infty)})^{-1} }}_{T_2}.
\end{align*}

For the pseudo-inverse, $\lrp{ \Zt }^+ = \lrp{\Zt}^\rT \lrp{\Zt \lrp{\Zt}^\rT}^{-1}$, we use the Woodbury formula and develop its inverse part as follows:
\begin{align}
\lrp{\Zt \lrp{\Zt}^\rT}^{-1}
= \tP^{-\rT} \lrp{\mI-\lrp{\mI+Y}^{-1}Y} \tP^{-1},
\label{eq:Woodbury}
\end{align}
where $\tP \widetilde{V}^\rT = V_k^\rT \Xt$ for a unitary matrix $\widetilde{V}$, and
\[
Y = \widetilde{V}^\rT \Gt^\rT V_k (\Lambda_k^{(\infty)})^{-1} \tP^{-\rT} + \tP^{-1} (\Lambda_k^{(\infty)})^{-1} V_k^\rT \Gt \widetilde{V} + \widetilde{P}^{-1} (\Lambda_k^{(\infty)})^{-1} V_k^\rT \Gt \Gt^\rT V_k (\Lambda_k^{(\infty)})^{-1} \tP^{-\rT},
\]
as defined in equation~\eqref{eq:Y_def} in \Cref{fact:Y_bound}.
\Cref{fact:Y_bound} will also provide a bound on $\lrn{ \lrp{\mI+Y}^{-1}Y }$.

\sloppy
We now use equation~\eqref{eq:Woodbury} and the definition of $\Zt$ to expand term $T_1$.
Since we have $\lrp{\Xt}^\rT V_k \widetilde{P}^{-\rT} \widetilde{P}^{-1} = \lrp{V_k^\rT \Xt}^+$ and  $\lrp{\Xt}^\rT V_k \widetilde{P}^{-\rT} = \widetilde{V}$, we obtain that
\begin{align*}
T_1 &\leq
\lrn{ \Lambda_{-m}^{(\infty)} W_m^\rT \Xt \lrp{ \lrp{V_k^\rT \Xt}^+ + \lrp{\Gt}^\rT V_k (\Lambda_k^{(\infty)})^{-1} \widetilde{P}^{-\rT} \widetilde{P}^{-1}  } (\Lambda_k^{(\infty)})^{-1} } \\
&+ \lrn{ \Lambda_{-m}^{(\infty)} W_m^\rT \Xt \lrp{ \widetilde{V} \lrp{\mI+Y}^{-1} Y \widetilde{P}^{-1} + \lrp{\Gt}^\rT V_k (\Lambda_k^{(\infty)})^{-1} \widetilde{P}^{-\rT} \lrp{\mI+Y}^{-1} Y \widetilde{P}^{-1} } (\Lambda_k^{(\infty)})^{-1} } \\
&\stackrel{\1}{\leq} \frac{\lambda_{m+1}^{(\infty)}}{\lambda_k^{(\infty)}} \lrp{h_t + \Delta_2 + \frac{\frac{7}{3}\Delta_2}{1-\frac{7}{3}\Delta_2} + \frac{\frac{7}{3}\Delta_2^2}{1-\frac{7}{3}\Delta_2} }
\leq \frac{\lambda_{m+1}^{(\infty)}}{\lambda_k^{(\infty)}} \lrp{h_t + 15\Delta_2},
\end{align*}
where $\1$ follows from \Cref{fact:Y_bound}.
Similarly, by the bound on $\lrn{\Gt}$ and \Cref{fact:Y_bound},
\begin{align*}
T_2 &\leq \lrn{ W_m^\rT \Gt \lrp{ \lrp{V_k^\rT \Xt}^+ + \lrp{\Gt}^\rT V_k (\Lambda_k^{(\infty)})^{-1} \tP^{-\rT} \tP^{-1} } (\Lambda_k^{(\infty)})^{-1} } \\
&+ \lrn{ W_m^\rT \Gt \lrp{ \widetilde{V} (\mI+Y)^{-1} Y \tP^{-1} + \lrp{\Gt}^\rT V_k (\Lambda_k^{(\infty)})^{-1} \tP^{-\rT} (\mI+Y)^{-1} Y \tP^{-1} } (\Lambda_k^{(\infty)})^{-1} } \\
&\leq \Delta_1 + \Delta_1 \Delta_2 + \Delta_1 \cdot \frac{\frac{7}{3}\Delta_2}{1-\frac{7}{3}\Delta_2} + \Delta_1 \cdot \Delta_2 \cdot \frac{\frac{7}{3}\Delta_2}{1-\frac{7}{3}\Delta_2}
\leq 6 \Delta_1.
\end{align*}
Combining the two bounds, we arrive at our result that
\[
h_{t+1} \leq \frac{\lambda_{m+1}^{(\infty)}}{\lambda_k^{(\infty)}} h_t + \lrp{ 15 \frac{\lambda_{m+1}^{(\infty)}}{\lambda_k^{(\infty)}} \Delta_2 + 6 \Delta_1 }.
\]
Rearranging the terms yields the final expression.
\end{proof}

\begin{fact}
\label{fact:Y_bound}
Define 
\begin{multline}
Y = \widetilde{V}^\rT \lrp{\Gt}^\rT V_k (\Lambda_k^{(\infty)})^{-1} \tP^{-\rT} + \tP^{-1} (\Lambda_k^{(\infty)})^{-1} V_k^\rT \Gt \widetilde{V} \\+ \widetilde{P}^{-1} (\Lambda_k^{(\infty)})^{-1} V_k^\rT \Gt \lrp{\Gt}^\rT V_k (\Lambda_k^{(\infty)})^{-1} \tP^{-\rT},
\label{eq:Y_def}
\end{multline}
for a unitary matrix $\widetilde{V}\in\real^{d\times d}$, for $\Lambda_k^{(\infty)}$ containing eigenvalues bigger than or equal to $\lambda_k^{(\infty)}$, and for $\tP \widetilde{V}^\rT = V_k^\rT \Xt$.
Then when $\lrn{V_k^\rT \Gt} \leq \Delta_2 \cdot \lambda_k^{(\infty)} \cdot \sigma^2_{\min}\lrp{V_k^\rT \Xt}$ for $\Delta_2\leq1/3$,
\[
\lrn{ \lrp{\Gt}^\rT V_k (\Lambda_k^{(\infty)})^{-1} \tP^{-\rT} } \leq \Delta_2 \cdot \sigma_{\min}\lrp{V_k^\rT \Xt},
\]
and that 
\[
\lrn{ \lrp{\mI+Y}^{-1}Y } \leq \frac{\frac{7}{3}\Delta_2}{1-\frac{7}{3}\Delta_2} \cdot \sigma_{\min}\lrp{V_k^\rT \Xt}.
\]
\end{fact}
\begin{proof}
First note that for $\tP \widetilde{V}^\rT = V_k^\rT \Xt$, $\|\tP^{-1}\|_2 \leq 1/\sigma_{\min}\lrp{V_k^\rT \Xt}$.
Then
\[
\lrn{ \lrp{\Gt}^\rT V_k (\Lambda_k^{(\infty)})^{-1} \tP^{-\rT} } 
\leq \lrn{V_k^\rT \Gt} \cdot \lrn{(\Lambda_k^{(\infty)})^{-1}}_2 \cdot \lrn{\tP^{-1}}_2.
\]
Plugging in the assumption yields the bound.

Similarly,
\[
\lrn{ \tP^{-1} (\Lambda_k^{(\infty)})^{-1} V_k^\rT \Gt \widetilde{V} } 
\leq \lrn{V_k^\rT \Gt} \cdot \lrn{(\Lambda_k^{(\infty)})^{-1}}_2 \cdot \lrn{\tP^{-1}}_2
\leq \Delta_2 \cdot \sigma_{\min}\lrp{V_k^\rT \Xt},
\]
and 
\begin{align*}
\lrn{ \widetilde{P}^{-1} (\Lambda_k^{(\infty)})^{-1} V_k^\rT \Gt \lrp{\Gt}^\rT V_k (\Lambda_k^{(\infty)})^{-1} \tP^{-\rT} }
&\leq \lrn{V_k^\rT \Gt}^2 \cdot \lrn{(\Lambda_k^{(\infty)})^{-1}}_2^2 \cdot \lrn{\tP^{-1}}_2^2 \\
&\leq \Delta_2^2 \cdot \sigma^2_{\min}\lrp{V_k^\rT \Xt}
\leq \frac{1}{3} \Delta_2 \cdot \sigma_{\min}\lrp{V_k^\rT \Xt}.
\end{align*}

Combining the three terms, we obtain that
\[
\lrn{Y} \leq \frac{7}{3} \Delta_2 \cdot \sigma_{\min}\lrp{V_k^\rT \Xt},
\]
and that 
\[
\lrn{(\mI+Y)^{-1}Y}
\leq \frac{\lrn{Y}}{1-\lrn{Y}_2} \leq \frac{\frac{7}{3}\Delta_2}{1-\frac{7}{3}\Delta_2} \cdot \sigma_{\min}\lrp{V_k^\rT \Xt}.
\]
\end{proof}

\begin{proof}[Proof of \Cref{lem:init}]
Following~\cite{hardt2014}, we can write $\cos \theta_k(V_k, \Xt) = \sigma_{\min}\lrp{V_k^\rT \Xt}$ and $\tan \theta_k(V_k, \Xt) = \lrn{\lrp{V_k^{\bot}}^\rT \Xt \lrp{V_k^\rT \Xt}^+}$, for $V_k^{\bot} = \bar{U}^*_{-k-1}$.
From Lemma A.5 of~\cite{Balcan2016}, we know that for $\lrn{\Gt} \leq c_1 \eta \lambda_k^{(\infty)}$ and for $\lrn{V_k^\rT \Gt} \leq c_2 \eta \lambda_k^{(\infty)} \cdot \sigma_{\min}\lrp{V_k^\rT \Xt}$, for all $t=0,1,\dots,T$,
\begin{align*}
\tan \theta_k(V_k, \Xt) + \frac{c_1}{c_1+c_2} 
&\leq \lrp{\frac{1+c_1\eta}{1-c_2\eta}}^{t} \lrp{ \tan \theta_k(V_k, U_k^{(0)}) + \frac{c_1}{c_1+c_2} } \\
&\leq \exp\lrp{\frac{(c_1+c_2)\eta}{1-c_2\eta}t} \lrp{ \tan \theta_k(V_k, U_k^{(0)}) + \frac{c_1}{c_1+c_2} }.
\end{align*}
Taking $c_1=1$ and $c_2=1/10$, as well as the range of $t$, $T \leq 1/\eta$ (where $\eta\leq1$) ensures
\[
\tan \theta_k(V_k, \Xt) \leq 4 \tan \theta_k(V_k, U_k^{(0)}) + 3.
\]
Therefore, $$\sigma_{\min}\lrp{V_k^\rT \Xt} = \cos \theta_k (V_k, \Xt) \geq \frac{1}{6} \cos \theta_k(V_k, U_k^{(0)}) = \frac{1}{6} \sigma_{\min}\lrp{V_k^\rT U_k^{(0)}},$$ 
for all $t\leq 1/\eta$.
\end{proof}

\begin{proof}[Proof of \Cref{lem:sg_bound}]
Similar to~\cite{hardt2014}, we use matrix Chernoff bound to establish this noise bound.
Since 
$\theta_j^{(t)} \sim \mathcal{N}\lrp{0,\Omega^{-1}}$,
we know that 
\[
\mathbb{P}\lrbb{ \lrn{\theta} \geq \frac{\sqrt{d}}{\sqrt{\sigma_{\min}(\Omega)}} t } \leq e^{-t^2/2}.
\]
Since $\psi$ is $L$-Lipschitz smooth, we obtain that 
\[
\mathbb{P}\lrbb{ \lrn{\nabla \psi(\theta)} \geq \frac{L\sqrt{d}}{\sqrt{\sigma_{\min}(\Omega)}} t } \leq e^{-t^2/2}.
\]
For the orthogonal matrix $\Xt\in\real^{d \times k}$,
\[
\mathbb{P}\lrbb{ \lrn{\lrp{\Xt}^\rT \Omega \theta} \geq \sqrt{k} \sqrt{\sigma_{\max}(\Omega)} \cdot t } \leq e^{-t^2/2}.
\]

Therefore, for $k\leq p$, with high probability, 
\[
\lrn{\nabla \psi(\theta_j) \theta_j^\rT \Omega \Xt} \leq \lrn{\nabla \psi(\theta_j)} \lrn{\lrp{\Xt}^\rT \Omega \theta_j} 
\lesssim L \sqrt{\kappa(\Omega)} \cdot \sqrt{kd} \leq L \sqrt{\kappa(\Omega)} \cdot \sqrt{pd},
\]
and that 
\[
\lrn{V_k^\rT \nabla \psi(\theta_j) \theta_j^\rT \Omega \Xt} \leq \lrn{V_k^\rT \nabla \psi(\theta_j)} \lrn{\lrp{\Xt}^\rT \Omega \theta_j} \lesssim L \sqrt{\kappa(\Omega)} \cdot k \leq L \sqrt{\kappa(\Omega)} \cdot p.
\]

We then invoke the matrix Chernoff bound (c.f. Lemma 3.5 of~\cite{hardt2014}) to obtain that with probability $1-\mathcal{O}\lrp{1/N^2}$, for any $k\leq p$,
\[
\lrn{\Gt} \lesssim L \sqrt{\kappa(\Omega)} \cdot \sqrt{\frac{pd \log^4(N) \log(d)}{N}} + L \sqrt{\kappa(\Omega)} \cdot \frac{\sqrt{pd}}{N^2},
\]
and that
\[
\lrn{V_k^\rT \Gt}  \lesssim L \sqrt{\kappa(\Omega)} \cdot \sqrt{\frac{p^2 \log^4(N) \log(d)}{N}} + L \sqrt{\kappa(\Omega)} \cdot \frac{p}{N^2}.
\]
\end{proof}


\subsection{Proofs for solving for $\Lambda$}
\label{sec:pf_Lambda}
\paragraph{Computation}
\begin{proof}[Proof of \Cref{fact:lambda}]
We can first explicitly compute that 
\begin{align*}
\Ep{\theta\sim q}{ \nabla_{\lambda_i} \log q_{(U,\Lambda)}(\theta|\vx) \log{q_{(U,\Lambda)}(\theta|\vx)}  }
= \frac{1}{2} u_i^\rT \Omega^{-1} u_i
= \frac{1}{2} \lrp{D_{i,i} + \lambda_i}^{-1}.
\end{align*}
\sloppy
To solve for $\lambda_i$, we can directly set the gradient equal to zero, i.e.  $\nabla_{\lambda_i} \KL{q}{p}
= \Ep{\theta\sim q}{ \nabla_{\lambda_i} \log q_{(U,\Lambda)}(\theta|\mathbf{x}) \log\frac{q_{(U,\Lambda)}(\theta|\mathbf{x})}{p(\theta|\mathbf{x})}  } =0 $, and obtain:
\begin{align}
\frac{1}{2} \lrp{D_{i,i} + \lambda_i}^{-1} 
&= \Ep{\theta\sim q}{ \nabla_{\lambda_i} \log q_{(U,\Lambda)}(\theta|\mathbf{x}) \log{q_{(U,\Lambda)}(\theta|\mathbf{x})}  } \nonumber\\
&= \Ep{\theta\sim q}{ \nabla_{\lambda_i} \log q_{(U,\Lambda)}(\theta|\mathbf{x}) \log p(\theta|\mathbf{x})  } \nonumber\\
&= - \Ep{\theta\sim q}{ \nabla_{\lambda_i} \log q_{(U,\Lambda)}(\theta|\mathbf{x}) \cdot \psi(\theta)  } \nonumber\\
&= \Ep{\theta\sim q}{ \lrp{ \frac{1}{2} u_i^\rT \theta \theta^\rT u_i - \frac{1}{2} u_i^\rT \Omega^{-1} u_i } \cdot \psi(\theta)  }. \label{eq:zero_grad_KL}
\end{align}
Also note that $\nabla_{\theta} \log q_{(U,\Lambda)}(\theta|\mathbf{x}) = - \Omega \theta$, which leads to the relationship that $u_i^\rT \theta = - u_i^\rT \Omega^{-1} \nabla_{\theta} \log q = - (D_{i,i}+\lambda_i)^{-1} u_i^\rT \nabla_{\theta} \log q$.
We can use this fact to transform the above equation:
\begin{align*}
&\Ep{\theta\sim q}{ \lrp{ \frac{1}{2} u_i^\rT \theta \theta^\rT u_i } \cdot \psi(\theta)  } \\
&= \frac{1}{2} (D_{i,i}+\lambda_i)^{-2} \cdot \Ep{\theta\sim q}{ \lrp{ u_i^\rT \nabla_{\theta} \log q \ \nabla_{\theta}^\rT \log q \ u_i } \cdot \psi(\theta)  } \\
&= - \frac{1}{2} (D_{i,i}+\lambda_i)^{-2} \lrp{ \int u_i^\rT \nabla_{\theta}^2 \log q \ u_i \cdot \psi(\theta) \ q \ \rd \theta 
+ \int u_i^\rT \nabla_{\theta}\psi(\theta) \nabla_{\theta}^\rT \log q \ u_i \cdot q\ \rd \theta } \\
&= \frac{1}{2} (D_{i,i}+\lambda_i)^{-2} \int u_i^\rT \Omega \ u_i \cdot \psi(\theta) \ q \ \rd \theta 
+ \frac{1}{2} (D_{i,i}+\lambda_i)^{-2} \int u_i^\rT \nabla_{\theta}^2 \psi(\theta) \ u_i \cdot q\ \rd \theta \\
&= \frac{1}{2} \Ep{\theta\sim q}{ u_i^\rT \Omega^{-1} u_i \cdot \psi(\theta)  }
+ \frac{1}{2} (D_{i,i}+\lambda_i)^{-2} \Ep{\theta\sim q}{ u_i^\rT \nabla_{\theta}^2 \psi(\theta) u_i }.
\end{align*}
Plugging this into the zero-gradient condition in equation~\eqref{eq:zero_grad_KL}, we obtain the following condition:
\[
\lambda_i = \Ep{\theta\sim q}{ u_i^\rT \nabla_{\theta}^2 \psi(\theta) u_i } - D_{i,i}.
\]
\end{proof}

\begin{proof}[Proof of \Cref{lem:lambda}]
To bound $\sum_{k=1}^p \lrp{\lambda^{(\infty)}_k - \lambda^{(T)}_k}^2$, where $$\lambda^{(T)}_k = \frac{1}{M} \sum_{j=1}^M \lrp{ u_k^{(T)} }^\rT \lrp{ \frac{\nabla \psi(\theta_j+\Delta \cdot u_k^{(T)}) - \nabla \psi(\theta_j-\Delta \cdot u_k^{(T)})}{2\Delta} } - \alpha,$$ we use Young's inequality to separate it into two terms:
\begin{align*}
\sum_{k=1}^p \lrp{\lambda^{(\infty)}_k - \lambda^{(T)}_k}^2
&\leq \underbrace{ \sum_{k=1}^p 2 \lrp{ \lambda^{(T)}_k + \alpha - \Ep{\theta\sim q}{ \lrp{u_k^{(T)}}^\rT \nabla_{\theta}^2 \psi(\theta) \lrp{u_k^{(T)}} } }^2 }_{\1} \\
&+ \underbrace{ \sum_{k=1}^p 2 \lrp{ \lambda^{(\infty)}_k + \alpha - \Ep{\theta\sim q}{ \lrp{u_k^{(T)}}^\rT \nabla_{\theta}^2 \psi(\theta) \lrp{u_k^{(T)}} } }^2 }_{\2}.
\end{align*}

We first bound term $\1$.
We start by proving that the stochastic estimate $\frac{1}{M} \sum_{j=1}^M u_i^\rT \nabla^2\psi(\theta_j) u_i$ is close to the true mean $\Ep{\theta\sim q}{ u_i^\rT \nabla_{\theta}^2 \psi(\theta) u_i }$.


\begin{lemma}
\label{lem:sg_noise_lambda}
Assume that there exists a positive definite $\Psi$ such that $- \Psi \preceq \nabla^2 \psi(\theta) \preceq \Psi, \forall \theta\in\real^d$.
We take number of samples $M = \frac{2 p \log(2 p/\delta)}{\epsilon^2} \sigma_{\max}^2 \lrp{\Psi}$ in \Cref{alg:General}. We obtain that for i.i.d. $\theta_j\sim q_{(U,\Lambda)}(\theta|\mathbf{x})$, 
$$\left| \frac{1}{M} \sum_{j=1}^M u_i^\rT \nabla^2\psi(\theta_j) u_i - \Ep{\theta\sim q}{ u_i^\rT \nabla_{\theta}^2 \psi(\theta) u_i } \right| \leq \epsilon/\sqrt{p}, \qquad \forall u_i \in \mathbb{R}^d, \lrn{u}_i\leq1,$$ 
$\forall i=1,\dots,p$ with probability $1-\delta$.
\end{lemma}
Since function $\psi(\theta)$ is $L$-Lipschitz smooth, $\sigma_{\max}(\Psi) = L$.

We then prove that the gradient difference $ \frac{\nabla \psi(\theta+\Delta \cdot u_i) - \nabla \psi(\theta-\Delta \cdot u_i)}{2\Delta} $ in \Cref{alg:General} accurately estimates the Hessian vector product $\nabla^2 \psi(\theta) \cdot u_i$.
\begin{fact}
\label{fact:Hess_estimate}
Assume the Hessian Lipschitz condition that $\lrn{\nabla^2 \psi(\theta) - \nabla^2 \psi(\theta')}_2 \leq L_{\rm{Hess}} \lrn{\theta-\theta'}, \forall \theta, \theta' \in\real^d$. 
Taking $\varDelta \leq \epsilon/\lrp{ \sqrt{p} \cdot L_{\rm{Hess}} }$ in \Cref{alg:General}, the Hessian vector product $\nabla^2\psi(\theta) \cdot u_i$ can be estimated to $\epsilon$ accuracy by the gradient difference:
$$\lrn{ \frac{ \nabla \psi(\theta + \varDelta u_i) - \nabla \psi(\theta - \varDelta u_i) }{2\varDelta} - \nabla^2 \psi(\theta) \cdot u_i } \leq \epsilon/\sqrt{p},$$
for any $\theta \in \mathbb{R}^d$ and $u_i \in \mathbb{R}^d$ and $\lrn{u}_i\leq1$.
\end{fact}
\Cref{fact:Hess_estimate} leads to the bound that for $\lambda_{k}^{(T)} + \alpha = \frac{1}{M} \sum_{j=1}^M \lrp{ u_k^{(T)} }^\rT \lrp{ \frac{\nabla \psi(\theta_j+\Delta \cdot u_k^{(T)}) - \nabla \psi(\theta_j-\Delta \cdot u_k^{(T)})}{2\Delta} }$,
$$\left| \lambda_{k}^{(T)} + \alpha - \frac{1}{M} \sum_{j=1}^M \lrp{u_k^{(T)}}^\rT \nabla^2\psi(\theta_j) u_k^{(T)} \right| \leq \epsilon/\sqrt{p},$$
since $\lrn{u_k^{(T)}} \leq 1$.

Combining \Cref{fact:Hess_estimate} and \Cref{lem:sg_noise_lambda}, we obtain that 
$\lrp{ \lambda_{k}^{(T)} + \alpha - \Ep{\theta\sim q}{ \lrp{u_k^{(T)}}^\rT \nabla_{\theta}^2 \psi(\theta) \lrp{u_k^{(T)}} } } \leq 4 \epsilon^2/p$, $\forall k=1,\dots,p$ with probability $1-\delta$, via $M = 2 p \log(2p/\delta) \cdot \frac{L^2}{\epsilon^2}$ gradient differences of $\varDelta \leq \epsilon/\lrp{ \sqrt{p} \cdot L_{\rm{Hess}} }$ apart.
This leads to the bound for term $\1$ that 
\begin{align}
\sum_{k=1}^p 2 \lrp{ \lambda^{(T)}_k + \alpha - \Ep{\theta\sim q}{ \lrp{u_k^{(T)}}^\rT \nabla_{\theta}^2 \psi(\theta) \lrp{u_k^{(T)}} } }^2 
\leq 8 \epsilon^2.
\label{eq:eig_bound_T1}
\end{align}

We then use \Cref{lem:Oja_convergence} to bound term $\2$.
%
%
When $p(\theta|\vx)$ is a normal distribution, $\Ep{\theta\sim q}{ \nabla_{\theta}^2 \psi(\theta) } = \nabla_{\theta}^2 \psi(\theta) = \Omega^{(\infty)}$.
Plugging into term $\2$ leads to the expression: $\sum_{k=1}^p 2 \lrp{ \lambda_{k}^{(\infty)} - \widetilde{\lambda}_k^{(T),(\infty)} }^2$,
where we denote $\widetilde{\lambda}_k^{(T),(\infty)} = \lrp{u_k^{(T)}}^\rT \Omega^{(\infty)} u_k^{(T)} - \alpha
= \lrp{u_k^{(T)}}^\rT \lrp{\Omega^{(\infty)} - \alpha\mI} u_k^{(T)}$.

From \Cref{lem:Oja_convergence}, we know that with number of stochastic gradient samples per iteration $N$ and
and number of iterations $T$ defined therein, $\widetilde{\lambda}_k^{(T),(\infty)} + \alpha = \lrp{u^{(T)}_k}^\rT \Omega^{(\infty)} u^{(T)}_k \geq \lambda^{(\infty)}_k + \alpha - 2\epsilon$,
$\forall k\in\{1,\dots,p\}$ with probability $1-\delta$.
Consequently,
\begin{align*}
\sum_{k=1}^p \lrp{ \lambda^{(\infty)}_k - \widetilde{\lambda}_k^{(T),(\infty)} }^2
&= \sum_{k=1}^p \lrp{\lambda^{(\infty)}_k}^2 + \sum_{k=1}^p \lrp{\widetilde{\lambda}_k^{(T),(\infty)}}^2 - 2 \sum_{k=1}^p \lambda_k^{(\infty)} \widetilde{\lambda}_k^{(T),(\infty)} \\
&\leq \sum_{k=1}^p \lrp{\lambda^{(\infty)}_k}^2 + \sum_{k=1}^p \lrp{\widetilde{\lambda}_k^{(T),(\infty)}}^2 - 2 \sum_{k=1}^p \lrp{\lambda_k^{(\infty)}}^2 + 4 \epsilon \sum_{k=1}^p \lambda_k^{(\infty)}.
\end{align*}
On the other hand, $\sum_{k=1}^p \lrp{ \widetilde{\lambda}_k^{(T),(\infty)} }^2 = \tr\lrp{\lrp{U^{(T)}}^\rT \lrp{\Omega^{(\infty)} - \alpha\mI}^2 \lrp{U^{(T)}}} \leq \tr\lrp{\lrp{\Omega^{(\infty)} - \alpha\mI}^2} = \tr\lrp{\lrp{U^{(\infty)} \Lambda^{(\infty)} \lrp{U^{(\infty)}}^\rT }^2} = \sum_{k=1}^d \lrp{\lambda^{(\infty)}_k}^2$, and therefore
\begin{align}
\sum_{k=1}^p 2 \lrp{ \lambda^{(\infty)}_k - \widetilde{\lambda}_k^{(T),(\infty)} }^2
\leq 2 \sum_{k=p+1}^d \lrp{\lambda^{(\infty)}_k}^2 + 8 \epsilon \sum_{k=1}^p \lambda^{(\infty)}_k.
\label{eq:eig_bound_T2}
\end{align}
Combining equations~\eqref{eq:eig_bound_T1} and~\eqref{eq:eig_bound_T2} leads to the final result that
\[
\sum_{k=1}^p \lrp{\lambda^{(\infty)}_k - \lambda^{(T)}_k}^2 
\leq 2 \sum_{k=p+1}^d \lrp{\lambda^{(\infty)}_k}^2 + 8 \epsilon \sum_{k=1}^p \lambda^{(\infty)}_k + 8 \epsilon^2,
\]
with probability $1-2\delta$.
\end{proof}

\subsubsection{Supporting Proofs}
\begin{proof}[Proof of \Cref{lem:sg_noise_lambda}]
Denote $u_i^\rT \widetilde{\Omega}(q) u_i = \frac{1}{M} \sum_{j=1}^M u_i^\rT \nabla^2\psi(\theta_j) u_i$, and $u_i^\rT \Omega(q) u_i = \Ep{\theta\sim q}{ u_i^\rT \nabla_{\theta}^2 \psi(\theta) u_i }$, for i.i.d. $\theta_j\sim q_{(U,\Lambda)}(\theta|\mathbf{x})$.

From the Hoeffding's inequality, we know that 
\begin{align*}
\mathbb{P}\lrp{ \left| u_i^\rT \widetilde{\Omega}(q) u_i - u_i^\rT \Omega(q) u_i \right| \geq \upsilon } \leq 2 \exp\lrp{ - \frac{M \cdot \upsilon^2 }{2 \cdot \sigma_{\max}^2 \lrp{\Psi} } },
\end{align*}
since $u_i^\rT \nabla^2\psi(\theta_j) u_i \in \lrb{-\sigma_{\max} \lrp{\Psi}, \sigma_{\max} \lrp{\Psi}}$.
This leads to the fact that 
\[
\left| u_i^\rT \widetilde{\Omega}(q) u_i - u_i^\rT \Omega(q) u_i \right| \leq \sqrt{\frac{ 2 \log(2/\delta) }{M}} \sigma_{\max} \lrp{\Psi},
\]
with probability $1-\delta$.

By the union bound, we know that with probability $1-\delta$,
\[
\left| u_i^\rT \widetilde{\Omega}(q) u_i - u_i^\rT \Omega(q) u_i \right| \leq \sqrt{\frac{ 2 \log(2p/\delta) }{M}} \sigma_{\max} \lrp{\Psi},
\forall i=1,\dots,p.
\]
Taking $M = \frac{2 p \log(2p/\delta)}{\epsilon^2} \sigma_{\max}^2 \lrp{\Psi}$, we obtain that $\left| u_i^\rT \widetilde{\Omega}(q) u_i - u_i^\rT \Omega(q) u_i \right| \leq \epsilon/\sqrt{p}, \ \forall i=1,\dots,p$ with probability $1-\delta$.
\end{proof}

\begin{proof}[Proof of \Cref{fact:Hess_estimate}]
By the mean value theorem,
\begin{align*}
\frac{ \nabla \psi(\theta + \varDelta \cdot u_i) - \nabla \psi(\theta - \varDelta \cdot u_i) }{2\varDelta}
= \frac{1}{2} \int_{-1}^1 \nabla^2 \psi(\theta+\eta \varDelta \cdot u_i) \rd \eta \cdot  u_i.
\end{align*}
Then by the Hessian Lipschitzness assumption on $\psi$, 
\[
\lrn{ \frac{1}{2} \int_{-1}^1 \nabla^2 \psi(\theta+\eta \varDelta \cdot u_i) \rd \eta - \nabla^2 \psi(\theta) }_2 \leq L_{\rm{Hess}} \varDelta \cdot \lrn{u_i} = L_{\rm{Hess}} \varDelta.
\]
We can therefore accurately estimate $\nabla^2\psi(\theta) \cdot u_i$ using two gradient evaluations by taking a small enough $\varDelta$:
\begin{align*}
&\lrn{ \frac{ \nabla \psi(\theta + \varDelta \cdot u_i) - \nabla \psi(\theta - \varDelta \cdot u_i) }{2\varDelta} - \nabla^2\psi(\theta) \cdot u_i }\\
&\leq \lrn{ \frac{1}{2} \int_{-1}^1 \nabla^2 \psi(\theta+\eta \varDelta \cdot u_i) \rd \eta \cdot u_i - \nabla^2\psi(\theta) \cdot u_i } \\
&\leq \lrn{ \frac{1}{2} \int_{-1}^1 \nabla^2 \psi(\theta+\eta \varDelta \cdot u_i) \rd \eta - \nabla^2\psi(\theta) }_2 \lrn{u_i} \\
&\leq L_{\rm{Hess}} \varDelta.
\end{align*}
\end{proof}

\section{Proof for the frequentist uncertainty quantification error}
\label{app:stat}
For any $T > 0$ and $\Omega^{(\infty)} = \sum_{i=1}^n x_ix_i^\top$, consider decomposing the error
\begin{align}
     \lrn{\Omega^* - \Omega^{(T)}}_F^2 &\leq 2 \lrn{\Omega^* - \Omega^{(\infty)} }_F^2 
     + 2 \lrn{ \Omega^{(\infty)} - \Omega^{(T)} }_F^2\nonumber\;,
\end{align}
where the first term corresponds to the statistical error and the second term corresponds to the optimization and approximation error that are bounded in \Cref{thm:KL_convergence}. 
\begin{proof}[Proof of \Cref{prop:stat}]
We start with bounding the first term in the above inequality.
Note that with vectorization, we can transform the matrix Frobenius norm to the vector $2$-norm:
\[ \lrn{M}_F = \lrn{ \mathrm{vec}\lrp{M} }. \]
Then for $\Omega^{(\infty)} = \sum_{i=1}^n x_i x_i^\top$ and $\Omega^* = \E{x x^\top}$, the first term can be written as
\begin{align*}
\lrn{\Omega^* - \frac{1}{n} \Omega^{(\infty)} }_F
= \lrn{ \E{ \mathrm{vec}\lrp{x x^\top} } - \frac{1}{n} \sum_{i=1}^n \mathrm{vec}\lrp{x_i x_i^\top} }.
\end{align*}
Then for any $\mathbf{v}\in \real^{d^2}, \lrn{\mathbf{v}}=1$, we know that
\[
\left| \lrw{\mathbf{v},\mathrm{vec}\lrp{x_i x_i^\top}} \right| 
\leq \lrn{ \mathrm{vec}\lrp{x_i x_i^\top} }
= \lrn{x_i x_i^\top}_F
= \lrn{x_i}^2
\leq R.
\]
We thus apply the Hoeffding's inequality on the sequence of $\lrw{\mathbf{v},\mathrm{vec}\lrp{x_1 x_1^\top}}, \dots, \lrw{\mathbf{v},\mathrm{vec}\lrp{x_n x_n^\top}}$ and obtain that with $1-\delta$ probability,
\[
\E{ \lrw{ \mathbf{v},\mathrm{vec}\lrp{x x^\top} } } - \frac{1}{n} \sum_{i=1}^n \lrw{ \mathbf{v},\mathrm{vec}\lrp{x_i x_i^\top} }
\leq 2 R \sqrt{\frac{1}{n}\log\frac{1}{\delta}}.
\]
Choosing $\mathbf{v} = \lrp{ \mathrm{vec}\lrp{\Omega^*} - \mathrm{vec}\lrp{\Omega^{(\infty)}} } / \lrn{\Omega^* - \Omega^{(\infty)}}$ leads to the result that 
\begin{align*}
\lrn{\Omega^* - \frac{1}{n} \Omega^{(\infty)} }_F
= \lrn{ \E{ \mathrm{vec}\lrp{x x^\top} } - \frac{1}{n} \sum_{i=1}^n \mathrm{vec}\lrp{x_i x_i^\top} }
\leq 2 R \sqrt{\frac{1}{n}\log\frac{1}{\delta}},
\end{align*}
with $1-\delta$ probability.

From \Cref{thm:KL_convergence}, we know that the second term is upper bounded with $1-\delta$ probability
\begin{align*}
\lrn{ \frac{1}{n} \Omega^{(\infty)} - \frac{1}{n} \Omega^{(T)}}_F^2
\leq \frac{3}{n^2} \cdot \sum_{k=p+1}^d \lrp{\lambda^{(\infty)}_k}^2
+ \frac{12}{n} \cdot \frac{\epsilon}{n} \cdot \sum_{k=1}^p \lambda^{(\infty)}_k + 8\frac{\epsilon^2}{n^2},
\end{align*}
if we take number of stochastic gradient samples per iteration $N = \widetilde{\boldsymbol{\Omega}}\lrp{ p d \cdot \frac{L^2}{\alpha^2} \cdot \frac{L^2}{\epsilon^2 \sqrt{\delta} } }$, the number of iterations $T = \boldsymbol{\Theta}\lrp{ \frac{L}{\epsilon} }$, as well as the number of samples for eigenvalue computation $n = {\boldsymbol{\Omega}}\lrp{ p \log(p/\delta) \cdot \frac{L^2}{\epsilon^2} }$.

Combining the two bounds, we have that with $1-\delta$ probability,
\[
\lrn{ \Omega^{*} - \frac{1}{n} \Omega^{(T)}}_F^2
\leq 6 \sum_{k=p+1}^d \lrp{ \frac{\lambda^{(\infty)}_k}{n} }^2
+ 24 \frac{\epsilon}{n} \sum_{k=1}^p \frac{\lambda^{(\infty)}_k}{n} + 16 \frac{\epsilon^2}{n^2}
+ 8 \frac{R^2}{n}\log\frac{2}{\delta}.
\]
In terms of the eigenvalues $\lambda^*_k$ of $\Omega^*$, we can apply our bound of $\lrn{\Omega^* - \frac{1}{n} \Omega^{(\infty)} }_F$ 
to obtain that when $n \geq 8 \frac{R^2}{(\alpha^*)^2} \log\frac{2}{\delta}$,
\begin{align*}
\lrn{\Omega^{*} - \frac{1}{n} \Omega^{(T)}}_F^2
&\leq 12 \sum_{k=p+1}^d \lrp{ \lambda^*_k }^2
+ 24 \epsilon^* \sum_{k=1}^p \lambda^*_k
+ 16 (\epsilon^*)^2
+ 56 \frac{R^2}{n}\log\frac{2}{\delta}
+ 48 \epsilon^* \sqrt{p} R \sqrt{\frac{1}{n}\log\frac{2}{\delta}} \\
&\leq 12 \sum_{k=p+1}^d \lrp{ \lambda^*_k }^2
+ 24 \epsilon^* \sum_{k=1}^p \lambda^*_k
+ 40 p (\epsilon^*)^2
+ 80 \frac{R^2}{n}\log\frac{2}{\delta} \\
&\lesssim \sum_{k=p+1}^d \lrp{ \lambda^*_k }^2
+ \epsilon^* \sum_{k=1}^p \lambda^*_k
+ \frac{R^2}{n}\log\frac{1}{\delta},
\end{align*}
with $1-\delta$ probability.
Here the accuracy for estimating $\Omega^*$, $\epsilon^* = \epsilon/n$. 
In terms of computation resource, the number of stochastic gradient samples per iteration $N = \widetilde{\boldsymbol{\Omega}}\lrp{ p d \cdot \lrp{\frac{L^*}{\alpha^*}}^2 \cdot \frac{(L^*)^2}{(\epsilon^*)^2 \sqrt{\delta} } }$, the number of iterations $T = \boldsymbol{\Theta}\lrp{ \frac{L^*}{\epsilon^*} }$, as well as the number of samples for eigenvalue computation $n = {\boldsymbol{\Omega}}\lrp{ p \log(p/\delta) \cdot \frac{(L^*)^2}{(\epsilon^*)^2} }$.

Choosing the same allocation rule as in \Cref{thm:KL_convergence}, we obtain that with a computation budget that allows for $\Pi$ gradient evaluations,
\[
\lrn{\Omega^{*} - \frac{1}{n} \Omega^{(T)}}_F^2
\lesssim \sum_{k=p+1}^d \lrp{\lambda^{*}_k}^2
+ \lrp{\frac{pd}{\Pi}}^{1/3} \cdot \lrp{\frac{L^*}{\alpha^*}}^{2/3} \cdot \frac{L^*}{\delta^{1/6}} \cdot \sum_{k=1}^p \lambda^{*}_k 
+ \frac{R^2}{n}\log\frac{1}{\delta},
\]
with $1-\delta$ probability.
\end{proof}

\section{Proofs for solving for the non-Gaussian cases}
\label{sec:non-normal_proof}
Define the Bures-Wasserstein distance on the space of positive definite matrices:
\[
d^2_W\lrp{\Omega_1,\Omega_2}
= \tr\lrp{ \Omega_1 + \Omega_2 - 2 \lrp{\Omega_1^{1/2}\Omega_2\Omega_1^{1/2}}^{1/2} }.
\]

In what follows we formally state the three assumptions mentioned in the main text.
\paragraph{Assumptions in non-Gaussian cases}
\begin{enumerate}[label=\rm{A}{{\arabic*}}]
    \item Function $\psi$ is twice differentiable, and $\exists$ symmetric positive semi-definite $\Psi$ such that $\nabla^2 \psi(\theta) \preceq \alpha\mI + \Psi, \forall \theta\in\real^d$. Denote the Lipschitz smoothness of $\psi$ to be $L = \alpha + \sigma_{\max}(\Psi)$. \label{assumption:smooth}
    \item For any symmetric positive definite matrix $\Omega$ such that $\alpha\cdot\mI\preceq\Omega\preceq L\cdot\mI$, $\Ep{\theta\sim \mathcal{N}\lrp{0,\Omega^{-1}}}{ \nabla^2 \psi(\theta) } \succeq \alpha \mI$. \label{assumption:convex}
    \item There exists $0<\rho<1$ such that for any $\alpha \cdot \mI \preceq \Omega_1, \Omega_2 \preceq L \cdot \mI$,
    \begin{align*}
        d_W\lrp{ \Ep{ \theta\sim \mathcal{N}\lrp{0,\ \Omega_1^{-1}} }{ \nabla^2 \psi(\theta) } , \Ep{ \theta'\sim \mathcal{N}\lrp{0,\ \Omega_2^{-1}} }{ \nabla^2 \psi(\theta') } }
        \leq \rho \cdot d_W\lrp{\Omega_1,\Omega_2}.
    \end{align*}
    \label{assumption:Hessian_Lip}
\end{enumerate}

Note that the above Assumptions \ref{assumption:smooth}--\ref{assumption:Hessian_Lip} are implied by the simpler (but stronger) conditions that: $\alpha\mI \preceq \nabla^2 \psi(\theta) \preceq \alpha\mI + \Psi, \forall \theta\in\real^d$, and that
$$\lrn{ \lrp{\nabla^2 \psi(\theta)}^{1/2} - \lrp{\nabla^2 \psi(\theta')}^{1/2}}_F \leq  \rho \sqrt{\frac{\alpha}{L} \log\frac{\alpha}{L}} \cdot \lrn{\theta-\theta'}, \forall \theta, \theta' \in \real^d,$$ as proved in the following \Cref{lem:equiv_non-normal}.
\begin{lemma}
\label{lem:equiv_non-normal}
Assume that $\nabla^2 \psi(\theta) \succeq \alpha\mI, \forall \theta\in\real^d$, and that
$$\lrn{ \lrp{\nabla^2 \psi(\theta)}^{1/2} - \lrp{\nabla^2 \psi(\theta')}^{1/2}}_F \leq \rho \sqrt{\frac{\alpha}{L} \log\frac{\alpha}{L}} \cdot \lrn{\theta-\theta'}, \forall \theta, \theta' \in \real^d.$$
Then $\Ep{\theta\sim \mathcal{N}\lrp{0,\Omega^{-1}}}{ \nabla^2 \psi(\theta) } \succeq \alpha \mI$, and
\begin{align*}
d_W\lrp{ \Ep{ \theta\sim \mathcal{N}\lrp{0,\ \Omega_1^{-1}} }{ \nabla^2 \psi(\theta) } , \Ep{ \theta'\sim \mathcal{N}\lrp{0,\ \Omega_2^{-1}} }{ \nabla^2 \psi(\theta') } }
\leq \rho \cdot d_W\lrp{\Omega_1,\Omega_2},
\end{align*}
for any $\Omega,\Omega_1,\Omega_2\succeq\alpha\mI$ and $\Omega,\Omega_1,\Omega_2\preceq L\mI$.
\end{lemma}


\begin{theorem}
\label{thm:non-normal_appnd}[Restatement of \Cref{thm:non-normal}]
Assume that Assumptions \ref{assumption:smooth}--\ref{assumption:Hessian_Lip} hold.
Then take in the inner loop of the SVI\_General Algorithm number of stochastic gradient samples per iteration $N = \widetilde{\boldsymbol{\Omega}}\lrp{ \max\lrbb{ \frac{pd}{ m_0^2 } , \frac{p^2}{ m_0^4 } } \cdot \frac{L^3}{\alpha^3} \cdot \frac{L^2}{\epsilon^2 \sqrt{\delta} } }$, number of iterations $T = \widetilde{\boldsymbol{\Theta}}\lrp{ \frac{L}{\epsilon}}$, as well as number of samples for eigenvalue computation $M = \widetilde{{\boldsymbol{\Omega}}}\lrp{ p \cdot \frac{L^2}{\epsilon^2} }$.
After $K = \frac{\log\lrp{ (L^2+1) \cdot p + \sum_{i=p+1}^d \sigma_i(\Psi)^2}}{\log(\rho)} + \frac{\log(1/\epsilon)}{\log(\rho)} = \widetilde{\boldsymbol{\Omega}}(1)$ global iterations, we obtain $\lrp{U^{(T)}_{K},\Lambda^{(T)}_{K}}$ so that for $\epsilon \leq \frac{1}{2} \sum_{i=1}^p \Lambda^{(\infty)}_{k+1}(i)$, 
\begin{multline*}
\KL{q_{\lrp{U^{(T)}_{K},\Lambda^{(T)}_{K}}}(\theta \giv \mathbf{x})}{q_{\Omega^{(\infty)}_\infty}(\theta \giv \vx)}\\ 
\lesssim \lrp{ \frac{1}{(1-\rho)^2} \frac{L}{\alpha} \log\frac{L}{\alpha} \cdot \frac{1}{\alpha^2} } \cdot \lrp{\sum_{i=p+1}^d \sigma_i(\Psi)^2 + \epsilon \sum_{i=1}^p \sigma_i(\Psi) + \epsilon^2},
\end{multline*}
with $1-\delta$ probability.
Here $\Omega^{(\infty)}_\infty$ minimizes $\KL{\mathcal{N}\lrp{0,\Omega^{-1}}}{p(\theta|\vx)}$ over $\Omega$.
\end{theorem}



Before proving \Cref{thm:non-normal}, we first provide a lemma about the stationary solution of \Cref{alg:outer_loop}.
\begin{lemma}
\label{prop:stationarity}
Consider \Cref{alg:outer_loop} (SVI\_General) in the asymptotic limit where $p=d$, and $N, T, K\rightarrow\infty$.
When matrix $\Omega_{k}$ is the input to the inner loop \Cref{alg:Gaussian} (SVI\_Gauss), the output of it satisfies
\[
\Omega_{k+1}^{(\infty)} = D + U_{k+1}^{(\infty)} \Lambda_{k+1}^{(\infty)} \lrp{U_{k+1}^{(\infty)}}^\rT 
= \Ep{ \theta\sim\mathcal{N}\big( 0, (\Omega_{k})^{-1} \big) }{ \nabla^2 \psi(\theta) }.
\]
In addition, the stationary solution $\Omega_{\infty}^{(\infty)} = D + U_{\infty}^{(\infty)} \Lambda_{\infty}^{(\infty)} \lrp{ U_{\infty}^{(\infty)} }^\rT$ to \Cref{alg:outer_loop} (SVI\_General) satisfies that 
\[
\Omega_{\infty}^{(\infty)} = \Ep{ \theta\sim\mathcal{N}\big( 0, ( \Omega_{\infty}^{(\infty)} )^{-1} \big) }{ \nabla^2 \psi(\theta) },
\]
and minimizes $\KL{\mathcal{N}\lrp{0,\Omega^{-1}}}{p(\theta|\vx)}$ for $p(\theta|\vx) \propto \exp\lrp{-\psi(\theta) }$.
\end{lemma}

\begin{proof}[Proof of \Cref{thm:non-normal}]
We will prove the theorem via combining two components.
One component is when we take $\Omega_{k+1} = \Omega^{(T)}_k$, the error incurred in \Cref{alg:Gaussian} (SVI\_Gauss): $d_W\lrp{\Omega^{(T)}_k, \Omega^{(\infty)}_k}$.
We will apply \Cref{thm:KL_convergence} to prove this part.
Another component is the contraction as we approximately take the step of: $\Omega_{k+1} = \Ep{ \theta\sim\mathcal{N}\lrp{ 0, \lrp{\Omega_{k}}^{-1} } }{ \nabla^2 \psi(\theta) }$.

We start with the second component and focus on: starting from two different inputs $ \Omega_{k}$ and $\widetilde{\Omega}_{k} $ at the $k$-th iteration,
how $d_W\lrp{ \Omega^{(\infty)}_{k+1}, \widetilde{\Omega}^{(\infty)}_{k+1} }$ compares to $d_W\lrp{ \Omega_{k}, \widetilde{\Omega}_{k} }$.
From \Cref{prop:stationarity} we know that $\Omega_{k+1}^{(\infty)} = \Ep{ \theta\sim\mathcal{N}\big( 0, (\Omega_{k})^{-1} \big) }{ \nabla^2 \psi(\theta) }$.
Hence we obtain that 
\[
d_W\lrp{ \Omega^{(\infty)}_{k+1}, \widetilde{\Omega}^{(\infty)}_{k+1} }
= d_W\lrp{ \Ep{ \theta\sim\mathcal{N}\big( 0, (\Omega_{k})^{-1} \big) }{ \nabla^2 \psi(\theta) }, \Ep{ \tilde{\theta}\sim\mathcal{N}\big( 0, (\widetilde{\Omega}_{k})^{-1} \big) }{ \nabla^2 \psi(\tilde{\theta}) } }
\]
Applying our assumption that for $\alpha\mI\preceq\Omega_1,\Omega_2\preceq\Psi$,
\begin{align*}
d_W\lrp{ \Ep{ \theta\sim \mathcal{N}\lrp{0,\ \Omega_1^{-1}} }{ \nabla^2 \psi(\theta) } , \Ep{ \theta'\sim \mathcal{N}\lrp{0,\ \Omega_2^{-1}} }{ \nabla^2 \psi(\theta') } }
\leq \rho \cdot d_W\lrp{\Omega_1,\Omega_2},
\end{align*}
we obtain that 
\[
d_W\lrp{ \Omega^{(\infty)}_{k+1}, \widetilde{\Omega}^{(\infty)}_{k+1} }
\leq \rho \cdot d_W\lrp{\Omega_{k},\widetilde{\Omega}_{k}}.
\]
We take $\widetilde{\Omega}_{k} = \Omega^{(\infty)}_\infty$ and set the input to \Cref{eq:global_iter}: $\Omega_{k} = \Omega_{k}^{(T)}$, which is computed from the output of \Cref{eq:global_iter} during the previous round of \Cref{alg:outer_loop} (SVI\_General). 
We obtain from \Cref{prop:stationarity} about the stationary solution $\Omega^{(\infty)}_\infty$ that
\begin{align}
d_W\lrp{ \Omega^{(\infty)}_{k+1}, \Omega^{(\infty)}_\infty }
\leq \rho \cdot d_W\lrp{\Omega^{(T)}_{k}, \Omega^{(\infty)}_\infty}.
\label{eq:W_2_contraction}
\end{align}

We then incorporate the error incurred in \Cref{alg:Gaussian} (SVI\_Gauss):
\begin{align}
d_W\lrp{ \Omega^{(T)}_{k+1}, \Omega^{(\infty)}_\infty }
\leq d_W\lrp{ \Omega^{(T)}_{k+1}, \Omega^{(\infty)}_{k+1} } + d_W\lrp{ \Omega^{(\infty)}_{k+1}, \Omega^{(\infty)}_\infty }.
\label{eq:W2_error_decomp}
\end{align}
For $d_W\lrp{ \Omega^{(T)}_{k+1}, \Omega^{(\infty)}_{k+1} }$, we invoke \Cref{lem:W_2_bound} to have:
\[
d_W\lrp{ \Omega^{(T)}_{k+1}, \Omega^{(\infty)}_{k+1} }
\leq \sqrt{L} \cdot \sqrt{ \KL{ q_{\Omega^{(T)}_{k+1}} (\theta|\vx) }{ q_{\Omega^{(\infty)}_{k+1}} (\theta|\vx) } }.
\]
Applying \Cref{thm:KL_convergence} and \Cref{lem:Oja_convergence}, we know that 
taking number of stochastic gradient samples per iteration $N = \widetilde{\boldsymbol{\Omega}}\lrp{ \max\lrbb{ \frac{pd}{ m_0^2 } , \frac{p^2}{ m_0^4 } } \cdot \frac{L^4}{\alpha^2} \cdot \frac{ \kappa\lrp{ \Omega^{(T)}_k } }{\epsilon^2 \sqrt{\delta'} } }
= \widetilde{\boldsymbol{\Omega}}\lrp{ \max\lrbb{ \frac{pd}{ m_0^2 } , \frac{p^2}{ m_0^4 } } \cdot \frac{L^3}{\alpha^3} \cdot \frac{ L^2 }{\epsilon^2 \sqrt{\delta'} } }$, 
number of iterations $T = \boldsymbol{\Theta}\lrp{ \frac{L}{\epsilon} \log\frac{1}{m_0}}$,
and $M \geq 2 p \log(8p/\delta') \cdot \frac{L^2}{\epsilon^2}$, 
\begin{align*}
\KL{q_{\lrp{U^{(T)}_{k+1},\Lambda^{(T)}_{k+1}}}(\theta \giv \mathbf{x})}{q_{\Omega^{(\infty)}_{k+1}}(\theta \giv \vx)} 
\leq \frac{3}{2\alpha^2} \sum_{i=p+1}^d \lrp{ \Lambda^{(\infty)}_{k+1}(i) }^2 + \frac{6}{\alpha^2} \epsilon \sum_{i=1}^p \Lambda^{(\infty)}_{k+1}(i) + \frac{4\epsilon^2}{\alpha^2},
\end{align*}
with probability $1-\delta'$.
We thus have the bound for $d_W\lrp{ \Omega^{(T)}_{k+1}, \Omega^{(\infty)}_{k+1} }$:
\begin{align}
d_W\lrp{ \Omega^{(T)}_{k+1}, \Omega^{(\infty)}_{k+1} }
\leq \sqrt{\frac{3}{2}} \frac{\sqrt{L}}{\alpha} \sqrt{ \sum_{i=p+1}^d \lrp{ \Lambda^{(\infty)}_{k+1}(i) }^2 } + \sqrt{6} \frac{\sqrt{L}}{\alpha} \sqrt{\epsilon} \sqrt{ \sum_{i=1}^p \Lambda^{(\infty)}_{k+1}(i) } + 2 \frac{\sqrt{L}}{\alpha} \epsilon.
\label{eq:W2_iter_error}
\end{align}
Invoking Assumption~\ref{assumption:smooth}, $\nabla^2 \psi(\theta) \preceq \alpha\mI + \Psi$.
Then using linearity of expectation, we obtain that for $\Omega_{k+1}^{(\infty)} = \Ep{ \theta\sim\mathcal{N}\big( 0, (\Omega_{k})^{-1} \big) }{ \nabla^2 \psi(\theta) }$, $\Omega_{k+1}^{(\infty)} \preceq \alpha\mI + \Psi$.
Applying Weyl's monotonicity theorem to $\Omega_{k+1}^{(\infty)}$, we have $ \Lambda^{(\infty)}_{k+1}(i) \leq \sigma_i(\Psi)$.
Plugging this result and \Cref{eq:W2_iter_error} into \Cref{eq:W2_error_decomp}, we arrive at
\begin{multline*}
d_W\lrp{ \Omega^{(T)}_{k+1}, \Omega^{(\infty)}_\infty }\\
\leq d_W\lrp{ \Omega^{(\infty)}_{k+1}, \Omega^{(\infty)}_\infty } + \sqrt{\frac{3}{2}} \frac{\sqrt{L}}{\alpha} \sqrt{ \sum_{i=p+1}^d \sigma_i(\Psi)^2 } 
+ \sqrt{6} \frac{\sqrt{L}}{\alpha} \sqrt{\epsilon} \sqrt{ \sum_{i=1}^p \sigma_i(\Psi) }
+ 2 \frac{\sqrt{L}}{\alpha} \epsilon.
\end{multline*}

Further applying \Cref{eq:W_2_contraction},
we obtain a recurrent relationship between $\Omega_{K}^{(T)}$ and $\Omega_{K-1}^{(T)}$ and expand it:
\begin{align*}
&d_W\lrp{ \Omega^{(T)}_{K}, \Omega^{(\infty)}_\infty }\\
&\leq \rho \cdot d_W\lrp{\Omega^{(T)}_{K-1}, \Omega^{(\infty)}_\infty} + \sqrt{\frac{3}{2}} \frac{\sqrt{L}}{\alpha} \sqrt{ \sum_{i=p+1}^d \sigma_i(\Psi)^2 } + \sqrt{6}\frac{\sqrt{L}}{\alpha} \sqrt{\epsilon} \sqrt{ \sum_{i=1}^p \sigma_i(\Psi) } 
+ 2 \frac{\sqrt{L}}{\alpha} \epsilon \\
&\leq \rho^{K} \cdot d_W\lrp{\Omega_{0}, \Omega^{(\infty)}_\infty} 
+ \frac{\sqrt{3/2}}{1-\rho} \cdot \frac{\sqrt{L}}{\alpha} \sqrt{ \sum_{i=p+1}^d \sigma_i(\Psi)^2 } \\
&+ \frac{\sqrt{6}}{1-\rho} \cdot \frac{\sqrt{L}}{\alpha} \sqrt{\epsilon} \sqrt{ \sum_{i=1}^p \sigma_i(\Psi) } 
+ \frac{2}{1-\rho} \frac{\sqrt{L}}{\alpha} \epsilon,
\end{align*}
since the first input to \Cref{eq:global_iter} is $\Omega_{0}$. 

In terms of the KL-divergence,
\begin{align*}
\lefteqn{ \KL{q_{\lrp{U^{(T)}_{K},\Lambda^{(T)}_{K}}}(\theta \giv \mathbf{x})}{q_{\Omega^{(\infty)}_\infty}(\theta \giv \mathbf{x})} } \\
&\leq \lsi d_W^2\lrp{ \Omega^{(T)}_{K}, \Omega^{(\infty)}_\infty } \\
&\leq 4 \rho^{2K} \lsi \cdot d_W^2\lrp{\Omega_{0}, \Omega^{(\infty)}_\infty}
+ \frac{6}{(1-\rho)^2} \lsi \cdot \frac{L}{\alpha^2}  \sum_{i=p+1}^d \sigma_i(\Psi)^2 \\
&+ \frac{24}{(1-\rho)^2} \lsi \cdot \frac{L}{\alpha^2} \epsilon \sum_{i=1}^p \sigma_i(\Psi)
+ \frac{16}{(1-\rho)^2} \lsi \frac{L}{\alpha^2} \cdot \epsilon^2.
\end{align*}
Since $\Omega_0 = \alpha\mI_{d \times d} + \left[e_1,\dots,e_p\right] \cdot \mI_{p \times p} \cdot \left[e_1,\dots,e_p\right]^\rT$ and $\alpha\mI_{d \times d} \preceq \Omega^{(\infty)}_\infty \preceq \alpha\mI_{d \times d} + \Psi$,
\begin{align*}
d_W^2\lrp{ \Omega_{0}, \Omega^{(\infty)}_\infty }
&\leq L \cdot \KL{ \mathcal{N}\lrp{0, \lrp{\Omega^{(\infty)}_\infty}^{-1} }}{\mathcal{N}\lrp{ 0,\Omega_{0}^{-1} } } \\
&\leq \frac{L}{2\alpha^2} \lrn{\Omega_0 - \Omega^{(\infty)}_\infty}_F^2
\leq \frac{L}{\alpha^2} \lrp{ \lrn{\mI_{p \times p}}_F^2 + \lrn{\Psi}_F^2 }
\leq \frac{L}{\alpha^2}\lrp{ p + \sum_{i=1}^d \sigma_i(\Psi)^2 } \\
&\leq \frac{L}{\alpha^2}\lrp{ (L^2+1) \cdot p + \sum_{i=p+1}^d \sigma_i(\Psi)^2 }.
\end{align*}
Therefore, when $K = \frac{\log\lrp{ (L^2+1) \cdot p + \sum_{i=p+1}^d \sigma_i(\Psi)^2}}{\log(\rho)} + \frac{ \log\lrp{1/\epsilon^2} }{\log(\rho)} = \widetilde{\boldsymbol{\Omega}}(1)$,
\begin{align*}
&\KL{q_{\lrp{U^{(T)}_{K},\Lambda^{(T)}_{K}}}(\theta \giv \mathbf{x})}{q_{\Omega^{(\infty)}_\infty}(\theta \giv \mathbf{x})}  \\
&\leq \frac{6}{(1-\rho)^2} \lsi \cdot \frac{L}{\alpha^2}  \sum_{i=p+1}^d \sigma_i(\Psi)^2 
+ \frac{24}{(1-\rho)^2} \lsi \cdot \frac{L}{\alpha^2} \epsilon \sum_{i=1}^p \sigma_i(\Psi)
+ \frac{20}{(1-\rho)^2} \lsi \frac{L}{\alpha^2} \cdot \epsilon^2 \\
&\lesssim \lrp{ \frac{1}{(1-\rho)^2} \frac{L}{\alpha} \log\frac{L}{\alpha} \cdot \frac{1}{\alpha^2} } \cdot \lrp{\sum_{i=p+1}^d \sigma_i(\Psi)^2 + \epsilon \sum_{i=1}^p \sigma_i(\Psi) + \epsilon^2},
\end{align*}
with $(1-\delta' \cdot K)$ probability.
Choosing $\delta = \delta'/K$ gives the result.
\end{proof}

\subsection{Proofs for the auxiliary lemmas}
\begin{proof}[Proof of \cref{lem:equiv_non-normal}]
\sloppy
The fact that $\Ep{\theta\sim \mathcal{N}\lrp{0,\Omega^{-1}}}{ \nabla^2 \psi(\theta) } \succeq \alpha \mI$ is implied by the concavity of the least eigenvalue of a real symmetric matrix: 
$$\lambda_{\min} \lrp{ \Ep{\theta\sim \mathcal{N}\lrp{0,\Omega^{-1}}}{ \nabla^2 \psi(\theta) } } 
\geq \Ep{\theta\sim \mathcal{N}\lrp{0,\Omega^{-1}}}{ \lambda_{\min} \lrp{ \nabla^2 \psi(\theta) } }
\geq \alpha.$$

To upper bound $d_W\lrp{ \Ep{ \theta\sim \mathcal{N}\lrp{0,\ \Omega_1^{-1}} }{ \nabla^2 \psi(\theta) } , \Ep{ \theta'\sim \mathcal{N}\lrp{0,\ \Omega_2^{-1}} }{ \nabla^2 \psi(\theta') } }$,
we first note that $d_W^2(\Omega_1,\Omega_2) = \tr\lrp{ \Omega_1 + \Omega_2 - 2 \lrp{\Omega_1^{1/2}\Omega_2\Omega_1^{1/2}}^{1/2} }$ is a convex function in $\Omega_2$ (and in $\Omega_1$ by symmetry).
This is because function $f(\cdot) = (\cdot)^{1/2}$ is operator concave by the L\"{o}wner-Heinz theorem, leading to the fact that $\tr\lrp{\lrp{\Omega_1^{1/2} \Omega_2 \Omega_1^{1/2}}^{1/2}}$ is a concave function in $\Omega_2$.
Hence $d^2_W\lrp{\Omega_1,\Omega_2}$ is a convex function in $\Omega_2$.

Using this convex property, we have the following result.
Denote $\tilde{\gamma}$ as the optimal coupling between $\theta \sim \mathcal{N} \lrp{0,\Omega_1^{-1}}$ and $\theta' \sim \mathcal{N} \lrp{0,\Omega_2^{-1}}$, then under the assumption that $\lrn{ \lrp{\nabla^2 \psi(\theta)}^{1/2} - \lrp{\nabla^2 \psi(\theta')}^{1/2}}_F \leq \rho' \lrn{\theta-\theta'}$,
\begin{align*}
&d_W^2\lrp{ \Ep{ \theta\sim \mathcal{N}\lrp{0,\ \Omega_1^{-1}} }{ \nabla^2 \psi(\theta) } , \Ep{ \theta'\sim \mathcal{N}\lrp{0,\ \Omega_2^{-1}} }{ \nabla^2 \psi(\theta') } } \\
&\leq \int d_W^2\lrp{ \nabla^2 \psi(\theta) , \nabla^2 \psi(\theta') } \rd \tilde{\gamma}(\theta,\theta') \\
&\overset{\1}{\leq} \int \lrn{ \lrp{\nabla^2 \psi(\theta)}^{1/2} - \lrp{\nabla^2 \psi(\theta')}^{1/2}}_F^2 \rd \tilde{\gamma}(\theta,\theta') \\
&\leq \rho'^2 \int \lrn{\theta-\theta'}^2 \rd \tilde{\gamma}(\theta,\theta') \\
&= \rho'^2 d_W^2\lrp{\mathcal{N}(0,\Omega_2^{-1}), \mathcal{N}(0,\Omega_1^{-1})} \\
&= \rho'^2 d_W^2\lrp{\Omega_2^{-1}, \Omega_1^{-1}},
\end{align*}
where $\1$ follows from \Cref{lem:W_2_Fro}.

Applying \Cref{lem:W_2_bound}, we obtain that for $\Omega_1,\Omega_2\succeq\alpha\mI$ and $\Omega_1,\Omega_2\preceq L\mI$,
\begin{align*}
d_W^2\lrp{\Omega_2^{-1}, \Omega_1^{-1}}
&\leq L \cdot \KL{\mathcal{N}(0,\Omega_1)}{\mathcal{N}(0,\Omega_2)} \\
&= L \cdot \KL{\mathcal{N}(0,\Omega_2^{-1})}{\mathcal{N}(0,\Omega_1^{-1})} \\
&\leq \frac{L}{\alpha} \log\frac{L}{\alpha} \cdot d_W^2\lrp{\Omega_1, \Omega_2}.
\end{align*}
Taking $\rho' = \rho \sqrt{\frac{\alpha}{L} \log\frac{\alpha}{L}}$ yields the result that 
\[
d_W\lrp{ \Ep{ \theta\sim \mathcal{N}\lrp{0,\ \Omega_1^{-1}} }{ \nabla^2 \psi(\theta) } , \Ep{ \theta'\sim \mathcal{N}\lrp{0,\ \Omega_2^{-1}} }{ \nabla^2 \psi(\theta') } }
\leq \rho \cdot d_W(\Omega_1,\Omega_2).
\]
\end{proof}

\begin{proof}[Proof of \Cref{prop:stationarity}]
We first focus on the updates of \Cref{eq:sample_step_Gauss} and \Cref{eq:mult_step_Gauss} for the general posterior and its associated potential function $\psi$.
The stochastic update matrix $\widetilde{F}_\psi = \frac{1}{N} \sum_{i=1}^N \nabla \psi(\theta_i) \theta_i^\rT \Omega$ satisfy for any positive definite matrix $\Omega$ that
\[
F_\psi
=\mathbb{E}[\widetilde{F}_\psi] 
= \Ep{ \theta\sim\mathcal{N}\lrp{ 0, \lrp{\Omega}^{-1} } }{ \nabla \psi(\theta) \theta^\rT \Omega }
= \Ep{ \theta\sim\mathcal{N}\lrp{ 0, \lrp{\Omega}^{-1} } }{ \nabla^2 \psi(\theta) }.
\]
In the infinite sample limit where $N\rightarrow\infty$, the stochastic update matrix becomes the deterministic update matrix: $\widetilde{F}_\psi \rightarrow F_\psi$.
When we fix $\Omega_{k}$ in the input of $\textrm{SVI\_Gauss}$ \Cref{alg:Gaussian} (which is the inner loop of $\textrm{SVI\_General}$ \Cref{alg:outer_loop}), the update matrix $F_\psi = \Ep{ \theta\sim\mathcal{N}\lrp{ 0, \lrp{\Omega_{k}}^{-1} } }{ \nabla^2 \psi(\theta) }$ is a fixed matrix.
Taking $p=d$ and $T\rightarrow\infty$, the output of $\textrm{SVI\_Gauss}$ \Cref{alg:Gaussian} is
\[
\Omega_{k+1}^{(\infty)} = D + U_{k+1}^{(\infty)} \Lambda_{k+1}^{(\infty)} \lrp{U_{k+1}^{(\infty)}}^\rT 
= \Ep{ \theta\sim\mathcal{N}\big( 0, (\Omega_{k})^{-1} \big) }{ \nabla^2 \psi(\theta) },
\]
which is fed to the outer loop of $\textrm{SVI\_General}$ \Cref{alg:outer_loop}. 
The stationary solution $\Omega_{\infty}^{(\infty)}$ of the above update as $k\rightarrow\infty$ is
\[
\Omega_{\infty}^{(\infty)} = \Ep{ \theta\sim\mathcal{N}\big( 0, (\Omega_{\infty}^{(\infty)})^{-1} \big) }{ \nabla^2 \psi(\theta) }.
\]

We also prove that this solution optimizes the KL divergence $\KL{\mathcal{N}\lrp{0,\Omega^{-1}}}{p(\theta|\vx)}$ for $p(\theta|\vx) \propto \exp\lrp{-\psi(\theta) }$.
Taking gradient over the KL divergence:
\[
\nabla_{\Omega} \KL{\mathcal{N}\lrp{0,\Omega^{-1}}}{p(\theta|\vx)}
= \frac{1}{2} \Omega^{-1} - \frac{1}{2} \Omega^{-1} \Ep{\theta\sim\mathcal{N}\lrp{0,\Omega^{-1}}}{\nabla^2_\theta \psi(\theta)} \Omega^{-1}.
\]
Plugging the stationary solution $\Omega_{\infty}^{(\infty)} = \Ep{ \theta\sim\mathcal{N}\big( 0, (\Omega_{\infty}^{(\infty)})^{-1} \big) }{ \nabla^2 \psi(\theta) }$ into the above gradient evaluation, we obtain that it has zero gradient: 
\[\nabla_{\Omega} \KL{\mathcal{N}\lrp{0,\lrp{\Omega_{\infty}^{(\infty)}}^{-1}}}{p(\theta|\vx)} = 0.\]
From Assumption~\ref{assumption:convex}, we know that this stationary solution is in the interior of the space of positive definite matrices, and accordingly of the space of all normal distributions.
We also know from the convexity of the KL divergence that this stationary solution is the minimum of the KL divergence.
Therefore, the stationary solution satisfying $\Omega_{\infty}^{(\infty)} = \Ep{ \theta\sim\mathcal{N}\big( 0, (\Omega_{\infty}^{(\infty)})^{-1} \big) }{ \nabla^2 \psi(\theta) }$ minimizes the KL divergence.
\end{proof}

We now prove that the Wasserstein distance $d_W\lrp{\Omega_1,\Omega_2}$ is upper and lower bounded by the KL-divergence.
\begin{lemma}
\label{lem:W_2_bound}
Assume that the symmetric matrices $\alpha \cdot \mI \preceq \Omega_1 \preceq L \cdot \mI$ and that $\alpha \cdot \mI \preceq \Omega_2 \preceq L \cdot \mI$.
Then 
\[
d_W^2\lrp{\Omega_1,\Omega_2}
\leq L \cdot \KL{ \mathcal{N}\lrp{0,\Omega_2^{-1}}}{\mathcal{N}\lrp{0,\Omega_1^{-1}} },
\]
and
\begin{align*}
\KL{\mathcal{N}\lrp{0,\Omega_2^{-1}}}{\mathcal{N}\lrp{0,\Omega_1^{-1}}}
\leq \frac{1}{\alpha} \log\frac{L}{\alpha} \cdot d_W^2\lrp{\Omega_1,\Omega_2}.
\end{align*}
\end{lemma}

\begin{proof}[Proof of \Cref{lem:W_2_bound}]
For the upper bound on $d_W\lrp{\Omega_1,\Omega_2}$, we note that $d_W\lrp{\Omega_1,\Omega_2}$ also defines the $2$-Wasserstein distance between two centered normal distributions with covariances $\Omega_1$ and $\Omega_2$.
We hence denote $\rho_1$, $\rho_2$ as the density functions associated with $\mu_1 = \mathcal{N}\lrp{0,\Omega_1}$, and $\mu_2 = \mathcal{N}\lrp{0,\Omega_2}$ respectively and have $d_W\lrp{\mu_1,\mu_2}=d_W\lrp{\Omega_1,\Omega_2}$, where $d_W\lrp{\mu_1,\mu_2}$ is the $2$-Wasserstein distance between $\mu_1$ and $\mu_2$.
Note that $-\log \rho_2(\theta) = \frac{1}{2} \theta^\rT \Omega_2^{-1} \theta + C$, which is $1/L$-strongly convex and $1/\alpha$-Lipschitz smooth, which means that measure $\mu_2$ is $1/L$-strongly log-concave.

Using this fact, we leverage the Talagrand inequality for the $1/L$-strongly log-concave distribution $\rho_2$ and obtain that (c.f. \citep{otto2000generalization} and more concretely \citep{Yian_underdamped} for discussions about various generalizations of the strongly log-concave case):
\begin{align*}
d_W^2\lrp{\Omega_1,\Omega_2}
= d_W^2\lrp{\mu_1,\mu_2} 
\leq \frac{\KL{\mu_1}{\mu_2}}{1/L}.
\end{align*}
Plugging in the expression:
\[
\KL{\mu_1}{\mu_2} = \frac{1}{2} \lrp{ \log\frac{|\Omega_2|}{|\Omega_1|} - d + \tr\lrp{ \lrp{ \Omega_2 }^{-1} \Omega_1 } }
= \KL{\mathcal{N}\lrp{0,\Omega_2^{-1}}}{\mathcal{N}\lrp{0,\Omega_1^{-1}}},
\]
we obtain the result that
\[
d_W^2\lrp{\Omega_1,\Omega_2}
\leq L \cdot \KL{\mathcal{N}\lrp{0,\Omega_2^{-1}}}{\mathcal{N}\lrp{0,\Omega_1^{-1}}}.
\]

For the lower bound on $d_W\lrp{\Omega_1,\Omega_2}$, we similarly relate $\KL{\mu_1}{\mu_2}$ and $d_W^2\lrp{\mu_1,\mu_2}$.
Let $$\map = \Omega_2^{-1/2} \lrp{\Omega_2^{1/2} \Omega_1 \Omega_2^{1/2}}^{1/2} \Omega_2^{-1/2}$$ denote the optimal transport map from $\mu_2$ to $\mu_1$.
Then the $2$-Wasserstein distance can be equivalently represented as:
$d_W^2(\mu_1,\mu_2) = \Ep{\theta\sim\mu_2}{ \lrn{(\map-\mI) \theta}^2 } = \tr\lrp{\Omega_2\lrp{\map-\mI}^2}$.

By the change of variables formula, 
\[
\frac{\mu_2(\theta)}{\mu_1(\map \theta)}
= \det \map.
\]
Taking logarithm, the KL-divergence can be expressed as
\begin{align*}
\KL{\mu_1}{\mu_2} = \Ep{\theta'\sim\mu_1}{\log\frac{\mu_1(\theta')}{\mu_2(\theta')}}
&= \Ep{\theta\sim\mu_2}{\log\frac{\mu_1(\map \theta)}{\mu_2(\map \theta)}} \\
&= \Ep{\theta\sim\mu_2}{\log\mu_2(\theta) - \log\mu_2(\map \theta) - \log \det \map}.
\end{align*}
Let $\theta\sim\mu_2$. Then 
\begin{align*}
\KL{\mu_1}{\mu_2} 
&= \E{ \frac{1}{2} \theta^\rT \map \Omega_2^{-1} \map \theta - \frac{1}{2} \theta^\rT \Omega_2^{-1} \theta - \log\det\map } \\
&= \E{ \frac{1}{2} \lrn{\Omega_2^{-1/2}(\map-\mI)\theta}^2 + \theta^\rT \Omega_2^{-1} \map \theta - \theta^\rT \Omega_2^{-1} \theta - \log\det\map } \\
&\leq \E{ \frac{1}{2\sigma_{\min}(\Omega_2)} \lrn{(\map-\mI)\theta}^2 + \tr\lrp{ \theta \theta^\rT \Omega_2^{-1} \map - \theta \theta^\rT \Omega_2^{-1} } - \log\det\map } \\
&= \frac{1}{2\sigma_{\min}(\Omega_2)} d_W^2(\mu_1,\mu_2) + \tr \map - d - \log\det\map.
\end{align*}

We now bound $\tr \map - d - \log\det\map$.
Define function $f(z) = z - \log z - 1$.
Then we can write using $\lambda_i(\map)$ to denote the $i$-th eigenvalue of matrix $\map$,
\begin{align*}
\tr \map - d - \log\det\map
= \tr f(\map) 
= \sum_{i=1}^d f(\lambda_i(\map)).
\end{align*}
Applying the Caffarelli’s contraction theorem, we obtain that the eigenvalues of $\map$ are lower bounded: 
\[
\lambda_i(\map)
\geq \sqrt{\sigma_{\min}(\Omega_1) / \sigma_{\max}(\Omega_2)}, \quad \forall i=1,\dots,d.
\]
In addition, note that for all $z\geq z_0$, 
\[
f(z) \leq \max\lrbb{1,\log\frac{1}{z_0}} (z-1)^2.
\]
Plugging the lower bound of $\lambda_i(\map)$ inside, we obtain that 
\begin{align*}
\tr \map - d - \log\det\map
= \tr f(\map) 
&\leq \max\lrbb{1,\frac{1}{2}\log\frac{\sigma_{\max}(\Omega_2)}{\sigma_{\min}(\Omega_1)}} \sum_{i=1}^d \lrp{\lambda_i(\map)-1}^2 \\
&= \max\lrbb{1,\frac{1}{2}\log\frac{\sigma_{\max}(\Omega_2)}{\sigma_{\min}(\Omega_1)}} \lrn{\map-\mI}_F^2 \\
&\leq \frac{1}{\sigma_{\min}(\Omega_2)} \max\lrbb{1,\frac{1}{2}\log\frac{\sigma_{\max}(\Omega_2)}{\sigma_{\min}(\Omega_1)}} \tr\lrp{\Omega_2\lrp{\map-\mI}^2} \\
&= \frac{1}{\sigma_{\min}(\Omega_2)} \max\lrbb{1,\frac{1}{2}\log\frac{\sigma_{\max}(\Omega_2)}{\sigma_{\min}(\Omega_1)}} d_W^2(\mu_1,\mu_2).
\end{align*}

To sum up, we have proven that 
\begin{align*}
\KL{\mathcal{N}\lrp{0,\Omega_2^{-1}}}{\mathcal{N}\lrp{0,\Omega_1^{-1}}}
&= \KL{\mu_1}{\mu_2} \\
&\leq \frac{2}{\sigma_{\min}(\Omega_2)} \max\lrbb{1,\frac{1}{2}\log\frac{\sigma_{\max}(\Omega_2)}{\sigma_{\min}(\Omega_1)}} d_W^2(\mu_1,\mu_2) \\
&= \frac{2}{\sigma_{\min}(\Omega_2)} \max\lrbb{1,\frac{1}{2}\log\frac{\sigma_{\max}(\Omega_2)}{\sigma_{\min}(\Omega_1)}} d_W^2\lrp{\Omega_1,\Omega_2}.
\end{align*}
Since $\alpha \cdot \mI \preceq \Omega_1 \preceq L \cdot \mI$ and $\alpha \cdot \mI \preceq \Omega_2 \preceq L \cdot \mI$,
\begin{align*}
\KL{\mathcal{N}\lrp{0,\Omega_2^{-1}}}{\mathcal{N}\lrp{0,\Omega_1^{-1}}}
&\leq \frac{1}{\sigma_{\min}(\Omega_2)} \log\frac{\sigma_{\max}(\Omega_2)}{\sigma_{\min}(\Omega_1)} d_W^2\lrp{\Omega_1,\Omega_2} \\
&= \frac{1}{\alpha} \log\frac{L}{\alpha} d_W^2\lrp{\Omega_1,\Omega_2}.
\end{align*}

\end{proof}

In addition, we have the following Frobenius norm bound on $d_W\lrp{\Omega_1,\Omega_2}$.
\begin{lemma}
For positive semi-definite matrices $\Omega_1$, $\Omega_2$,
\[
d_W\lrp{\Omega_1,\Omega_2}
\leq \lrn{ \Omega_1^{1/2} - \Omega_2^{1/2} }_F.
\]
\label{lem:W_2_Fro}
\end{lemma}

\begin{proof}[Proof of \Cref{lem:W_2_Fro}]
To bound $d_W\lrp{\Omega_1,\Omega_2}$, we first rewrite $d^2_W\lrp{\Omega_1,\Omega_2}$ as
\[
d^2_W\lrp{\Omega_1,\Omega_2}
= \tr\lrp{ \Omega_1 + \Omega_2 - 2 \lrp{\Omega_1 \Omega_2}^{1/2} }.
\]
Note that by the Lieb and Thirring inequality (c.f., \citep{traceineq0}), 
\[
\tr\lrp{ \Omega_1^{1/2} \Omega_2^{1/2} }
\leq \tr\lrp{ \lrp{\Omega_1 \Omega_2}^{1/2} }.
\]
Also note that $\tr\lrp{ \lrp{\Omega_1 \Omega_2}^{1/2} } = \tr\lrp{ \lrp{\Omega_2 \Omega_1}^{1/2} }$ because of the properties of matrix square root and trace.
Plugging into the expression of $d^2_W\lrp{\Omega_1,\Omega_2}$ yields
\[
d^2_W\lrp{\Omega_1,\Omega_2} 
\leq \tr\lrp{\lrp{ \Omega_1^{1/2} - \Omega_2^{1/2} }^2}
= \lrn{ \Omega_1^{1/2} - \Omega_2^{1/2} }_F^2.
\]
\end{proof}

\section{Details of the Empirical Study about Gaussian posterior and Bayesian Logistic Regression}
\label{sec:empirical-logistic}
\subsection{Additional plots of the Gaussian posterior study}
\Cref{{fig:Gaussian_full_with_mf}} demonstrates the optimal KL achieved by Gaussian inferential models given sufficient computational budget. While low-rank inferential models trade statistical accuracy for computational efficiency, when sufficient computational resource is accessible, Gaussian inferential models with arppoximating ranks as high as the true rank can effectively recover the true Gaussian posterior.

\begin{figure}[h]
    \centering
    \includegraphics[width = 0.8\textwidth]{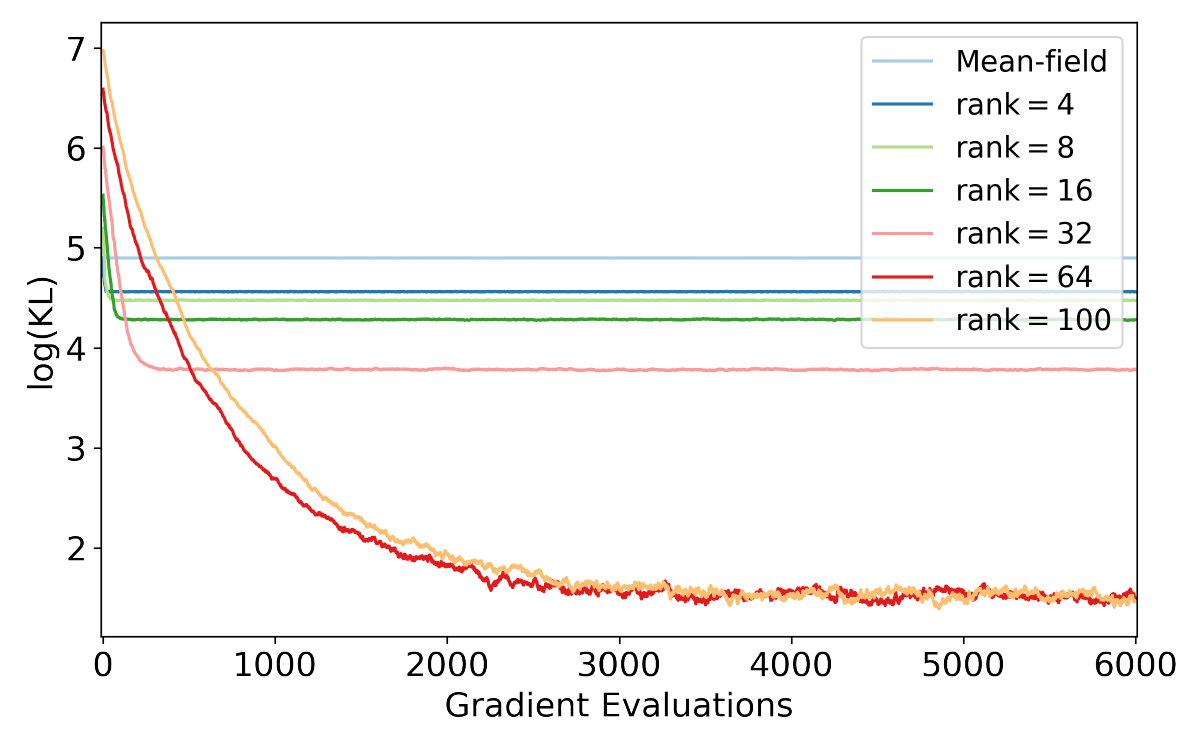}
    \caption{KL Convergence result for Gaussian inferential models with rank $p = 4, 8, 16, 32$ with $6000$ epochs. Inferential models with high ranks can achieve higher statistical accuracy in the case of sufficient computational budget on Gaussian Posterior.} 
    \label{fig:Gaussian_full_with_mf}
\end{figure}

\subsection{Dataset Information}
The original cardiac arrhythmia dataset \citep{Dua_2019} aims to distinguish the presence of different types of cardiac arrhythmia in the patients, and classify each medical record into 16 subtypes using linear-valued and nominal features. 
We preprocess the data to project the original feature space into a subspace with dimension $d = 110$, so as to effectively capture the covariance structure of the dataset. $Z$-score normalization is performed to standardize features for performance and convergence concerns. A binary target variable is then introduced based on the presence of different types of cardiac arrhythmia. 

Many such features are not Gaussian distributed and are heavily correlated, posing difficulties for inference and causing mean-field variational approximation family to have large bias.
As a result, more flexible inferential models are desirable to deal with the practical data.
On the other hand, a noticeable amount of these features are independent to each other and contribute little to the classification result. Unlike the mean-field Gaussian family, the low-rank Gaussian inferential models excel at capturing the inherent dependence of data yet providing computational efficiency. 



\subsection{Experimental setup}
We aim to provide a variational approximation $q_{(U,\Lambda)}(\theta|\mathbf{x}) = \N(\mu, \Omega ^{-1})$ to the posterior $p(\theta \giv \mathbf{x})$ of the cardiac arrhythmia dataset, where the precision matrix of the variational family is parameterized as $\Omega = \alpha \mI_d + U \Lambda U^\rT$. Rank $p$ is varied to study the statistical and computational trade-offs. We focus on learning the low-rank components of the precision matrix and set the diagonal structure to be isotropic with some positive constant $\alpha$ for simplicity. We then run the preconditioned SGD (\cref{alg:Gaussian}) to sequentially optimize $U$ and $\Lambda$ with respect to the KL divergence until convergence is achieved. To compare with the mean-field family, we parameterize the precision matrix as $\Omega = \diag(u_0^2)$ with diagonal structure only, and 
optimize $u_0$ using the same approach. In both inferential models, optimal fixed learning rates are applied.

Note that in contrast to the learning process of the mean vector $\mu$ in the inferential models, optimizing the precision matrix $\Omega$ tends to be much more computationally expensive in practice. To study the trade-offs, it is therefore sufficient to fix $\mu$ to be the estimated result returned by NUTS sampler, and optimize $\Omega$ only. The estimation result of $\mu$ is obtained from 20k samples generated from 2 parallel chains when using the NUTS algorithm. For initialization, we adopt a prior distribution $p(\theta) = \N(\mu, \beta \mI_d)$ with some positive number $\beta$ for both $U$ and $\Lambda$.

Results are averaged over 30 random seeds, with each optimized over $6000$ training epochs, reported along with the $95 \%$ confidence interval respectively.

\subsection{Additional plots of the Bayesian logistic regression study}
\label{appnd:logistic_results}
\begin{figure}[h]
    \centering
    \includegraphics[scale = 0.55]{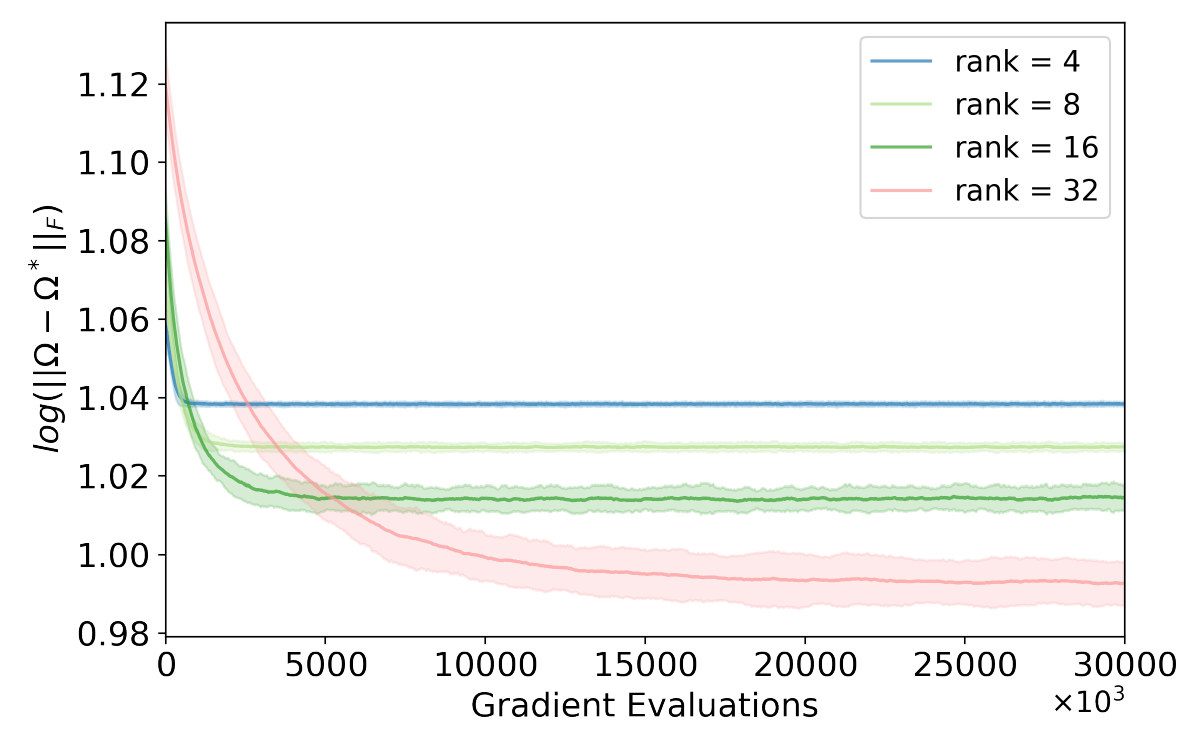}
    \caption{Convergence result for optimal low-rank inferential models with rank $p = 4, 8, 16, 32$ with $6000$ epochs. Inferential models with high ranks can achieve higher statistical accuracy in the case of sufficient computational budget on Cardiac arrhythmia dataset.} 
    \label{fig:lr_full_without_mf}
\end{figure}

\begin{figure}[h!]
    \centering
    \hspace*{-1cm}
    \includegraphics[scale = 0.45]{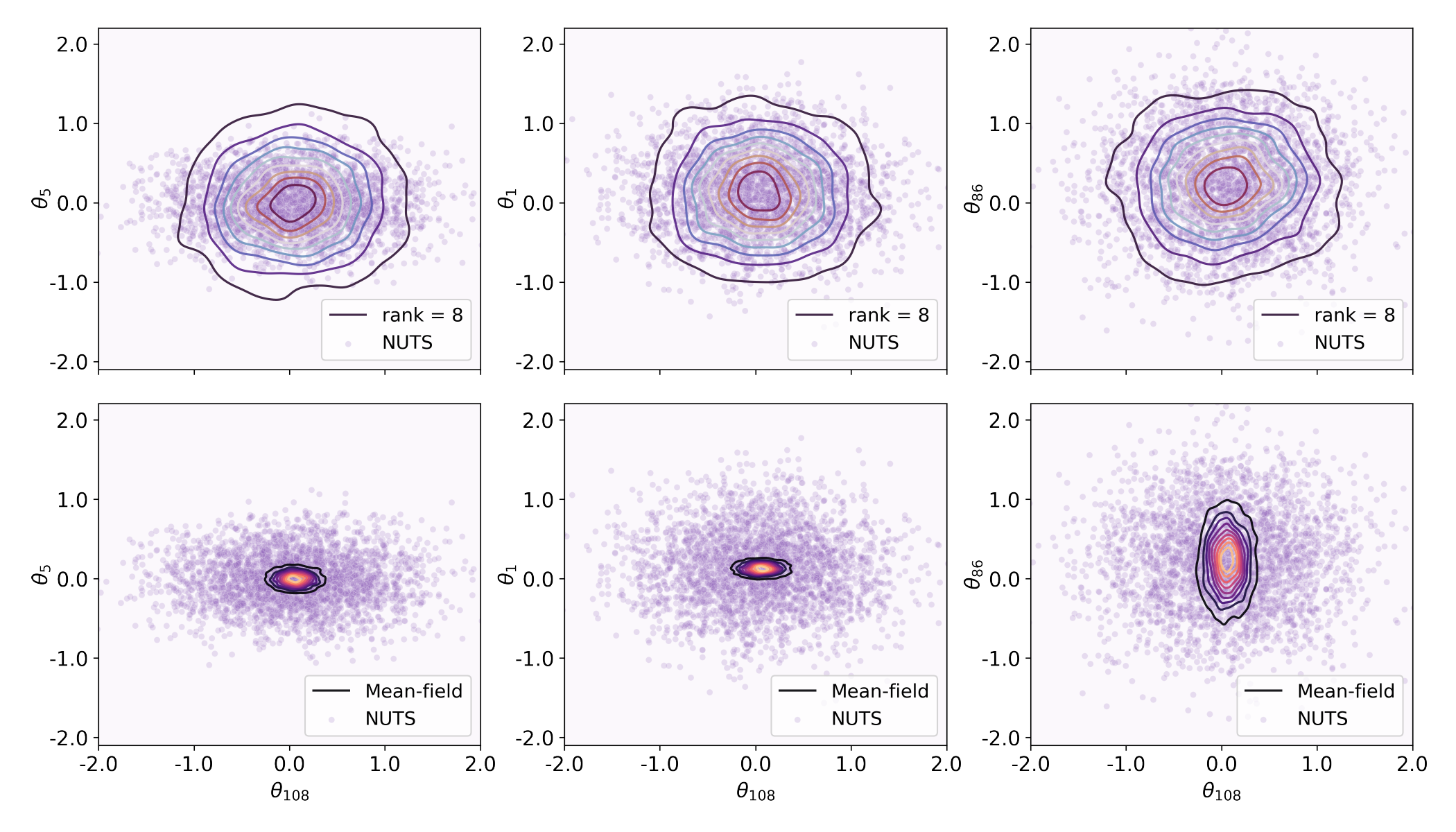}
    \caption{Low-rank inferential models lead to better approximation compared to mean-field family. Bivariate marginals of both inference models are compared against NUTS baseline for selected dimensions. Each distribution contour graph results from $3,000$ samples drawn from the baseline and its variational approximation. Top: bivariate marginals of low-rank inference model with rank $p = 8$ against NUTS baseline. Bottom: bivariate marginals of mean-field model against NUTS baseline. } 
    \label{fig:datacontour}
\end{figure}

We compare the Frobenius norm difference between baseline $\Omega^*$ obtained by the NUTS and the precision matrix $\Omega$ of the inferential models with rank $p = 4, 8, 16, 32$. The complete convergence result is provided in \Cref{fig:lr_full_without_mf}. It suggests given sufficient computational budget, higher statistical accuracy can be achieved with the more flexible high-rank inferential models.
Moreover, we have shown in \Cref{fig:Logistic_Regression} that the low-rank Gaussian inferential models outperform the standard mean-field family. More specifically, \Cref{fig:datacontour} compares bivariate marginals of the rank-$8$ inferential model and the mean-field one versus baseline obtained from the NUTS. We expect the performance difference to become more significant when the correlation among data is stronger. 

 \begin{figure}[h!]
    \centering
    \includegraphics[width=0.8\textwidth, keepaspectratio]{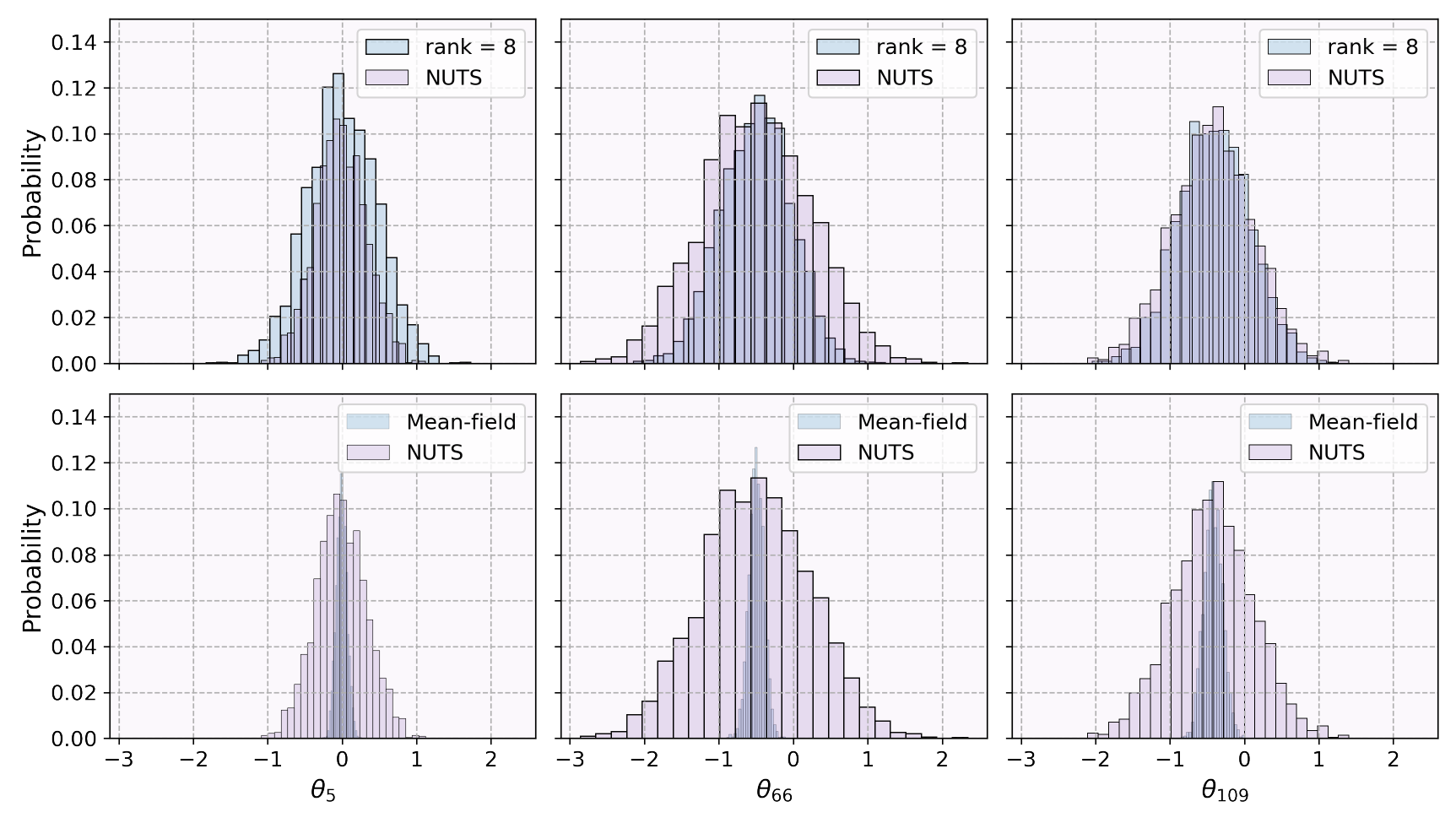}
    \caption{Low-rank variational approximation outperforms mean-field inferential model in terms of statistical accuracy. Comparison is performed against NUTS baseline given sufficient computational budget on the cardiac arrhythmia dataset (dimension d = 110). Marginal distributions are plotted for selected coordinates. Top: Low-rank inference model with rank $p=8$ against NUTS. Bottom: Mean-field model against NUTS. }
    \label{fig:Logistic_Regression_hist}
\end{figure}

\clearpage

\section{Future Works}
Extending the results in this paper to understand the statistical and computational trade-offs in other classes of inferential models is a promising avenue of future work. 
In particular, consider the mixture models. 
Another venue of possible extension is to provably improve the convergence of the variational inference methods, e.g.  leveraging gradient descent over the Bures–Wasserstein manifold for fast convergence towards the stationary solution \citep{Sinho2022}.

\clearpage


\end{document}